%% file: IJAR2024.tex
\documentclass[3p,12pt]{elsarticle}

\usepackage{amssymb}
\usepackage{amsmath,amsfonts}
\usepackage{url}
\usepackage{graphicx}
\usepackage{framed,multirow}
\usepackage{lineno}
\usepackage{xcolor}
\usepackage{colortbl}
\usepackage{bbm}
\usepackage{subfig}
\journal{International Journal of Approximate Reasoning}

\input{preambule}

\definecolor{mygray}{gray}{.9}
\begin{document}

\begin{frontmatter}

\title{Evidential time-to-event prediction  with \\
calibrated uncertainty quantification}

\author[inst1]{Ling Huang\corref{cor1}}
\ead{iweisskohl@gmail.com}
\cortext[cor1]{Corresponding author.}

\author[inst1]{Yucheng Xing}
\author[inst1,inst6]{Swapnil Mishra}
\author[inst4,inst5]{Thierry Den{\oe}ux}
\author[inst1,inst6]{Mengling Feng}

\affiliation[inst1]{organization={Saw Swee Hock School of Public Health, National University of Singapore, Singapore}}
\affiliation[inst6]{organization={Institute of Data Science, National University of Singapore, Singapore} }
\affiliation[inst4]{organization={Université de Technologie de Compiègne, CNRS, Heudiasyc, France}}
\affiliation[inst5]{organization={Institut universitaire de France, Paris, France}}
            
\begin{abstract}
Time-to-event analysis provides insights into clinical prognosis and treatment recommendations.  However, this task is more challenging than standard regression problems due to the presence of censored observations. Additionally, the lack of confidence assessment, model robustness, and prediction calibration  raises concerns about the reliability of predictions. 
To address these challenges, we propose an evidential regression model specifically designed for time-to-event prediction.
The proposed model quantifies both epistemic and aleatory uncertainties using Gaussian Random Fuzzy Numbers and belief functions, providing clinicians with uncertainty-aware survival time predictions.
The model is trained by minimizing a generalized negative log-likelihood function accounting for data censoring.
Experimental evaluations using simulated datasets with different data distributions and censoring conditions, as well as real-world datasets across diverse clinical applications, demonstrate that our model delivers both accurate and reliable performance, outperforming state-of-the-art methods. These results highlight the potential of our approach for enhancing clinical decision-making in survival analysis.
\end{abstract}


\end{frontmatter}

%\linenumbers

\section{Introduction}

Time-to-event analysis, also known as survival analysis, concerns the prediction of the time it takes for an event of interest to occur, such as time to death, disease recurrence, or treatment failure. Accurately predicting the  time of occurence of such events can provide valuable clinical insights such as risk stratification \cite{herold2020validation}, or clinical decisions such as treatment planning \cite{zhong2021gefitinib} and predictive prognostic \cite{gyorffy2010online}.

%\paragraph{Data censoring} 
Time-to-event data often include censored observations, as the exact event time is unknown for some learning instances. This feature makes the  ground-truth information partially uncertain and the estimation of survival probabilities difficult.  
The Cox proportional hazards model  \cite{cox1972regression} offers a simple approach for handling censoring using the proportional hazards (PH) assumption across covariates. The PH assumption holds that the hazard ratio (i.e., the risk of an event happening at any given time point) for two subjects is constant over time.
Faraggi and Simon \cite{faraggi1995neural} first extended the Cox model by replacing its linear predictor with a one-hidden layer multilayer perceptron (MLP). Deep learning-based Cox models such as DeepSurv \cite{katzman2018deepsurv} and Cox-CC \cite{kvamme2019time} have recently shown promising performance. 
Despite its widespread use, the PH assumption may prove unrealistic in some problems, potentially leading to biased parameter estimates and inaccurate predictions. Furthermore, the Cox model estimates the baseline hazard function solely based on observed event times, which can introduce extra biases or information loss when data are limited. For that reason, it has been proposed to relax the PH assumption. For instance, Kvamme et al. \cite{kvamme2019time} proposed a time-dependent Cox model (Cox-Time) to account for time-varying covariates. Although these methods have better prediction accuracy as compared to the baseline Cox model, their adoption in clinical decisions remains limited due to lack of confidence in their predictions.

%\paragraph{Prediction uncertainty} 
Prediction uncertainty undermines clinician trust in AI models and delays their adoption in clinical workflows, posing another challenge to time-to-event analysis. High prediction uncertainty makes it difficult to tailor treatment plans, negatively impacting patient outcomes. There are two main types of uncertainty: aleatory and epistemic uncertainty. For censored datasets,  aleatory uncertainty arises from natural variability in event times, while epistemic uncertainty is introduced by data censoring. 
Current approaches, mostly based on Bayesian inference,  quantify uncertainty based on prediction variants using techniques such as ensembling or dropout. For example, Li et al. \cite{li2020bayesian} characterize prognostic uncertainties by calculating credible intervals using a sequential Bayesian boosting algorithm, while  Chai et al. \cite{chai2024uncertainty} quantify uncertainty using the Monte Carlo dropout technique. However, these methods lack flexibility in representing and combining imperfect (i.e., uncertain, unreliable, or conflicting) information.

%\paragraph{Confidence calibration} 
The challenge of applying time-to-event models in clinical practice lies in prediction reliability. Calibration, also known as reliability in Bayesian analysis \cite{dawid1982well}, is a statistical concept that refers to the consistency of estimates relative to observed measurements, and is important in high-risk tasks such as disease diagnosis \cite{genders2011clinical} and tumor segmentation \cite{mehrtash2020confidence, huang2025deep}. Nevertheless, the calibration issue has been relatively underexplored in time-to-event models. 
Early research mainly relied on the Brier scores and negative log-likelihood, or their extensions to assess calibration performance \cite{vinzamuri2017pre, zhang2018nonparametric}.
More recently, Chapfuwa et al. \cite{chapfuwa2020calibration} accounted for prediction calibration and uncertainty with a survival-function matching estimator, which estimates and compares conditional survival distributions. However, most of the methods are based on probability calibration; as pointed out by Huang et al.  \cite{huang2024review}, probabilistic models can potentially lead to poor prediction calibration in complex data distribution scenarios.

%\paragraph{Study problem and proposals} 
In this paper\footnote{This paper is an extended version of the short paper presented at the 8th International Conference on Belief Functions (BELIEF 2024) \cite{huang2024evidential}.}, we propose a new approach towards accurate, well-calibrated, and clinically applicable time-to-event prediction via evidence-based uncertainty quantification and information fusion. The prediction model aggregates evidence about time to event within the framework of Epistemic Random Fuzzy Set theory \cite{denoeux2021belief,denoeux2023reasoning}, an extension of the Dempster-Shafer theory of evidence \cite{dempster67,shafer1976mathematical}, without relying on any restrictive distributional assumption. The proposed approach is based on the Gaussian random fuzzy number (GRFN)-based regression model introduced in \cite{denoeux22,denoeux2023quantifying}, modified to account for data censoring through a generalized negative log-likelihood optimization function. GRFNs are a newly introduced family of random fuzzy subsets of the real line that generalizes Gaussian random variables and Gaussian possibility distributions. The model outputs are the most plausible event time as well as variance and precision information quantifying, respectively, aleatory and epistemic prediction uncertainties.
 The model outputs make it possible to compute conservative prediction intervals with specified belief degrees and confidence calibration.
 
The rest of this paper is organized as follows. A review of previous work, including classical and machine-learning-based approaches to survival analysis on the one hand, and random fuzzy sets on the other hand, is first presented in Section \ref{sec: related}. Our approach is then introduced in Section \ref{sec: methods}, and experimental results are reported in Section \ref{sec: exp}. Finally, Section \ref{sec:conclu} concludes the paper and presents some directions for further research.

\section{Review of previous work}
\label{sec: related}

We started by introducing some basic definitions related to survival analysis and briefly introducing the baseline Cox model in Section \ref{subsec:basic}. A brief survey of machine-learning-based approaches to survival analysis is then presented in Section \ref{subsec: time-to-event}. Finally, necessary notions about random fuzzy sets and Gaussian random fuzzy numbers are introduced in Sections \ref{subsec: GRFN}. 

\subsection{Basic definitions and Cox model}
\label{subsec:basic}

\paragraph{Basic definitions} 
In the medical domain, the  \emph{survival function} gives the chance that a subject, described by a vector $x$ of covariates, is still alive after time $t$; it is formally defined as
\begin{equation*}
    S(t|x)=P(T\ge t|x)=\int_{t}^{+\infty}f(s|x)ds,
    \label{eq: S(t)}
\end{equation*}
where $T$ is the positive random event time and $f(\cdot|x)$ is the conditional probability density function (pdf) of $T$ given $x$. The  \emph{hazard function} is commonly used to quantify the risk of an event, such as death, occurring at a given time, given that the subject has survived up to that point. It is derived by considering the probability that  $T$ takes a value between $t$ and $t+\Delta t$, conditional on $T \geq t$, i.e., $P(t \leq T < t+\Delta t \mid T \geq t)$. This conditional probability is then normalized by dividing it by the time interval $\Delta t$, yielding a rate expressed as the probability per unit time. The hazard function is formally defined as
\begin{equation*}
    \lambda(t|x)=\lim_{\bigtriangleup t\to 0}\frac{P(t \le T <t+\Delta t\mid T\ge t; x)}{ \Delta t}=\frac{f(t|x)}{S(t|x)}.
\end{equation*}

\paragraph{Cox model} The Cox model assume the hazard function for subject $i$ to be of the form
\[
\lambda(t|x_i)=\lambda_0(t)\exp(-\beta^Tx_i),
\]
where $\lambda_0(t)$ is the baseline hazard, and $\beta$ is a vector of coefficients. This model is often referred to as the ``proportional hazard'' (PH) model because the ratio of hazards for two subjects does not depend on time. Indeed,
\[
\frac{\lambda(t|x_i)}{\lambda(t|x_j)}=\exp(-\beta^T(x_i-x_j)).
\]
Parameter $\beta$ is estimated using the maximum likelihood method.

Although this simple model has been widely adopted by the scientific community, it presents some limitations. In particular, the PH assumption is sometimes too restrictive. Even more importantly,  it cannot properly model nonlinearities and interaction effects (often present in data) out of the box. 

\subsection{Machine-learning approaches to survival analysis} 
\label{subsec: time-to-event}

Recently, machine learning (ML) and, in particular, deep learning have become valuable complements to traditional statistical methods, significantly improving the accuracy and quality of survival analysis. While ML performs well with large training datasets along with definitive ground truth, applying ML to survival analysis poses unique challenges. Collecting large-scale survival datasets is expensive, and survival times are often only partially known due to censoring caused by early-end experiments or lack of follow-up. Research in survival analysis using ML primarily focuses on two key areas: (1) using advanced learning techniques for information representation and (2) addressing data censoring through specially designed loss functions. In the following we will briefly review these two key research areas and illustrate them with some representative methods. 

\paragraph{Survival trees}

Survival trees are a variant of classification and regression trees specifically designed to handle censored data. The earliest application of tree structures for survival data was described by Ciampi et al. \cite{ciampi1981approach}. Among tree-based methods, Random Survival Forest (RSF) \cite{ishwaran2008random} is one of the most widely used non-parametric survival models. RSF extends the random forest algorithm \cite{breiman2001random} and ensemble learning technique \cite{dietterich2002ensemble} by employing the log-rank test as a splitting criterion, computing cumulative hazards at the leaf nodes, and averaging them across the ensemble. This approach provides a flexible continuous-time method that does not rely on the proportionality assumption, making it particularly effective for modeling complex survival distributions. Despite these advantages, RSF struggles with high-dimensional data and lacks interpretability. Regularized forests were introduced in  \cite{ishwaran2011random} to address the former issue. Additionally, other ensemble-based methods such as Bagging Survival Trees \cite{hothorn2004bagging}  and Boosting for survival analysis \cite{hothorn2006survival} have gained popularity for their ability to improve predictive accuracy and robustness in survival tasks. 

\paragraph{Support Vector Machines}

Adapting Support Vector Machines (SVMs) to survival analysis is appealing due to their non-linear modeling capabilities. Extensions of SVMs to survival data often involve adapting Support Vector Regression (SVR) models with survival-specific loss functions to handle censoring. For instance, Van Belle et al. \cite{van2007support} and Evers and Messow \cite{evers2008sparse} extended SVMs to maximize the concordance index for comparable pairs of observations. An efficient variant of this method was introduced by Van Belle et al. \cite{van2008survival} under the name ``SurvivalSVM''. Another approach to handling censored data focuses on predicting the relative order of survival times, ranking individuals by their likelihood of survival, as demonstrated by adaptations of the ranking SVM model \cite{shivaswamy2007support, van2011support}. More recently, P{\"o}lsterl et al. \cite{polsterl2015fast}  introduced fast training strategies for SurvivalSVM to enhance its applicability to large-scale databases. More analysis about SVM for survival analysis can be found in \cite{VANBELLE2011107}. Despite their strengths, SVM-based survival models have limited interpretability, as they focus on ranking subjects without providing explicit hazard or survival functions. This lack of interpretability reduces their clinical acceptance and adoption.

\paragraph{Neural Networks} 

Neural networks (NNs) are widely used in survival analysis because of their flexibility in modeling nonlinear relationships and handling high-dimensional data with powerful feature learning capabilities. Faraggi and Simon  \cite{faraggi1995neural} were the first to extend the Cox model by incorporating a single-hidden-layer multilayer perceptron (MLP) as a nonlinear predictor. More recently, Katzman et al. \cite{katzman2018deepsurv} revisited these models in the context of deep learning with the development of DeepSurv, demonstrating that modern neural networks can outperform classical Cox models in terms of the C-index. Other similar works include SurvivalNet \cite{christ2017survivalnet}, which fits proportional Cox models with neural networks and optimizes hyperparameters using Bayesian methods, and Zhu's model \cite{zhu2017wsisa}, which extended Cox modeling to image data by replacing the MLP in DeepSurv with a convolutional neural network for lung cancer pathology and whole-slide histopathological images. Despite these advancements, most methods remain constrained by the proportionality assumption of the Cox model. To address this issue, Kvamme et al. \cite{kvamme2019time} proposed Cox-Time, which removes the proportionality restriction and introduces an alternative loss function that works well for both proportional and non-proportional cases. However, neural network-based models are prone to overfitting, especially when applied to small datasets, and often lack interpretability, limiting their adoption in clinical settings.

\paragraph{Discrete models}
An alternative approach to time-to-event prediction is to discretize the time and compute the hazard or survival function on this predetermined time grid. Luck et al. \cite{luck2017deep} introduced a method similar to DeepSurv \cite{katzman2018deepsurv} but added a set of discrete outputs for survival predictions and used an isotonic regression loss to optimize over the time grid. Fotso \cite{fotso2018deep} proposed a multi-task neural network that parameterizes logistic regression to directly compute survival probabilities on the time grid. Lee et al. \cite{lee2018deephit} developed DeepHit, a method that estimates the probability mass function using a neural network and optimizes it with a hybrid loss function combining log-likelihood and ranking loss. Although discrete models generally have good performance, the use of time intervals can result in a loss of precision and reliance on subjective interval boundaries.

\paragraph{Discussion}

Most deep neural network models in survival analysis benchmark their performance against traditional ML models, such as RSF. However, those models are often treated as black boxes, making it difficult to understand how predictions are generated. This lack of transparency limits the trust patients and clinicians place in the predictions of these models \cite{huang2024evidential}. To address this challenge, some methods integrate Bayesian inference with survival models to enhance interpretability and quantify uncertainty \cite{raftery1996accounting, fard2016bayesian, hageman2024estimating}. For example, Raftery et al. \cite{raftery1996accounting} applied Bayesian model averaging to Cox PH models, arguing that Bayesian methods provide both interpretability and robust uncertainty reasoning. More recently, Fard et al. \cite{fard2016bayesian} proposed a framework that combines Bayesian network representations with the Accelerated Failure Time model \cite{wei1992accelerated}, extrapolating prior probabilities to future time points. Similarly, Hageman et al. \cite{hageman2024estimating} estimated uncertainty in cardiovascular risk predictions using a Bayesian survival analysis approach. 

A fundamental limitation of Bayesian inference is its reliance on prior knowledge about parameters, an often unrealistic assumption. The Dempster-Shafer theory of evidence   \cite{dempster67,shafer1976mathematical,denoeux20b} and its recent generalization,  Epistemic Random Fuzzy Set (ERFS) theory \cite{denoeux2021belief, denoeux2023reasoning} overcome this limitation and have been successfully used in ML and medical image processing (see, e.g., \cite{huang2023application, huang2024review}.  However, the application of this framework to survival analysis remains to be explored. In the rest of this section, we provide a brief introduction to ERFS, with a particular focus on Gaussian random fuzzy numbers, a parameterized model of random fuzzy subsets of the real line used in this paper.

\subsection{Gaussian random fuzzy numberq}

\label{subsec: GRFN}

The Dempster-Shafer (DS) theory of evidence \cite{dempster67, shafer1976mathematical,denoeux20b} is a framework for modeling, reasoning with, and combining imperfect information. Most applications of DS theory consider belief functions on finite frames
of discernment \cite{xu2016evidential, lian2018joint, tong2021evidential, huang2023application}, mainly due to the lack of practical belief function models that can handle continuous variables while remaining compatible with Dempster's rule of combination. Recently, Epistemic Random Fuzzy Set (ERFS) theory \cite{denoeux2021belief, denoeux2023reasoning, denoeux2024uncertainty, denoeux2024combination}, has addressed this limitation. ERFS theory extends both DS theory and possibility theory \cite{zadeh1978fuzzy}, representing uncertain and/or fuzzy evidence as random fuzzy sets that induce belief functions. Within ERFS, independent pieces of evidence are combined using a generalized product-intersection rule, which unifies Dempster's rule and the normalized product-intersection operator of possibility theory.  Gaussian random fuzzy numbers (GRFNs), a parameterized family of random fuzzy subsets of the real line closed under the product-intersection rule, have shown great promise for quantifying both aleatory and epistemic uncertainty in prediction tasks. These notions are briefly summarized below.

\paragraph{Random fuzzy sets} Let $(\Omega,\Sigma_\Omega,P)$ and $(\Theta,\Sigma_\Theta)$ be, respectively, a probability space and a measurable space. A random fuzzy set (RFS) $\tX$ is a mapping from $\Omega$ to the set $\calF(\Theta)$ of fuzzy subsets of $\Theta$ such that, for any $\alpha\in[0,1]$, the mapping $^\alpha\tX: \omega \mapsto {^\alpha}\tX(\omega)$, where ${^\alpha}\tX(\omega)$ is the $\alpha$-cut of $\tX(\omega)$, is strongly measurable \cite{nguyen78}. In ERFS theory,  $\Omega$ represents a set of interpretations of a piece of evidence about a variable $X$ taking values in $\Theta$. If $\omega\in\Omega$ holds, the value of $X$ is constrained by possibility distribution $\tX(\omega)$; the possibility and necessity that $X\in A$ for any $A\in \Sigma_\Theta$ are, respectively,
\[
\Pi_{\tX(\omega)}(A)=\sup_{\theta\in A}\tX(\omega)(\theta)
\]
and
\[
N_{\tX(\omega)}(A)=1-\Pi_{\tX(\omega)}(A^c),
\]
where $A^c$ denotes the complement of $A$ in $\Theta$. The expected possibility and necessity of $A$ are, then,
\[
Pl_\tX(A)=\int_\Omega \Pi_{\tX(\omega)}(A) dP(\omega)
\]
and $Bel_\tX(A)=1-Pl_\tX(A^c)$. It can be shown that mapping $Bel_\tX: A\mapsto Bel_\tX(A)$ is a belief function, and $Pl_\tX$ is the dual plausibility function \cite{couso11}. 

\paragraph{Product-intersection rule} Let us now consider two pieces of evidence represented by probability spaces $(\Omega_1,\Sigma_1,P_1)$ and $(\Omega_2,\Sigma_2,P_2)$ and mappings $\tX_1:\Omega_1\rightarrow \calF(\theta)$ and $\tX_2:\Omega_1\rightarrow \calF(\theta)$. Assuming the two pieces of evidence to be independent, they can be jointly represented by the mapping 
\[
\begin{array}{rll}
\tX_1\oplus\tX_2:\Omega_1\times\Omega_2 &\rightarrow &\calF(\Theta)\\
(\omega_1,\omega_2) & \mapsto &\tX_1(\omega_1)\odot \tX_2(\omega_2),
\end{array}
\]
where $\odot$ denotes the normalized product intersection of fuzzy sets, and the probability space $(\Omega_1\times\Omega_2,\Sigma_1\times\Sigma_2,\tP_{12})$, where $\tP_{12}$ is the probability measure obtained by conditioning the product measure $P_1\times P_2$ on the fuzzy subset $\tTheta$ if consistent pairs of interpretations, defined by $\tTheta(\omega_1,\omega_2)=\sup_{\theta\in\Theta} \tX_1(\omega_1)(\theta) \cdot \tX_1(\omega_2)(\theta)$. This combination rule is commutative and associative; it extends Dempster's rule of combination \cite{shafer1976mathematical}. The unnormalized product-intersection rule $\boxplus$ is similar to $\oplus$, but without the conditioning step. It can be seen as an approximation of $\oplus$ when the conflict between pieces of evidence is small.  More details about ERFS theory can be found in \cite{denoeux2023reasoning}.

\paragraph{Gaussian random fuzzy numbers (GRFNs)} Before introducing GRFNs, it is necessary to define Gaussian fuzzy numbers (GFNs). A GFN with mode $m\in\reels$ and precision $h\in\reels_+$ is a fuzzy subset of $\reels$ with  membership function
\begin{equation*}
    x \mapsto \exp\left(-\frac{h}{2} (x-m)^2\right).
\end{equation*}
It is denoted as $\GFN(m,h)$. The family of GFNs is closed under the normalized product intersection: for any  two GFNs, $\GFN(m_1, h_1)$ and $\GFN(m_2, h_2)$, we have $\GFN(m_1, h_1)\odot\GFN(m_2, h_2)=\GFN(m_{12}, h_1+h_2)$  with 
\begin{equation*}
    m_{12}=\frac{h_1 m_1+h_2 m_2}{h_1+h_2}. 
%    \label{eq: m_12}
\end{equation*}

A GRFN can be defined as a GFN whose mode is a Gaussian random variable (GRV) $M\sim N(\mu, \sigma^2)$. It is, thus, defined by (1) the probability space $(\reels,\calB(\reels),P_{\mu,\sigma^2})$, where $\calB(\reels)$ is the Borel sigma-algebra on $\reels$ and $P_{\mu,\sigma^2}$ is the Gaussian probability measure on $\reels$ with parameters $(\mu,\sigma^2)$, and (2) the mapping $\tX: M \mapsto \GFN(M,h)$. We note $\tX\sim\tN(\mu,\sigma^2,h)$. As shown in \cite{denoeux2023reasoning}, the contour function of GRFN $\tX\sim \tN(\mu,\sigma^2,h)$  is
\begin{equation}
\label{eq:pl}
pl_\tX(x)=\frac{1}{\sqrt{1+h\sigma^2}}\exp\left(- \frac{h(x-\mu)^2}{2(1+h\sigma^2)}\right)
\end{equation}
and the degrees of belief and plausibility of a real interval $[x,y]$  are, respectively,
\begin{multline}
\label{eq:belint}
Bel_\tX([x,y])=\Phi\fracpar{y-\mu}{\sigma} -\Phi\fracpar{x-\mu}{\sigma} - \\
pl_\tX(x)\left[\Phi\fracpar{(x+y)/2-\mu+(y-x)h\sigma^2/2}{\sigma\sqrt{h\sigma^2+1}}-\Phi\fracpar{x-\mu}{\sigma\sqrt{h\sigma^2+1}} \right]- \\
pl_\tX(y)\left[\Phi\fracpar{y-\mu}{\sigma\sqrt{h\sigma^2+1}}-\Phi\fracpar{(x+y)/2-\mu-(y-x)h\sigma^2/2}{\sigma\sqrt{h\sigma^2+1}}\right],
\end{multline}
and
\begin{multline}
\label{eq:plint}
Pl_\tX([x,y])=\Phi\fracpar{y-\mu}{\sigma} -\Phi\fracpar{x-\mu}{\sigma} + pl_\tX(x)\Phi\fracpar{x-\mu}{\sigma\sqrt{h\sigma^2+1}} + \\
pl_\tX(y)\left[1-\Phi\fracpar{y-\mu}{\sigma\sqrt{h\sigma^2+1}}\right],
\end{multline}
where $\Phi$ denotes the standard normal cumulative distribution function (cdf). In particular, the lower and upper cdfs of $\tX$ are, respectively,
\begin{equation*}
%\label{eq:lowercdf}
Bel_\tX((-\infty,y])=\Phi\fracpar{y-\mu}{\sigma}  - pl_\tX(y) \Phi\fracpar{y-\mu}{\sigma\sqrt{h\sigma^2+1}}
\end{equation*}
and
\begin{equation*}
%\label{eq:uppercdf}
Pl_\tX((-\infty,y])=\Phi\fracpar{y-\mu}{\sigma} + pl_\tX(y)\left[1-\Phi\fracpar{y-\mu}{\sigma\sqrt{h\sigma^2+1}}\right].
\end{equation*}
As shown in \cite{denoeux2023reasoning}, the product-intersection of two independent GRFNs $\tX_1\sim \tN(\mu_1,\sigma_1^2,h_1)$ and $\tX_2\sim \tN(\mu_2,\sigma_2^2,h_2)$ is a GRFN $\tX_{12} \sim \tN(\mu_{12},\sigma^2_{12},h_1+h_2)$. Formulas for $\mu_{12}$ and $\sigma^2_{12}$ are given in \cite{denoeux2023reasoning}, and generalized in \cite{denoeux2024combination} for the case of $n$ dependent GRFNs. The unnormalized combination of $\tX_1$ and $\tX_2$ is easily calculated as 
\begin{equation}
\label{eq:comb}
    \tX_1\boxplus\tX_2 \sim \tN\left(\frac{h_1\mu_1+h_2\mu_2}{h_1+h_2}, \frac{h_1^2\sigma^2_1+h_2^2\sigma^2_2}{(h_1+h_2)^2},h_1+h_2\right).
\end{equation}

\paragraph{Lognormal random fuzzy numbers}
%\label{subsec: tGRFN}
A GRFN models uncertainty about a variable that can take any value on the real line. However, in time-to-event analysis, the response variable is always positive. For such tasks, event prediction and the corresponding uncertainty are more appropriately represented by a \emph{lognormal random fuzzy number} introduced in \cite{denoeux2023parametric}. In general, let $\tX:\Omega\rightarrow\calF(\theta)$ denote a RFS, and let $\psi$ be a one-to-one mapping from $\Theta$ to a set $\Lambda$. Zadeh's extension principle \cite{zadeh75} allows us to extend mapping $\psi$ to fuzzy subsets of $\Theta$; specifically, we  define mapping $\tpsi:  \calF(\Theta)  \rightarrow \calF(\Lambda)$ as
\begin{equation*}
\label{eq:ext_princ}
\forall \calF(\Theta), \quad \tpsi(\tF)(\lambda)=\sup_{\lambda=\psi(\theta)} \tF(\theta)=\tF(\psi^{-1}(\lambda)).
\end{equation*}
By composing $\tpsi$ with $\tX$, we obtain a new RFS $\tpsi\circ\tX: \Omega \rightarrow \calF(\Lambda)$. For any measurable subset $C$ of $\Lambda$, we have 
\begin{equation}
\label{eq:belpl}
Bel_{\tpsi\circ\tX}(C)=Bel_{\tX}(\psi^{-1}(C)) \quad \text{and} \quad  Pl_{\tpsi\circ\tX}(C)=Pl_{\tX}(\psi^{-1}(C)).
\end{equation}
Furthermore, given two RFS $\tX_1: \Omega_1\rightarrow \Theta$ and $\tX_2: \Omega_2\rightarrow \Theta$, we have
\[
\tpsi\circ(\tX_1\oplus\tX_2)=(\tpsi\circ\tX_1) \oplus (\tpsi\circ\tX_2).
\]
Taking $\tX\sim \tN(\mu,\sigma^2,h)$, $\Lambda=\reels_+$ and $\psi=\exp$, we obtain a lognormal random fuzzy number $\tY:\widetilde{\exp}\circ \tX$,  denoted by $\tY\sim T\tN(\mu,\sigma^2,h,\log)$. We can remark that $\tY\sim T\tN(\mu,\sigma^2,h,\log)$ iff $\widetilde{\log}\circ \tY\sim \tN(\mu,\sigma^2,h)$.

\section{Proposed method}
\label{sec: methods}

\begin{figure}
\centering
\includegraphics[width=\textwidth]{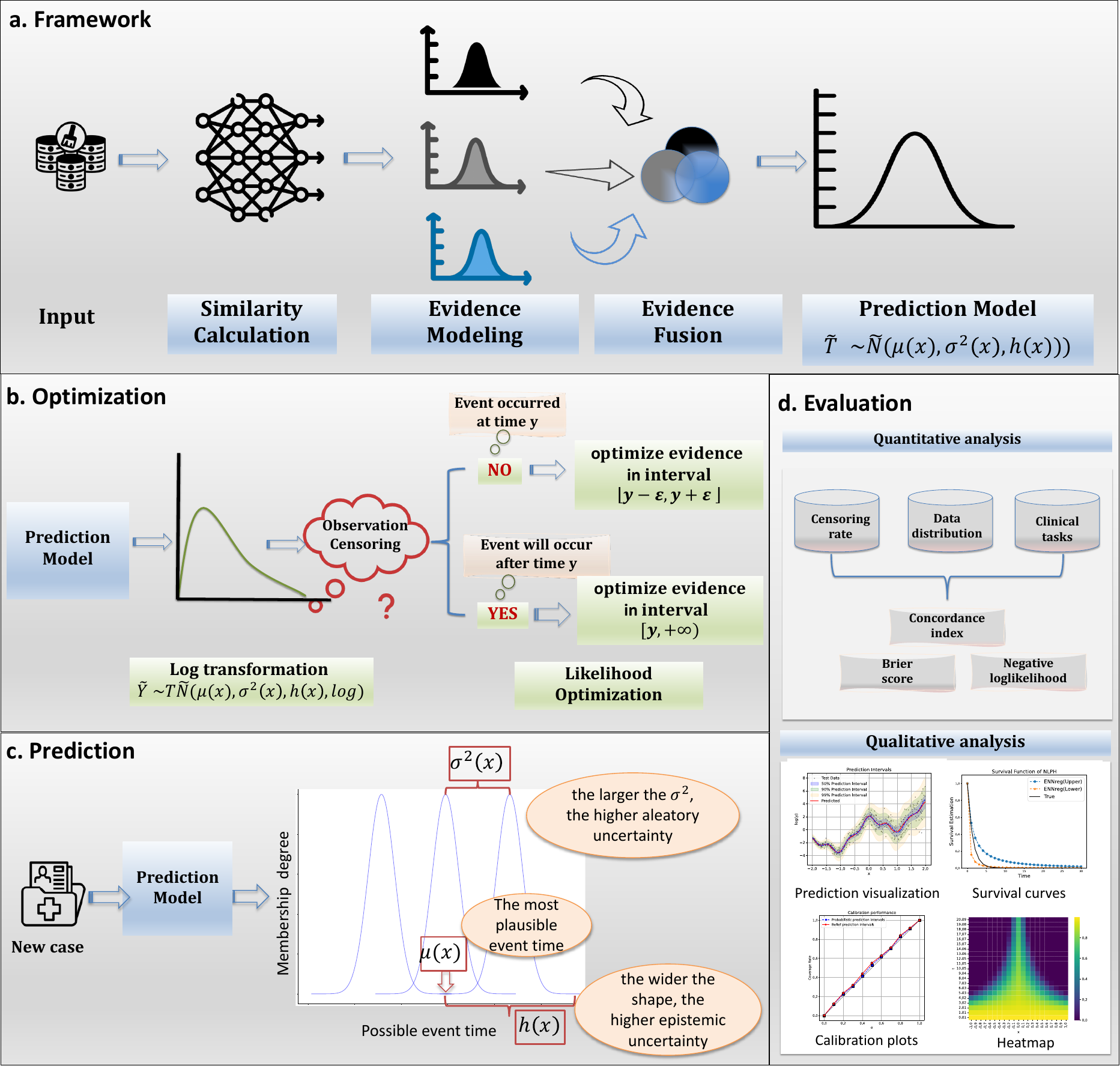}
\caption{Overview of the study. }
\label{fig: overall}
\end{figure}

%\subsection{Study overview}
Figure \ref{fig: overall} shows an overview of this study. The time-to-event prediction model can be decomposed into similarity calculation, evidence modeling, and evidence fusion steps. First, similarities between the input vector and prototypes are computed. The evidence from each prototype is then represented by a GRFN, and the prototype-based GRFNs are combined in the evidence fusion layer. The model is fit by minimizing a generalized negative log-likelihood loss function considering both censored and uncensored observations. The optimized model outputs a GRFN indicating the most plausible event time as well as measures of aleatory and  epistemic uncertainties about the predicted time. The evaluation of the framework includes both quantitative analysis (i.e., censoring rate, data distribution, and clinical tasks) and qualitative analysis (i.e., survival heatmaps and curves,  calibration plots). The prediction model is described below in Section \ref{subsec:ENNreg}, and the loss function is detailed in Section \ref{subsec:loss}.

\subsection{Evidential time-to-event prediction} 
\label{subsec:ENNreg}

To predict event times, we adopt the evidential regression model ENNreg introduced in \cite{denoeux2023quantifying}. We construct a GRFN-based evidential time-to-event prediction model that predicts the response $Y = \log T$, where $T$ represents the event time. Using ENNreg,   evidence about the event time is obtained by comparing the input vector $x$ with prototypes, and modeled using GRFNS. These GRFNs are then combined using  the unnormalized product-intersection rule. Similar to the model presented in \cite{denoeux2023quantifying}, the evidential time-to-event prediction model can be decomposed into a radial basis function (RBF) layer computing similarities to prototypes, an evidence mapping layer and an evidence fusion layer.

\paragraph{RBF layer} Let $p_1, \ldots, p_K$ denote $K$ vectors in the $p$-dimensional feature space, called  prototypes. The similarity between the input vector $x$ and each prototype $p_k$ is computed as
   \begin{equation*}
       s_k(x)=\exp(-\gamma_k^2 \| x-p_k \|^2),
%       \label{eq: si}
   \end{equation*}
where $\gamma_k$ is a positive scale parameter. 

\paragraph{Evidence mapping} 

In the evidence mapping layer, the evidence of prototype $p_k$ is represented by a GRFN $\tY_k(x) \sim \tilde{N}(\mu_k (x), \sigma_k^2, s_k(x)h_k)$ with mean  $\mu_k (x)$ defined as  a linear combination of the inputs, 
   \begin{equation*}
 %  \label{eq:mu}
       \mu_k(x)=\beta_{k}^{T}x +\beta_{k0},
   \end{equation*} 
where $\beta_{k}$ is a $p$-dimensional vector of coefficients, and $\beta_{k0}$ is a scalar parameter. The response variable $Y$ is, thus, locally approximated by a linear function of $x$, in a neighborhood of prototype $p_k$. The quantities $\mu_k(x)$ and $\sigma^2_k$ can be seen as estimates of the conditional expectation and variance, respectively, of the response $Y=\log T$ given that $x$ is close to $p_k$. The precision of $\tY(x)$ depends on a parameter $h_k$ and on the similarity $s_k(x)$ between input $x$ and prototype $p_k$. The farther $x$ is from $p_k$, the more imprecise is $\tY_k(x)$. In the limit, as $s_k(x)\rightarrow 0$, the precision tends to zero, and the belief function becomes totally uninformative ($Bel_{\tY(x)}(A)=0$ for all measurable $A\subset\reels$). 

\paragraph{Evidence fusion} 
The evidence fusion layer combines evidence from the $K$ prototypes using the unnormalized product-intersection rule \eqref{eq:comb}. The combined GRFN 
\begin{subequations}
\begin{equation}
\label{eq:mux}
\tY(x) \sim \tN(\mu(x),\sigma^2(x),h(x))
\end{equation}
is given by 
\begin{equation}
\mu(x)=\frac{\sum_{k=1}^{K} s_k(x)h_k\mu_k(x)}{\sum_{k=1}^{K} s_k(x)h_k}, \quad
\sigma^2(x)=\frac{\sum_{k=1}^{K}s^2_k(x)h^2_k\sigma^2_k}{(\sum_{k=1}^{K} s_k(x)h_k)^2},
\end{equation}
\end{subequations}
 and 
$
h(x)=\sum_{k=1}^{K}s_k(x)h_k.
$
RFS $\tY(x)$ represents uncertainty about the response $Y=\log T$, while  uncertainty about the time to event $T=\exp(Y)$ is described by lognormal random fuzzy number $\tT(x) \sim T\tN(\mu(x),\sigma^2(x),h(x),\log)$. The belief and plausibility that $T$ belongs to any interval $[t_1,t_2]$ can be computed from \eqref{eq:belint}-\eqref{eq:plint} and \eqref{eq:belpl} as
\[
Bel_{\tT(x)}([t_1,t_2])=Bel_{\tY(x)}([\log t_1,\log t_2])
\]
and
\[
Pl_{\tT(x)}([t_1,t_2])=Pl_{\tY(x)}([\log t_1,\log t_2]).
\]
The contour function of $\tT(x)$ is
\[
pl_{\tT(x)}(t)=
pl_{\tY(x)}(y))=\frac{1}{\sqrt{1+h(x)\sigma^2(x)}}\exp\left(- \frac{h(x)[\log t-\mu(x)]^2}{2[1+h(x)\sigma^2(x)]}\right).
\]
The mode of $pl_{\tT(x)}$ is $\exp(\mu(x))$, which is the most plausible value of $T$ (while $\mu(x)$ is the most plausible value of $Y=\log T$). 
The variance output $\sigma^2(x)$ estimates the conditional variance of $Y$ given the input $x$, reflecting aleatory uncertainty (larger $\sigma^2(x)$ indicates more uncertainty). The precision output $h(x)$ gets smaller and tends to zero when the distances to prototypes increase, reflecting epistemic uncertainty (smaller $h(x)$ corresponds to higher uncertainty).

Finally, we note that it is possible, from the predicted RFSs, to compute lower and upper estimated conditional survival functions as, respectively,
\begin{subequations}
\label{eq:lower_upper}
\begin{equation}
\label{eq:lower}
S_*(t|x)=Bel_{\tT(x)}([t,+\infty))=Bel_{\tY(x)}([\log t,+\infty))
\end{equation}
and
\begin{equation}
\label{eq:upper}
S^*(t|x)=Pl_{\tT(x)}([t,+\infty))=Pl_{\tY(x)}([\log t,+\infty)).
\end{equation}
\end{subequations}

%%%%%%%%%%%%%%%%%%%%%%%%%%%%%%%%%%%%%%%%%%%%%%%%%%%%%%%%%%%%%%%%%%%%%%%%%%%%%%%%%%%%%%%%%%%%%%%%%%%%%%%%%%%%%%%%%%%%%%%%%%%%%%

%\subsection{Prediction Calibration} 

\subsection{Optimization with data censoring}
\label{subsec:loss}

In medical applications, the true event times may be unknown for some subjects. This can occur when the follow-up time for a subject is not long enough for the event to happen, when a subject quits the study before its termination, or when there is imprecision in the data collection. Instead of observing the true event time $t$, we then observe a possibly right-censored event time $t^*=\min (t, \tau)$, where $\tau$ is the censoring time. In addition, we observe the binary variable $d= I(t=t^*)$ indicating whether the observed time $t^*$ is equal to the true event time ($d=1$) or not ($d=0$).

Let $\tY$ be the output GRFN given by \eqref{eq:mux}, $y^*=\log t^*$ the logarithm of the observed time, and $d$ the censoring variable for some subject. To optimize the proposed framework, we extend the negative log-likelihood loss function defined in \cite{denoeux2023quantifying} and adapt it to account for data censoring by optimizing the evidence in two situations: 
\begin{itemize}
    \item Even if the observation is not censored, the continuous event time $y^*$ is always observed with finite precision. Therefore, instead of observing an exact value, we actually observe an interval $[y^*]_{\epsilon}=[y^*-\epsilon, y^*+\epsilon]$ assumed to contain $y$, for some small $\epsilon>0$. The degrees of belief and plausibility of what has been observed are thus, respectively, $Bel([y^*]_{\epsilon})$ and $Pl([y^*]_{\epsilon})$.
    \item If the observation is censored, we only know that $y\in [y^*,\infty)$.  The degrees of belief and plausibility of the observation are, then, $Bel([y^*,\infty))$ and $Pl([y^*,\infty))$.
\end{itemize}
The generalized negative loss function introduced in \cite{denoeux2023quantifying} can thus be adapted to censored data as follows:
\begin{equation*}
     \calL_{\lambda,\epsilon}(\tY,y^*,d)= \eta \overline{\calL}(\tY,y^*,d)+(1-\eta) \underline{\calL}(\tY,y^*,d),
 \end{equation*}
 where 
\begin{align*}
    \overline{\calL}(\tY,y^*,d)&=- d \log Bel_{\tY}([y^*-\epsilon,y^*+\epsilon]) - (1-d) \log Bel_{\tY}([y^*,\infty)),\\
    \underline{\calL}(\tY,y^*,d)&=- d \log Pl_{\tY}([y^*-\epsilon,y^*+\epsilon])- (1-d)\log Pl_{\tY}([y^*,\infty)),
\end{align*}
and $\eta$ is a hyperparameter that controls the cautiousness of the prediction (the smaller, the more cautious). We found empirically that a higher censoring rate requires a larger $\lambda$. The whole network is trained by minimizing the regularized average loss:
\[
C_{\eta,\epsilon,\xi,\rho}(\Psi)= \frac1n\sum_{i=1}^n \calL_{\lambda,\epsilon}(\tY(x_i;\Psi),y^*_i,d_i) + \frac{\xi}K \sum_{k=1}^K h_k + \frac{\rho}K \sum_{k=1}^K \gamma_k^2,
\]
where $n$ is the number of observations, $\Psi$ is the vector of all parameters (including the prototypes), $\tY(x_i;\Psi)$ is the network output for input $x_i$, $\xi$ and $\rho$ are two regularization coefficients.  The first regularizer coefficients $\xi$ reduces the impact of prototypes used for prediction, for example, setting $h_k = 0$ amounts to discard prototype $k$. The second regularizer hyperparameter $\rho$ shrinks the solution towards a linear model \cite{denoeux2023quantifying}.
%-------------------------------------------------------------------------

\section{Experimental  results} 
\label{sec: exp}
In this section, the proposed framework described in Section \ref{sec: methods} is evaluated through a series of experiments. First, an illustrative example is used to assess its robustness to data censoring. Next, the framework is tested on three simulated survival datasets to evaluate its robustness to different data distributions. Finally, it is validated on real-world survival and intensive care unit (ICU) datasets to demonstrate its applicability in real-world settings. The evaluation criteria are first described in Section \ref{subsec: criteria}. Experimental results are then reported in Sections \ref{subsec: exp_illustrative}, \ref{subsec: simulated}, and \ref{subsec: real_world}.

\subsection{Evaluation criteria}
\label{subsec: criteria}

Evaluation criteria for time-to-event prediction involve assessing how well a model predicts time-to-event outcomes and how effectively it handles censored data. 
The \emph{concordance index} is one of the most widely used criteria that measure how well the model can correctly predict the order of events \cite{harrell1982evaluating}. Since it depends only on the ordering of the predictions, this criterion allows us to use the relative risk function instead of a metric for predicting survival time. Antolini et al. \cite{antolini2005time} proposed a time-dependent   concordance index defined as
\begin{equation}
\label{eq:Cidx}
   C_{idx}=P\{S(T_i\mid x_i) < S(T_i\mid x_j )\mid T_i< T_j, D_i=1\},
\end{equation}
which is more flexible. To account for tied event times and survival estimates, the authors make the modifications listed by Ishwaran et al. \cite{ishwaran2008random} to ensure that predictions independent of $x$, $\widehat{S}(t| x) = \widehat{S}(t)$, yield $C_{idx} = 0.5$ for unique event times. In this paper, we chose time-dependent $C_{idx}$ to evaluate how well our model predicts the ordering of patient death times.

The \emph{Brier score} (BS) is a common metric for evaluating the calibration of a model's predictions. To get binary value from time-to-event data, we chose a fixed time $t$ and labeled data according to whether or not an observed event time was shorter or longer than $t$. 
In \cite{graf1999assessment}, Graf et al. generalize the Brier score to account for censoring by weighting the scores by the inverse censoring distribution,
\begin{equation*}
    BS(t)=\frac{1}{N}  \sum_{1}^{N} \left [\frac{\widehat{S}(t|x_i)^2 I \{T_i \le t, D_i=1\} }{\widehat{G}(T_i)}+ \frac{(1-\widehat{S}(t|x_i))^2 I \{T_i > t\} }{\widehat{G}(t)} \right ],
\end{equation*}
where $N$ is the number of observations, $\widehat{G}(t)$ is the Kaplan-Meier estimate of the censoring survival function. The BS can be extended from a single duration $t$ to an interval by computing the integrated Brier score (IBS) as
\begin{equation}
\label{eq:IBS}
    IBS=\frac{1}{t_2-t_1}\int_{t_1}^{t_2} BS(t) dt.
\end{equation}
In practice,  this integral can be approximated by numerical integration with the time span equal to the duration of the test set. We followed the suggestion in \cite{kvamme2019time} and used 100 grid points to obtain stable scores.

\emph{Binomial log-likelihood} (BLL) is another commonly used metric in regression tasks that measures the calibration of the estimations. Using the same inverse censoring weighting as for the Brier score, we can apply this metric to censored duration time data,
\begin{equation*}
    BLL(t)=\frac{1}{N}\sum_{i=1}^{N} \left [ \frac{\log[1-\widehat{S}(t|x_i)] I  \{T_i \le t, D_i=1\} }{\widehat{G}(T_i)}+ \frac{\log[\widehat{S}(t|x_i)] I  \{T_i >  t  \} }{\widehat{G}(t)}  \right ].
\end{equation*}
We can also extend this metric from a single duration $t$ to an interval by computing the integrated binomial log-likelihood (IBLL) as
\begin{equation*}
    IBLL=\frac{1}{t_2-t_1}\int_{t_1}^{t_2} BLL(t) dt.
\end{equation*}

In this paper, following \cite{denoeux2023quantifying}, we also use  \emph{Belief prediction intervals} (BPIs)  to evaluate the calibration of an evidential predictive model and, therefore, to study the reliability of the model's predictions. 
BPIs are intervals centered around the most plausible response, $\mu(x)$, with a degree of belief $\alpha$ that the true value of the response variable lies within the interval. An $\alpha$-level BPI $\calB_\alpha(x)$ thus verifies $Bel_{\tilde{Y}(x)} (\calB_\alpha(x)) = \alpha$. Predictions are considered calibrated if, for all $\alpha \in (0, 1]$, the $\alpha$-level BPIs have a coverage probability at least equal to $\alpha$, i.e.,
\begin{equation}
    \forall \alpha \in (0, 1], \quad P_{X,Y} (Y \in \mathcal{B}_\alpha (X))\ge \alpha.
\end{equation}
Similarly to the probabilistic case, the calibration of evidential predictions can be evaluated using a calibration plot. Predictions are considered calibrated if the curve lies above the first diagonal, with closer proximity to the diagonal indicating higher precision in the predictions.

\subsection{Illustrative dataset}
\label{subsec: exp_illustrative}

\begin{figure}
%\colorbox{gray!20}{\parbox{0.99\textwidth}{\centering \textbf{No data censoring}}}\\
%\vspace{-1cm}
\begin{center}
    \subfloat[\label{fig: v2_0}No censoring]{\includegraphics[width=0.45\textwidth]{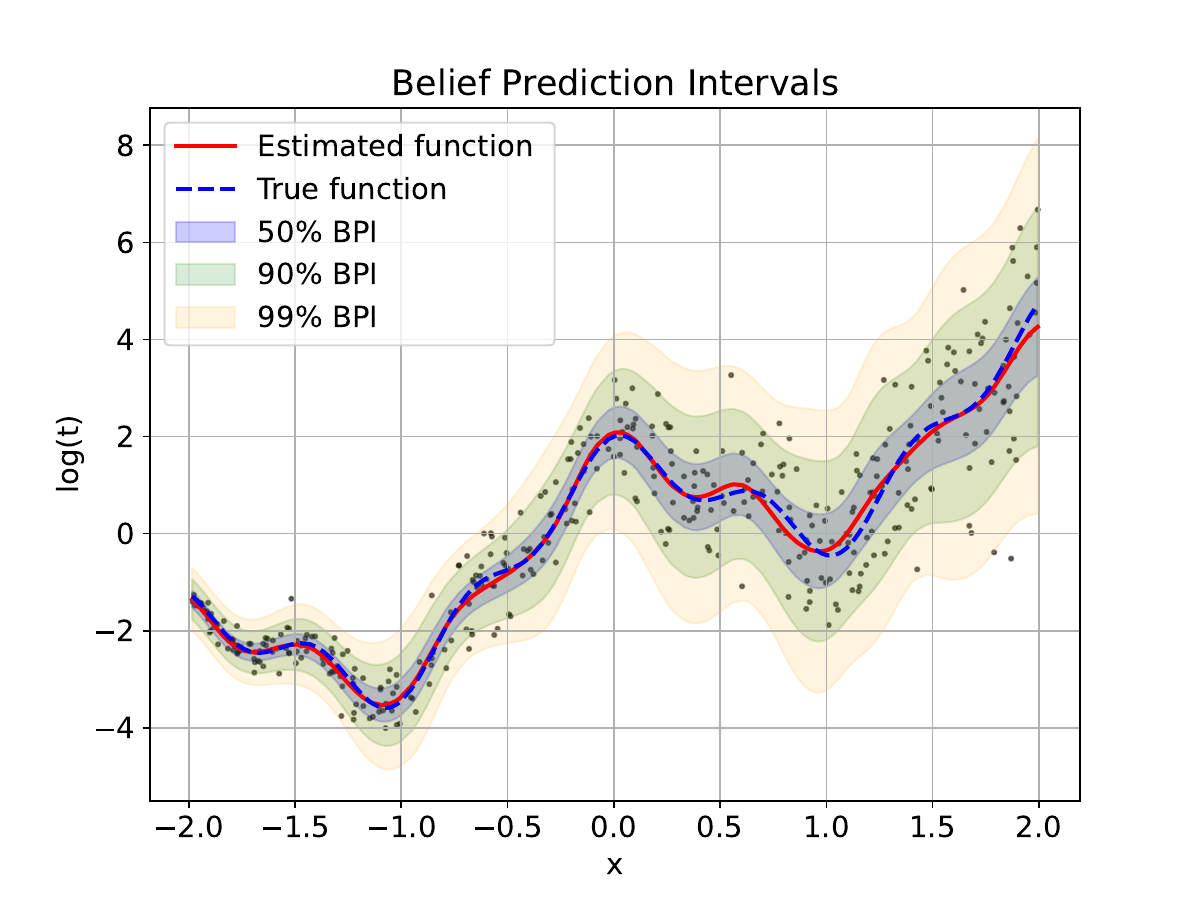}}
\subfloat[\label{fig: v2_0.5}50\% censoring]{\includegraphics[width=0.45\textwidth]{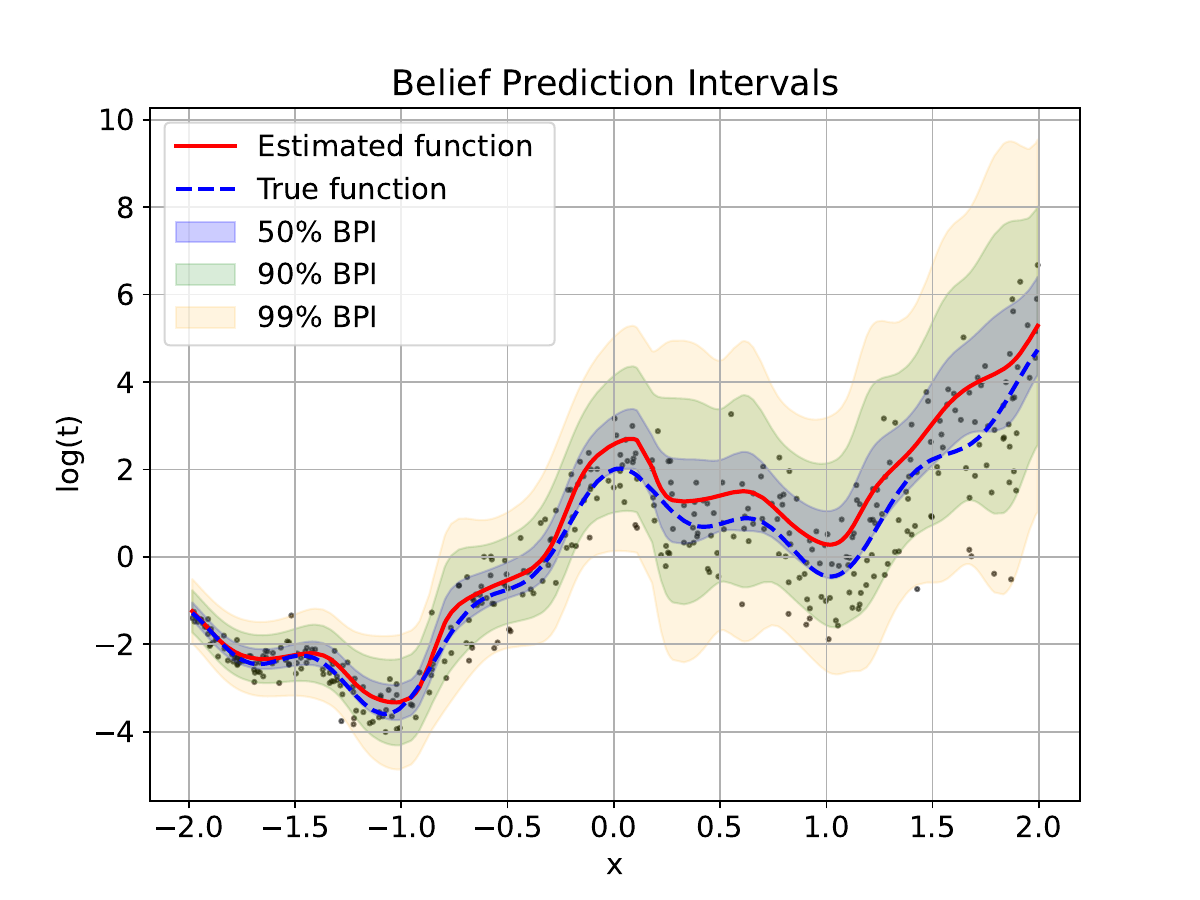}}
\end{center}
\caption{Visualized prediction performance and uncertainty on the illustrative dataset. Figures (a) and (d) show the simulated data, actual regression function (blue broken lines), and predicted function obtained from the trained model (red solid lines) for 0\% and 50\% censoring rates. Belief prediction intervals (BPIs) at levels $\alpha \in \{0.5, 0.9, 0.99\}$ are represented by shaded areas in blue, green, and orange. } 
\label{fig:pred_int}
\end{figure}

%\paragraph{Robustness to data censoring}
We first present experiments with a simple simulated dataset to illustrate the model's robustness to data censoring and calibration property. The  dataset was constructed with the following distribution: the input $X$ has a uniform distribution in the interval $[-2, 2]$, and the response is
\begin{equation}
T=\exp \left[1.5 X+2 \cos (3 X)^{3}+\frac{X^2+5}{3 \sqrt{5}} V\right],
\end{equation}
where $V \sim N(0, 1)$ is a standard normal random variable. Two elements were incorporated to simulate data censoring scenarios: the event censoring state indicator $d$ and a random censoring time $\tau$. The censored samples were randomly selected with a probability $p$ and assigned with $d=1$, and the uncensored samples were left with $d=0$. The censoring time was drawn from a uniform distribution between 0 and the maximum event time. For each censored sample, the event time was set to the minimum of the original time value and the randomly generated censoring time. Finally, a dataset with 2000 samples was generated, of which 1600 samples were used for training and 400 for testing. Two versions were considered, with censoring rates equal to 0 and 50\%. % and 90\%. %The model was initialized with $K = 40$ prototypes to train this dataset. 

\begin{figure}
\centering
%\colorbox{gray!20}{\parbox{0.9\textwidth}{\centering \textbf{Original reliability diagrams}}}\\
%\vspace{-0.2cm}
\subfloat[\label{fig: v2_0_c}No censoring]{\includegraphics[width=0.4\textwidth]{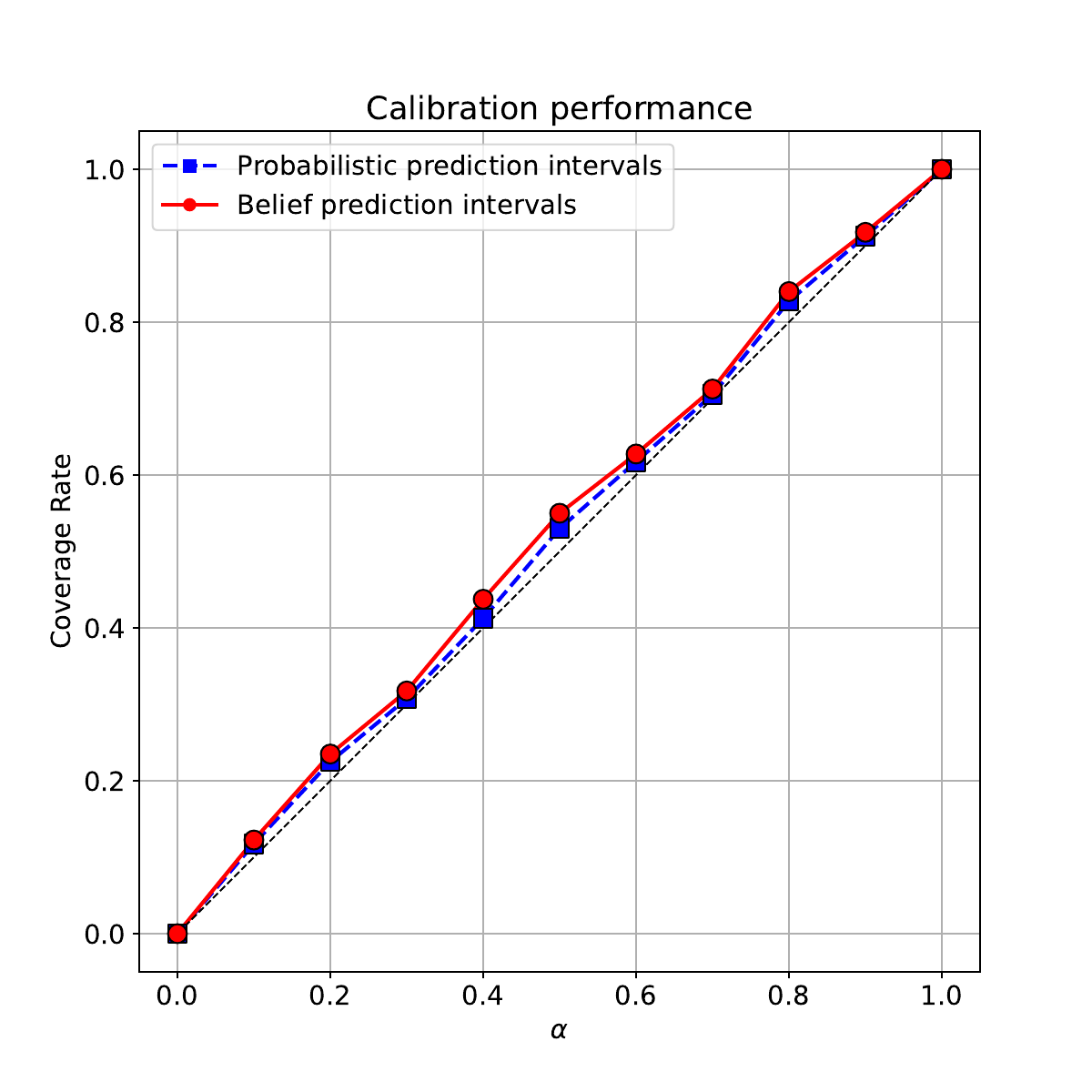}}
%\subfloat[\label{fig: v2_0.2_c}]{\includegraphics[width=0.5\textwidth]{Figures/evreg_v2/2000_0.2/calibration.pdf}}\\
\subfloat[\label{fig: v2_0.5_c}50\% censoring]{\includegraphics[width=0.4\textwidth]{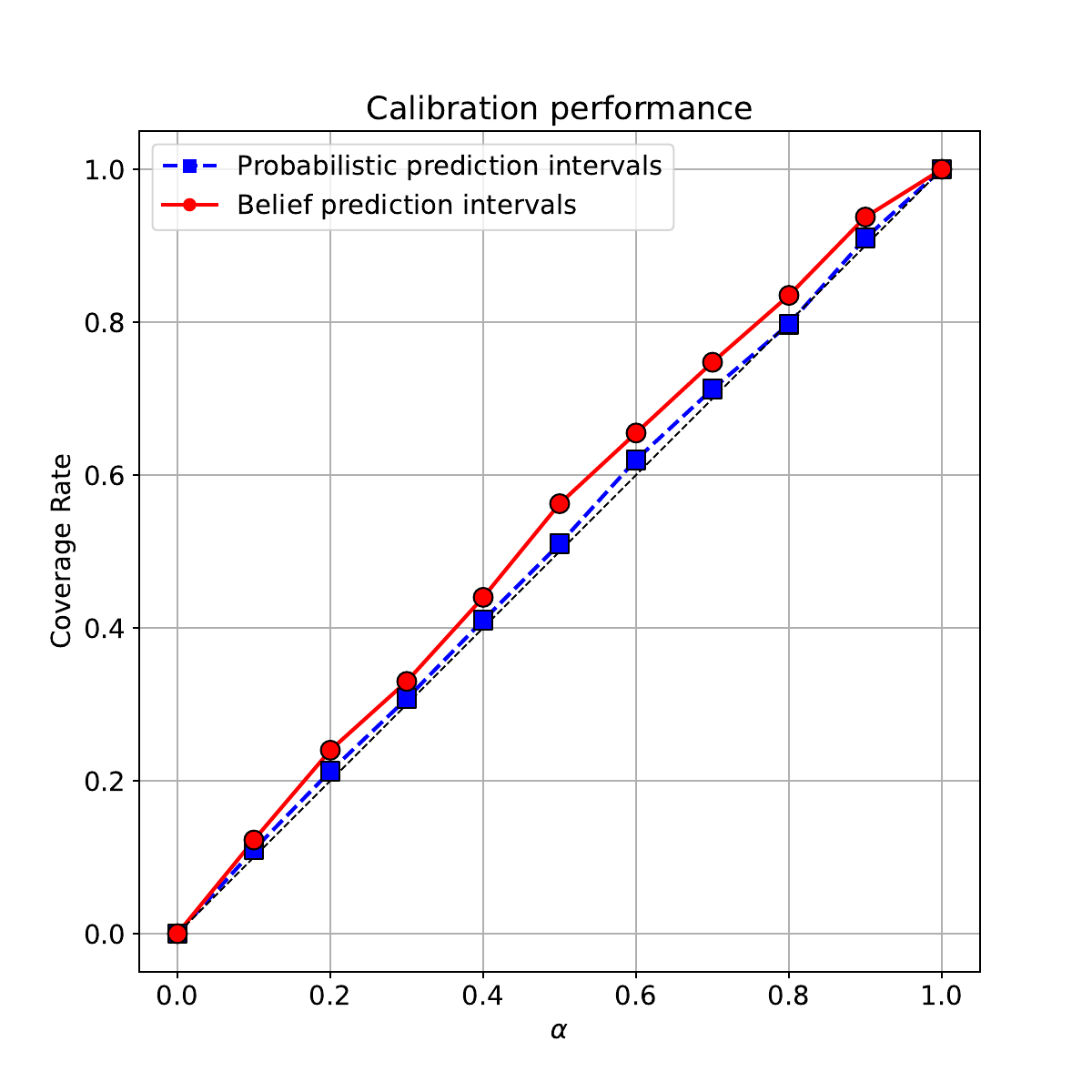}}
%\subfloat[\label{fig: v2_0.9_c}90\% censoring]{\includegraphics[width=0.33\textwidth]{Figures/illustrative_example/0.9/calibration.pdf}}\\
%\vspace{1cm}
%\colorbox{gray!20}{\parbox{0.9\textwidth}{\centering \textbf{Reliability diagrams after calibration}}}\\
%\vspace{-0.2cm}
%\subfloat[\label{fig: v2_0_ca}No censoring]{\includegraphics[width=0.33\textwidth]{Figures/illustrative_example/0/calibration2.pdf}}
%\subfloat[\label{fig: v2_0.5_ca}50\% censoring]{\includegraphics[width=0.33\textwidth]{Figures/illustrative_example/0.5/calibration2.pdf}}
%\subfloat[\label{fig: v2_0.9_ca}90\% censoring]{\includegraphics[width=0.33\textwidth]{Figures/illustrative_example/0.9/calibration2.pdf}}
\caption{Calibration plots for the illustrative dataset: belief prediction intervals (red solid lines) and probabilistic prediction intervals (blue broken lines).}
\label{fig: calibration}
\end{figure}

Figure \ref{fig:pred_int} displays the transformed observations $y=\log t$ and the network outputs $\mu(x)$ along with BPIs at levels $\alpha \in \{0.5, 0.9, 0.99\}$, with  0\% and 50\% censoring rates. Without censoring, our model prediction $\mu(x)$ (red line)  is close to the ground truth function (blue broken line), and the predicted BPIs effectively encompass all of the data points, as shown in Figure \ref{fig: v2_0}. 
When 50\% of the observations are censored (Figure \ref{fig: v2_0.5}), the predictions become slightly biased and tend to overestimate the time to event. However, the 90\% and 99\% belief intervals still contain most of the data. This is confirmed by Figure \ref{fig: calibration} showing calibration plots with and without data censoring. Without censoring (Figure \ref{fig: v2_0_c}), we can see that both prediction intervals are well-calibrated, and the calibration curve of BPIs is close to that of probabilistic prediction intervals, indicating that the output precision $h(x)$ is quite high (which is confirmed by Figure \ref{fig: v2_0}). With 50\% censoring rate (Figure \ref{fig: v2_0.5_c}),  the BPIs are more conservative than the probabilistic ones, as a result of lower output precision $h(x)$ (as seen in Figure \ref{fig: v2_0.5}) due to higher epistemic uncertainty. 

\subsection{Comparative results on simulated datasets}

\label{subsec: simulated}
The second experiment focuses on demonstrating the applicability and performance of our methods under different data distributions. 

\paragraph{Datasets} Following \cite{katzman2018deepsurv}, we generated patient death time $T$ as a function of a vector $x$ of 10 covariates using the exponential Cox model: 
\begin{equation*}
    T \sim \operatorname{Exp}(\lambda(t | x))=\operatorname{Exp}\left(\lambda_{0} \cdot e^{g(x)}\right).
\end{equation*}
The ten covariates were drawn independently from a uniform distribution in $[-1, 1]$. In all experiments, the log-risk function $g(x)$ only depends on two of the ten covariates. This allows us to verify that ENNreg discerns the relevant covariates from the noise. We then chose a censoring time to represent the “end of study” such that 50\% of the patients in the dataset have an observed event, $d = 1$. Three data distribution situations, i.e., linear proportional hazards (LPH), nonlinear (Gaussian) proportional hazards (NLPH), and nonlinear nonproportional hazards (NLNPH) were considered: 
\bd
\item[LPH:] Following \cite{katzman2018deepsurv}, we assumed a linear $g(x)$ so that the linear PH assumption holds: 
\begin{equation*}
g(x) = x_{0} + 2x_{1}.
\end{equation*}

\item[NLPH:] We assumed $g(x)$ to be Gaussian with $\lambda_{\max} = 5.0$ and scale factor $r = 0.5$ \cite{katzman2018deepsurv}:
\begin{equation*}
g(x)=\log \left(\lambda_{\max}\right) \exp \left(-\frac{x_{0}^{2}+x_{1}^{2}}{2 r^{2}}\right).
\end{equation*}

\item[NLNPH:] Following \cite{kvamme2019time}, we constructed a dataset with nonlinear nonproportional hazard function 
\begin{equation*}
    \lambda(t | x)=\lambda_{0}(t) \exp [g(t,x)],
\end{equation*} 
where $g(t,x)$ is a time-dependent function and $\lambda_{0}$ is set to 0.02. More details about the definition of $g(t,x)$ can be found in \cite{kvamme2019time}.
\ed
For the LPH and NLPH datasets, 5,000 samples were generated. For the more complex NLNPH dataset, 25,000 samples were generated.

\paragraph{Settings}  Each model was trained on the learning set with 500 epochs using the Adam optimization algorithm. The initial learning rate was set to $0.1$. An adjusted learning rate schedule was applied by reducing the learning rate when the training loss did not decrease in 100 epochs. An early stopping mechanism was applied when the validation performance did not improve in 20 training epochs. The model with the best performance on the validation set was saved as the final model for testing. For ENNreg, the number of prototypes was fixed to $K = 40$ for all datasets. Hyperparameter $\epsilon$ was fixed to $1e^{-4} \sigma_{Y}$, where $\sigma_{Y}$ is the estimated standard deviation of the response variable for all the simulations; hyperparameters $\xi$ and $\rho$ were fixed to 0. Hyperparameter $\lambda$ was set to  0.1.  Each dataset was randomly split into training, validation, and test sets containing, respectively, 60\%, 20\%, and 20\% of the observations. These random splits were repeated five times. Finally, the mean prediction and the 95\% confidence interval were calculated and reported. All methods were implemented in Python and were trained and tested on a 2023 Apple M2 Pro with a 12-core CPU and 19-core GPU, 32 GB CPU memory.

\begin{figure}
\centering
\subfloat[\label{fig: c_idx_LPH}$C_{idx}$ under LPH]{\includegraphics[width=0.33\textwidth]{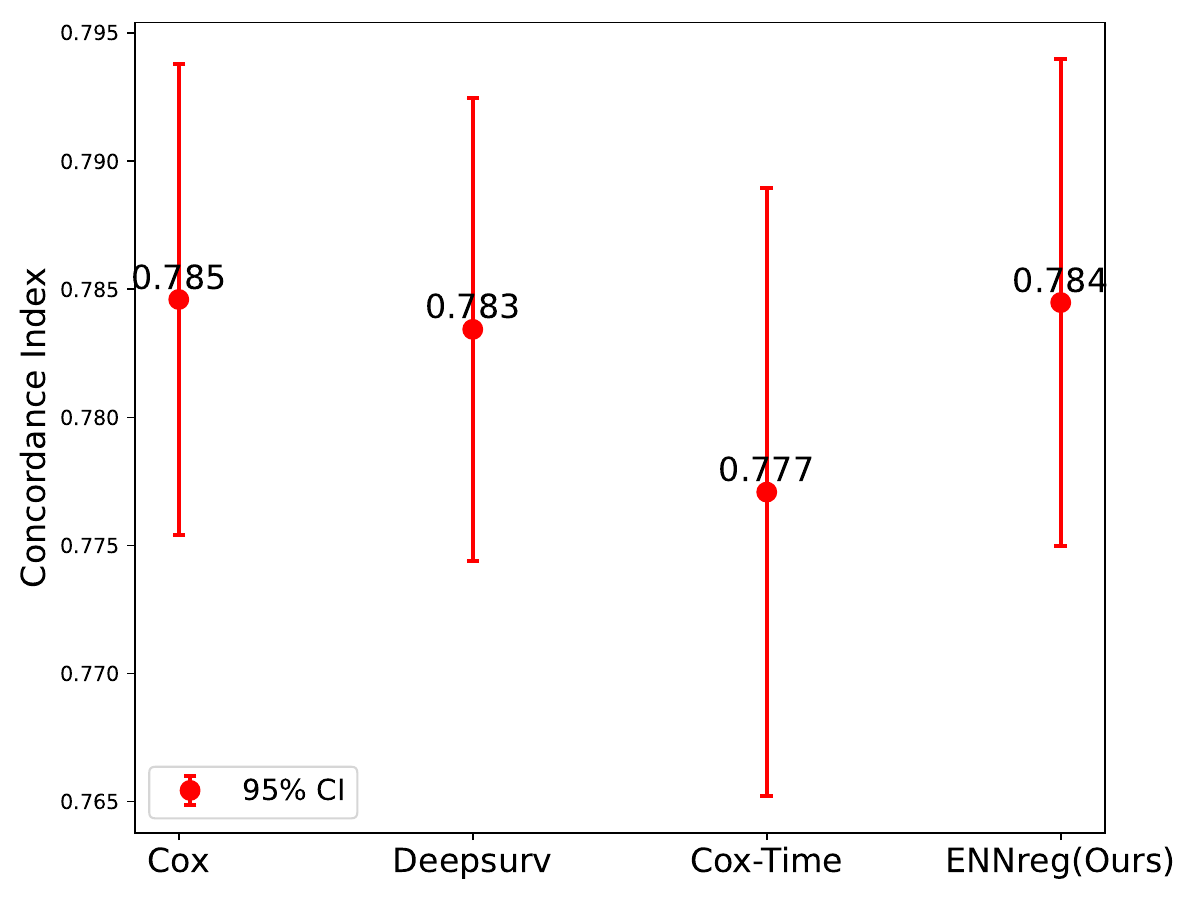}}
\subfloat[\label{fig: c_idx_NLPH}$C_{idx}$ under NLPH]{\includegraphics[width=0.33\textwidth]{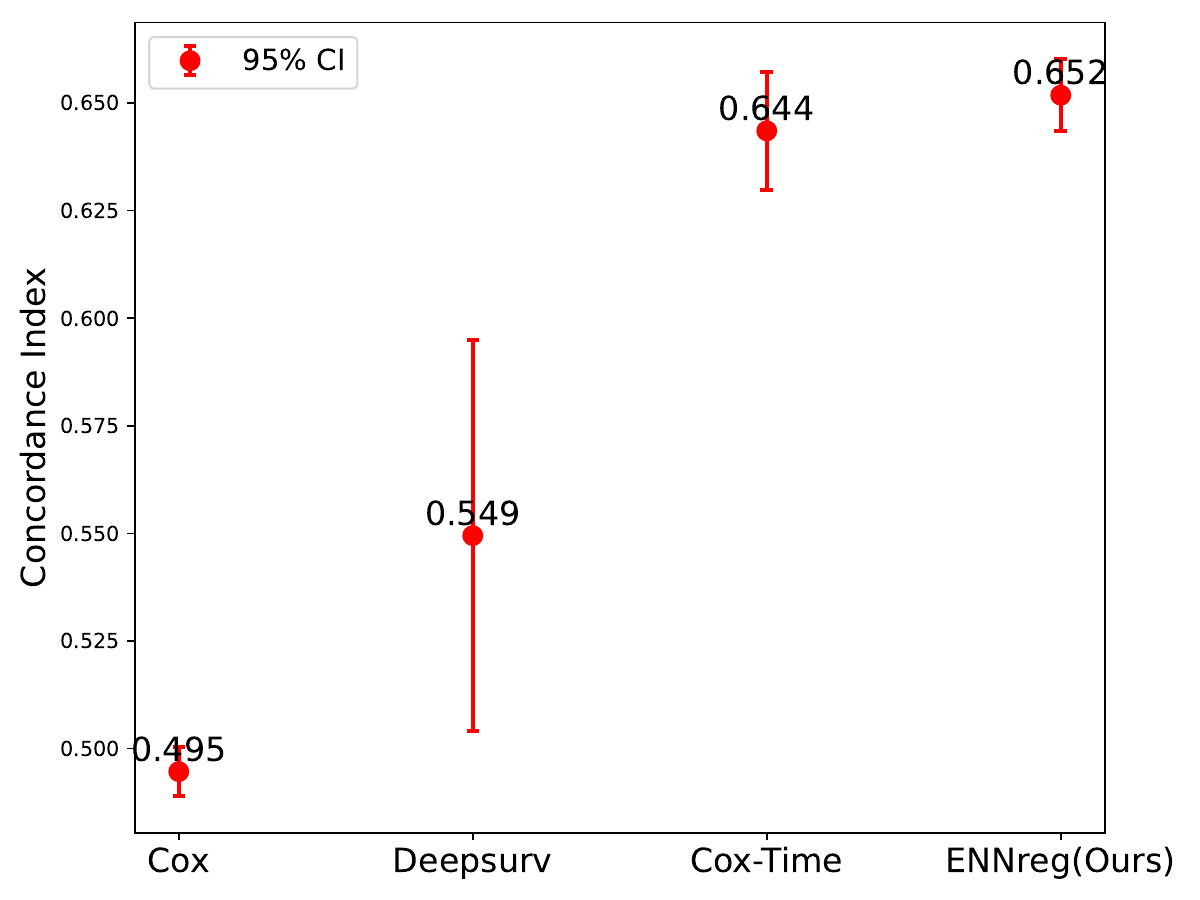}}
\subfloat[\label{fig: c_idx_NLNPH}$C_{idx}$ under NLNPH]{\includegraphics[width=0.33\textwidth]{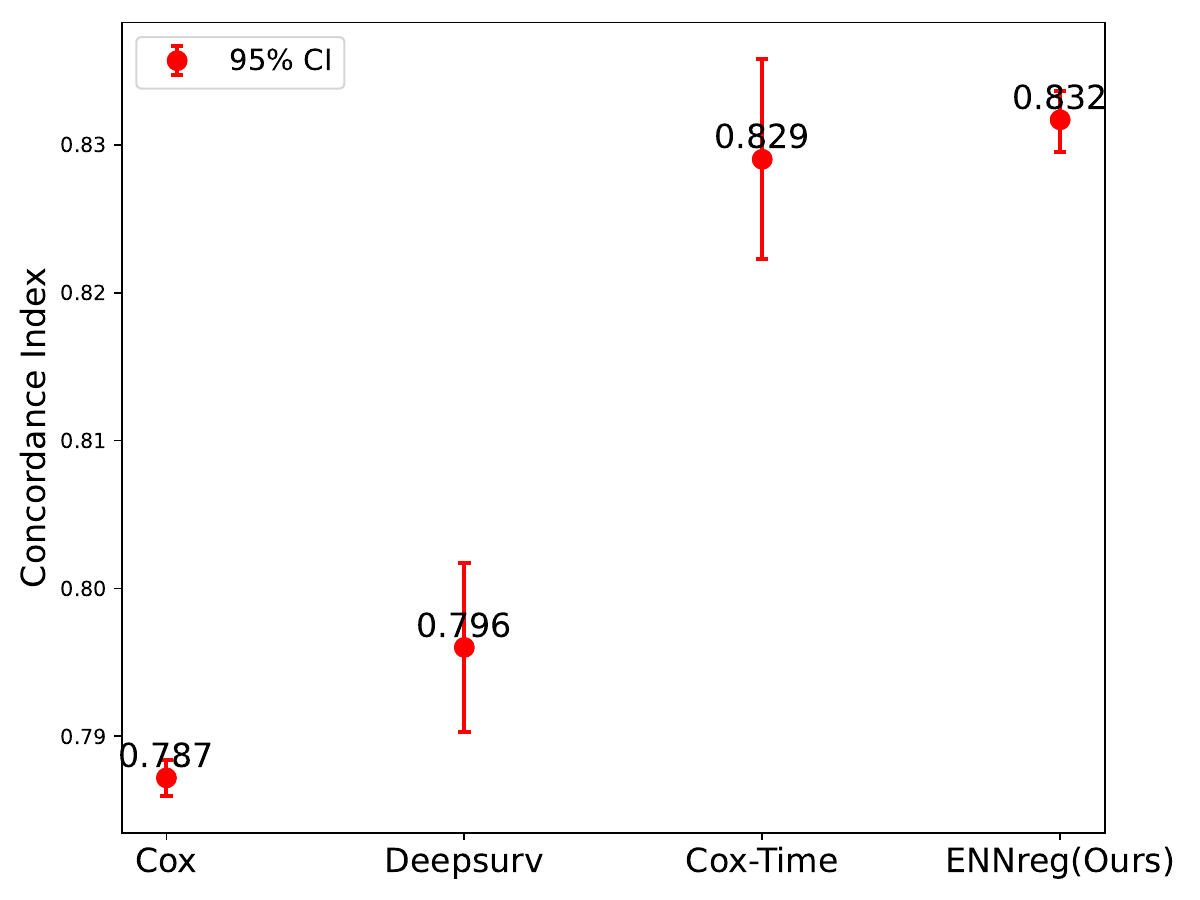}}\\

\subfloat[\label{fig: IBS_LPH}$IBS$ under LPH]{\includegraphics[width=0.33\textwidth]{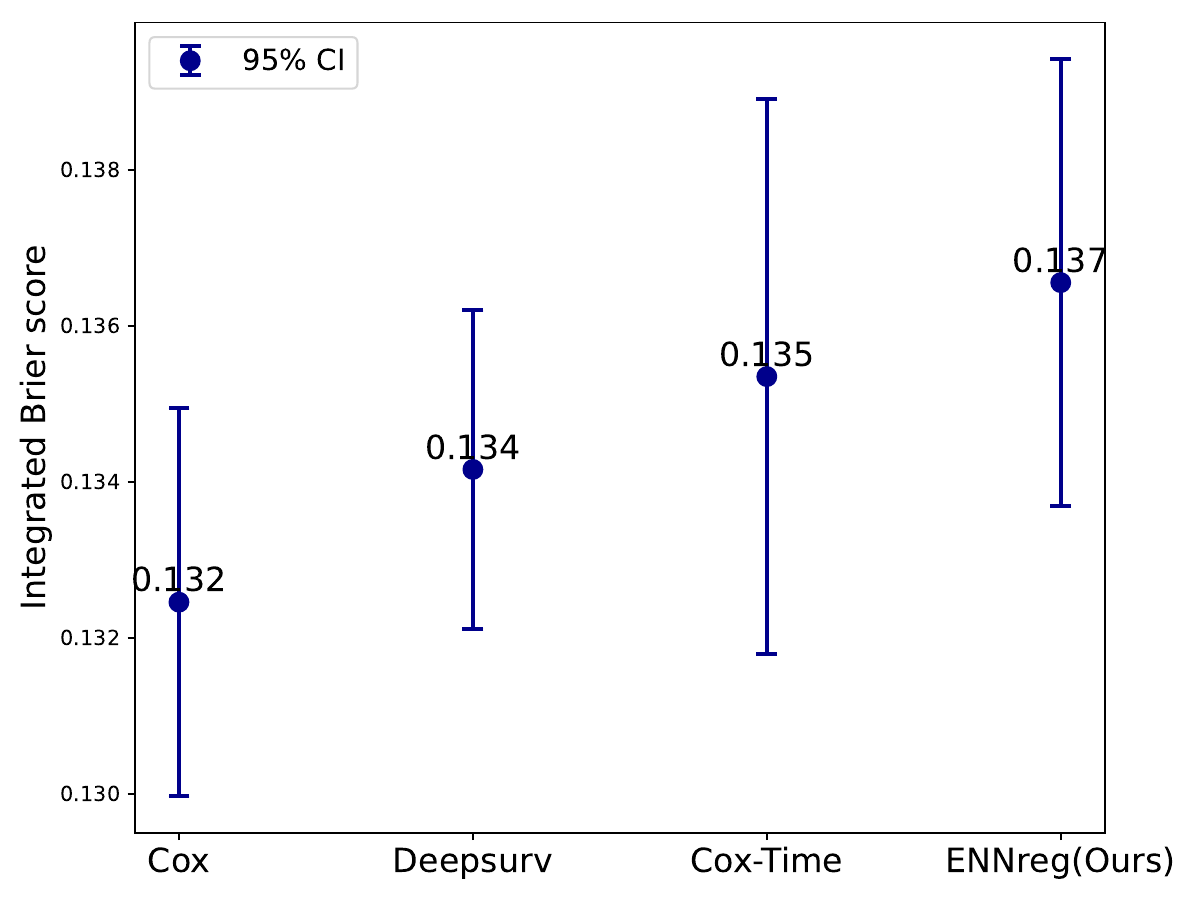}}
\subfloat[\label{fig: IBS_NLPH}$IBS$ under NLPH]{\includegraphics[width=0.33\textwidth]{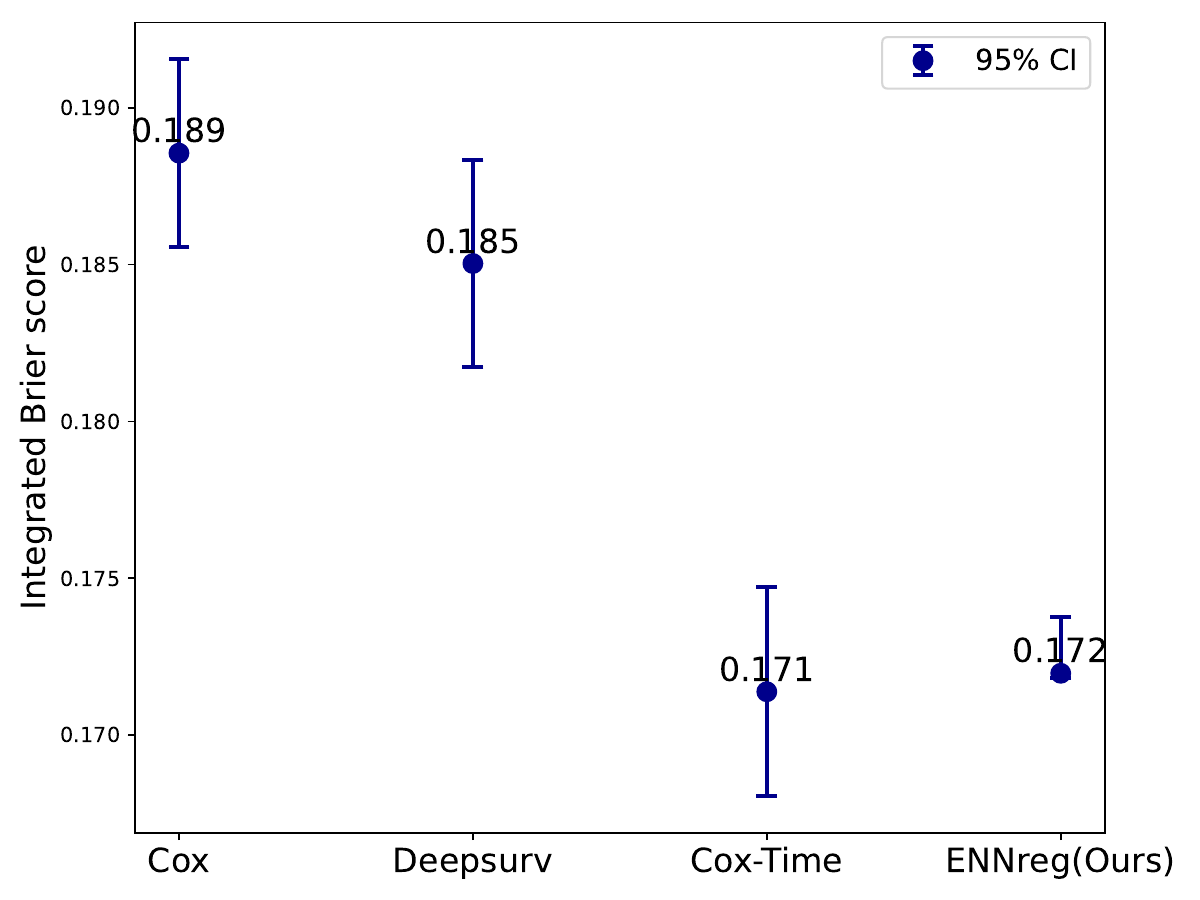}}
\subfloat[\label{fig: IBS_NLNPH}$IBS$ under NLNPH]{\includegraphics[width=0.33\textwidth]{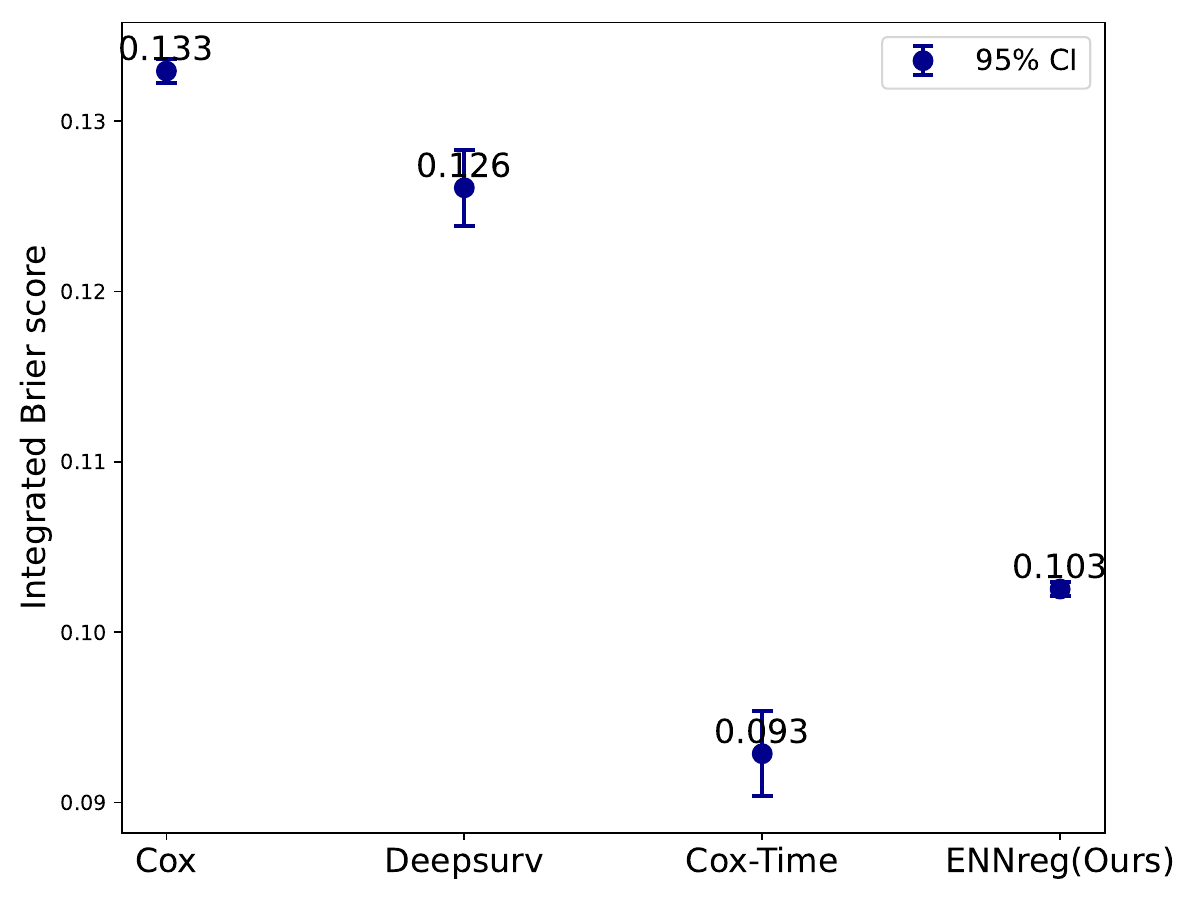}}\\

\subfloat[\label{fig: IBLL_LPH}$IBLL$ under LPH]{\includegraphics[width=0.33\textwidth]{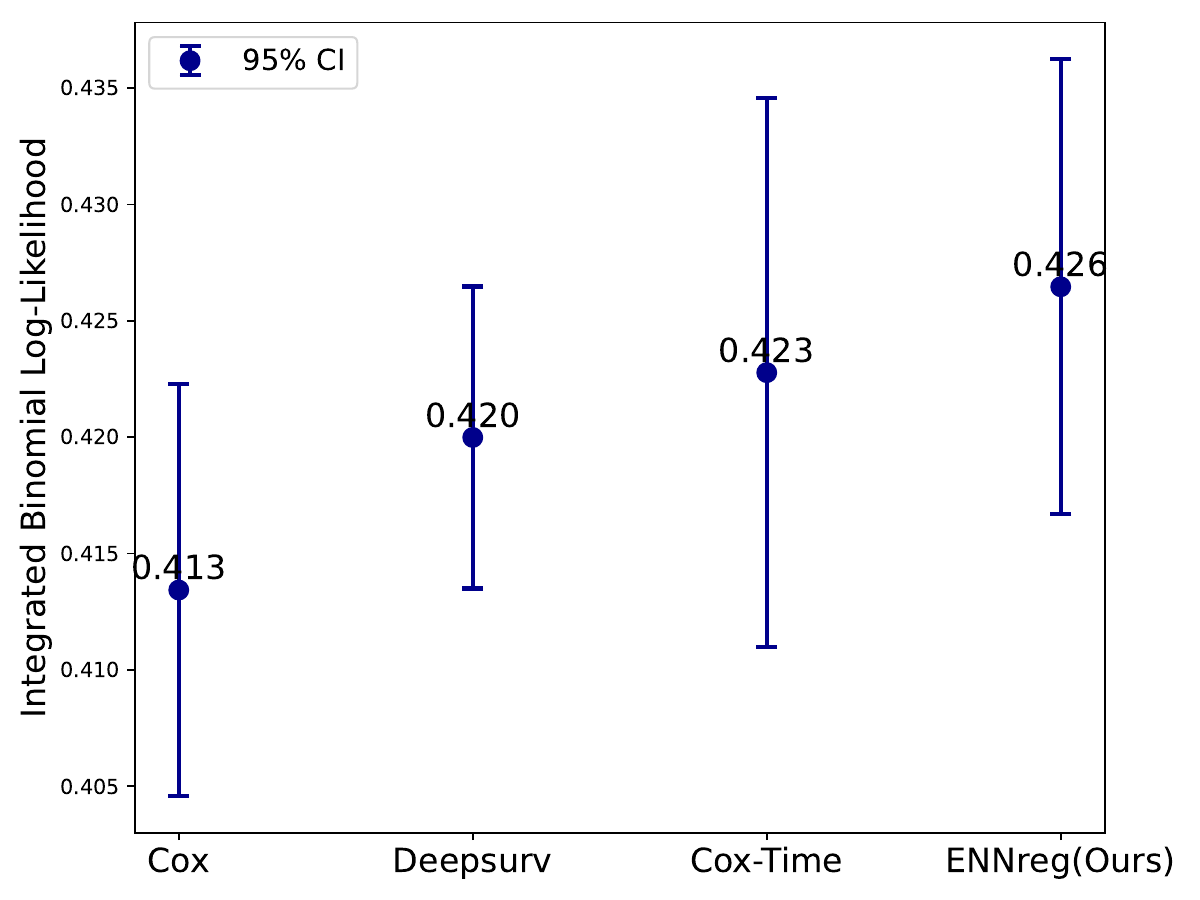}}
\subfloat[\label{fig: IBLL_NLPH}$IBLL$ under NLPH]{\includegraphics[width=0.33\textwidth]{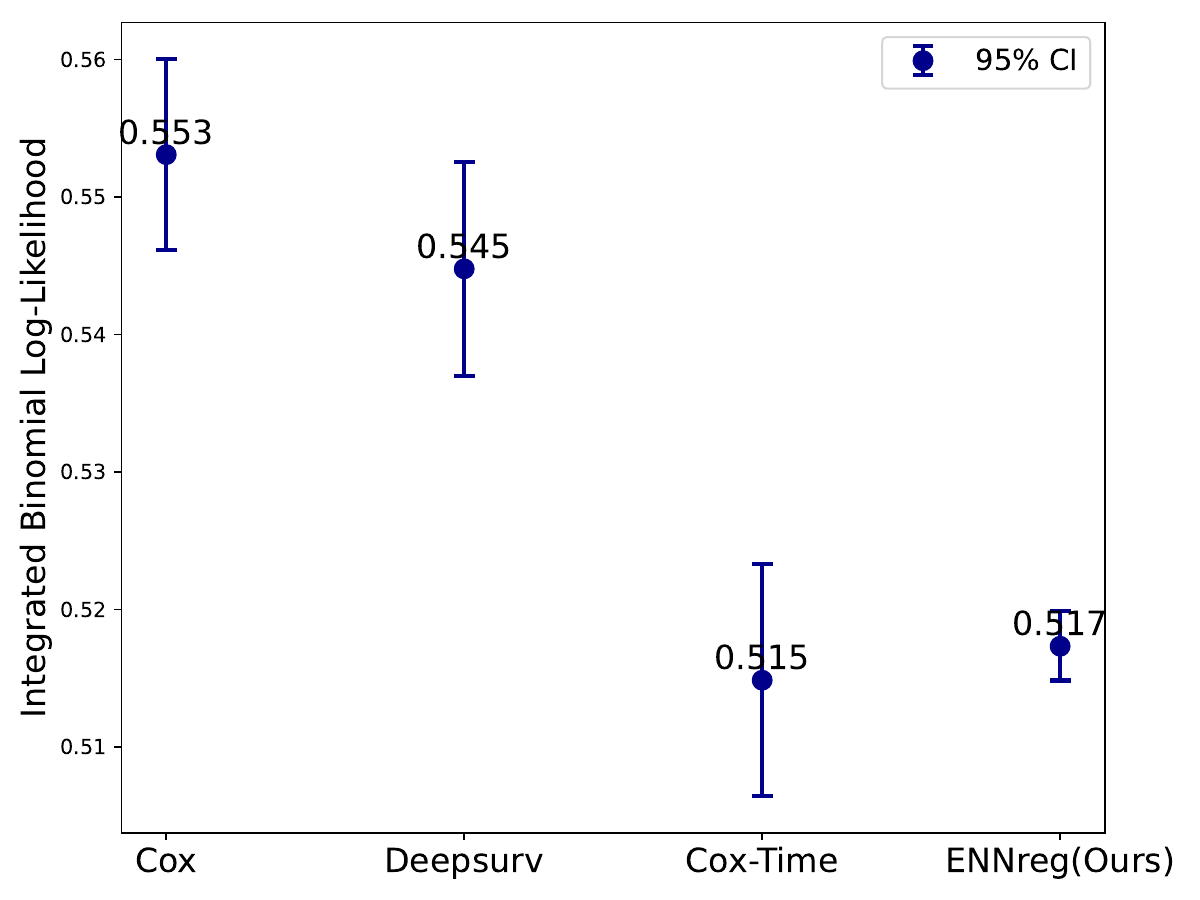}}
\subfloat[\label{fig: IBLL_NLNPH}$IBLL$ under NLNPH]{\includegraphics[width=0.33\textwidth]{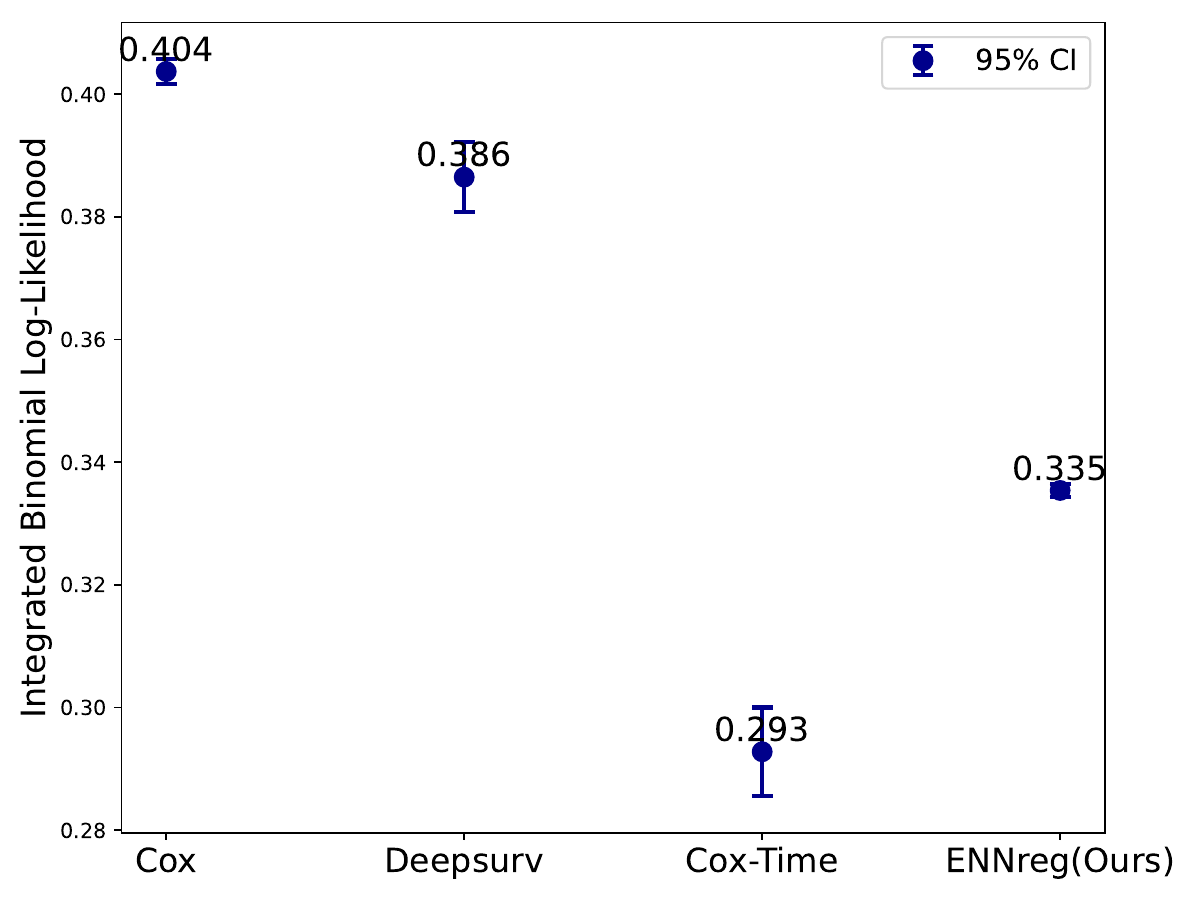}}
\\
\caption{Performance comparison on the simulated data. The left, middle, and right columns correspond to the prediction results of simulated datasets generated under the LPH, NLPH, and NLNPH assumptions, respectively. The four methods are, from left to right in each plot: Cox based on the LPH assumption, DeepSurv based on the NLPH assumption, Cox-Time based on the NLNPH assumption, and our ENNreg model. Larger values of the concordance index $C_{idx}$ (top row) and lower values of integrated Brier score $IBS$ (middle row) as well as integrated binomial log-likelihood $IBLL$ (bottom row) indicate better performance.}
\label{fig: risk_ass}
\end{figure}

\begin{figure}
\centering
\subfloat[\label{fig: cal_LPH} LPH]{\includegraphics[width=0.33\textwidth]{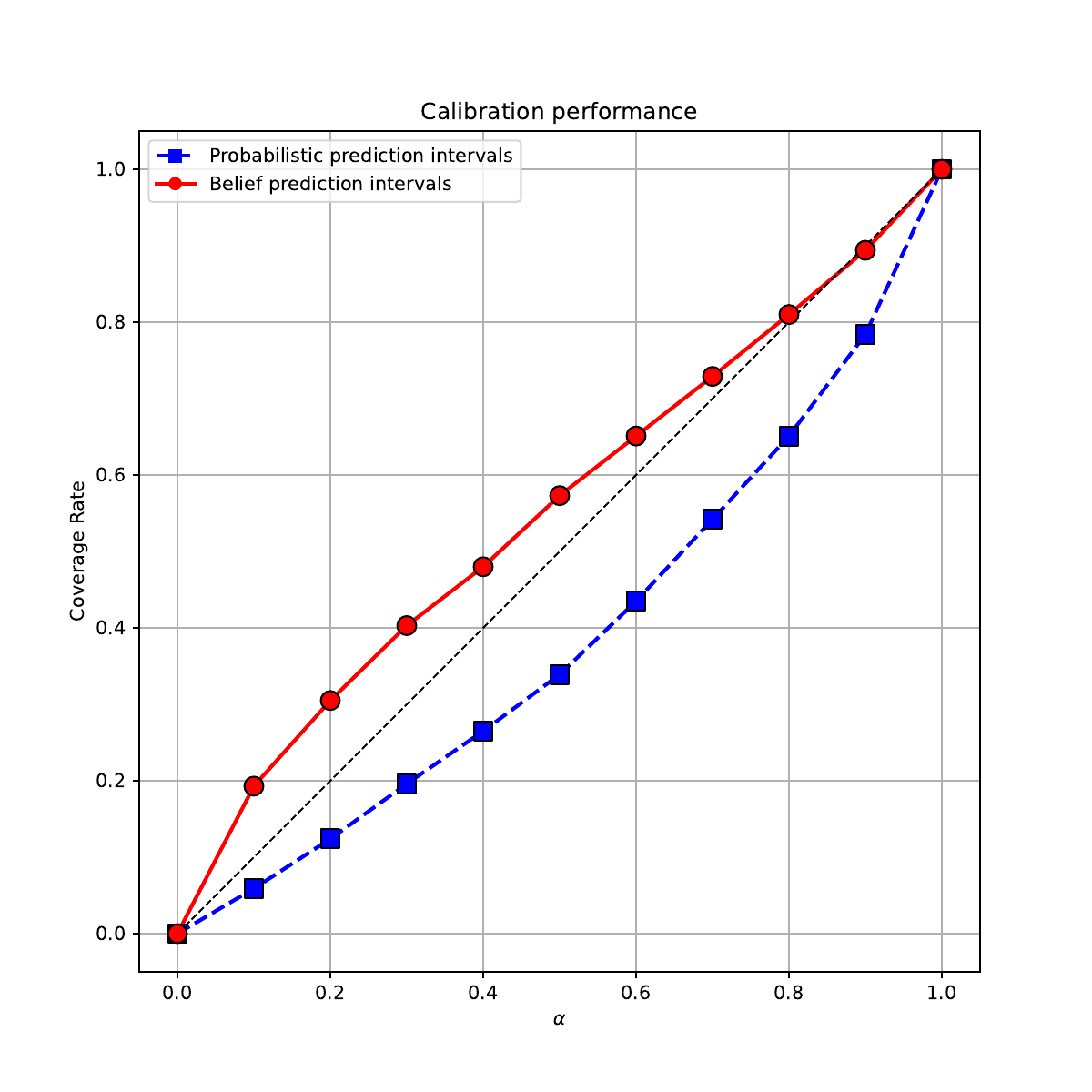}}
\subfloat[\label{fig: cal_NLPH} LPH]{\includegraphics[width=0.33\textwidth]{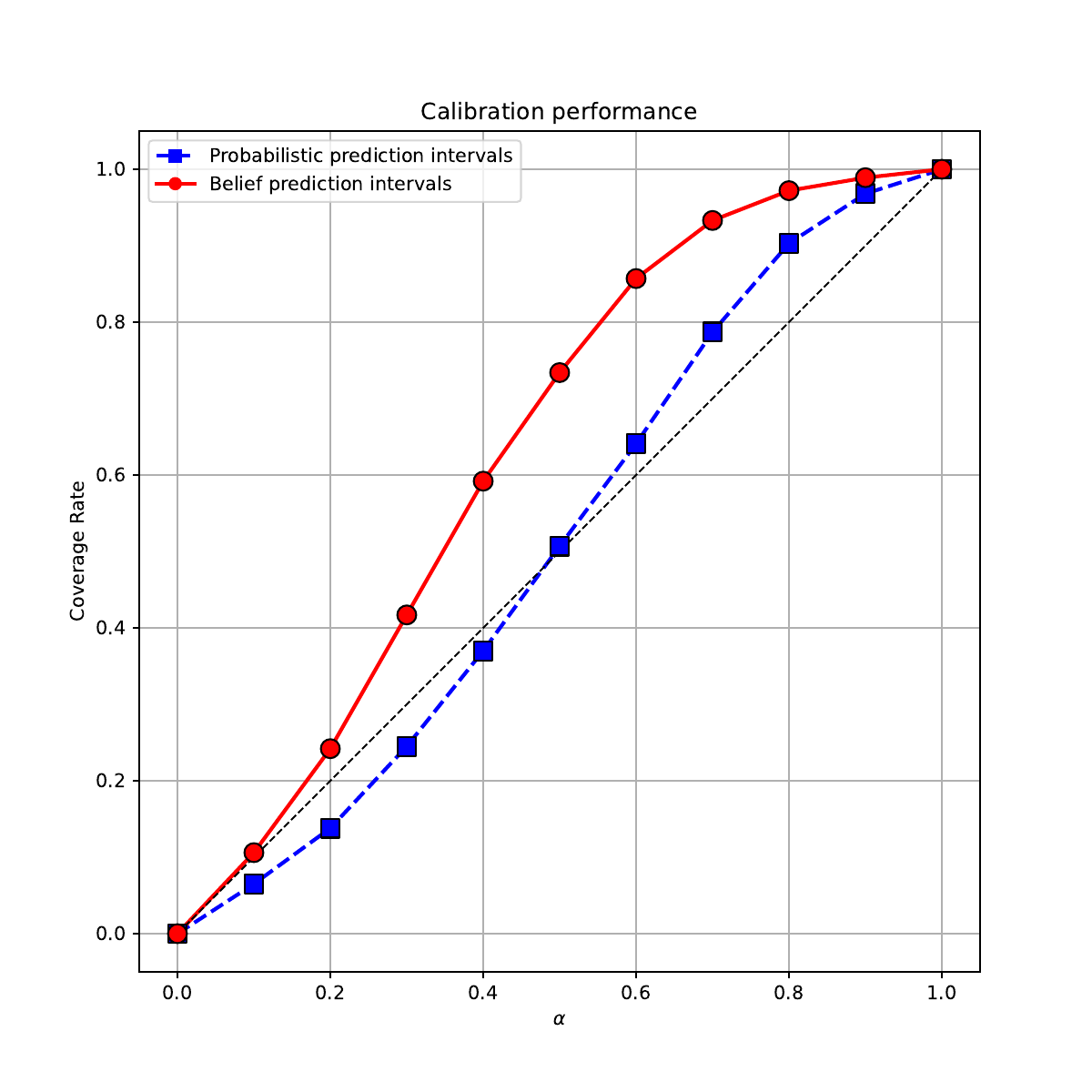}}
\subfloat[\label{fig: cal_NLNPH} NLNPH]{\includegraphics[width=0.33\textwidth]{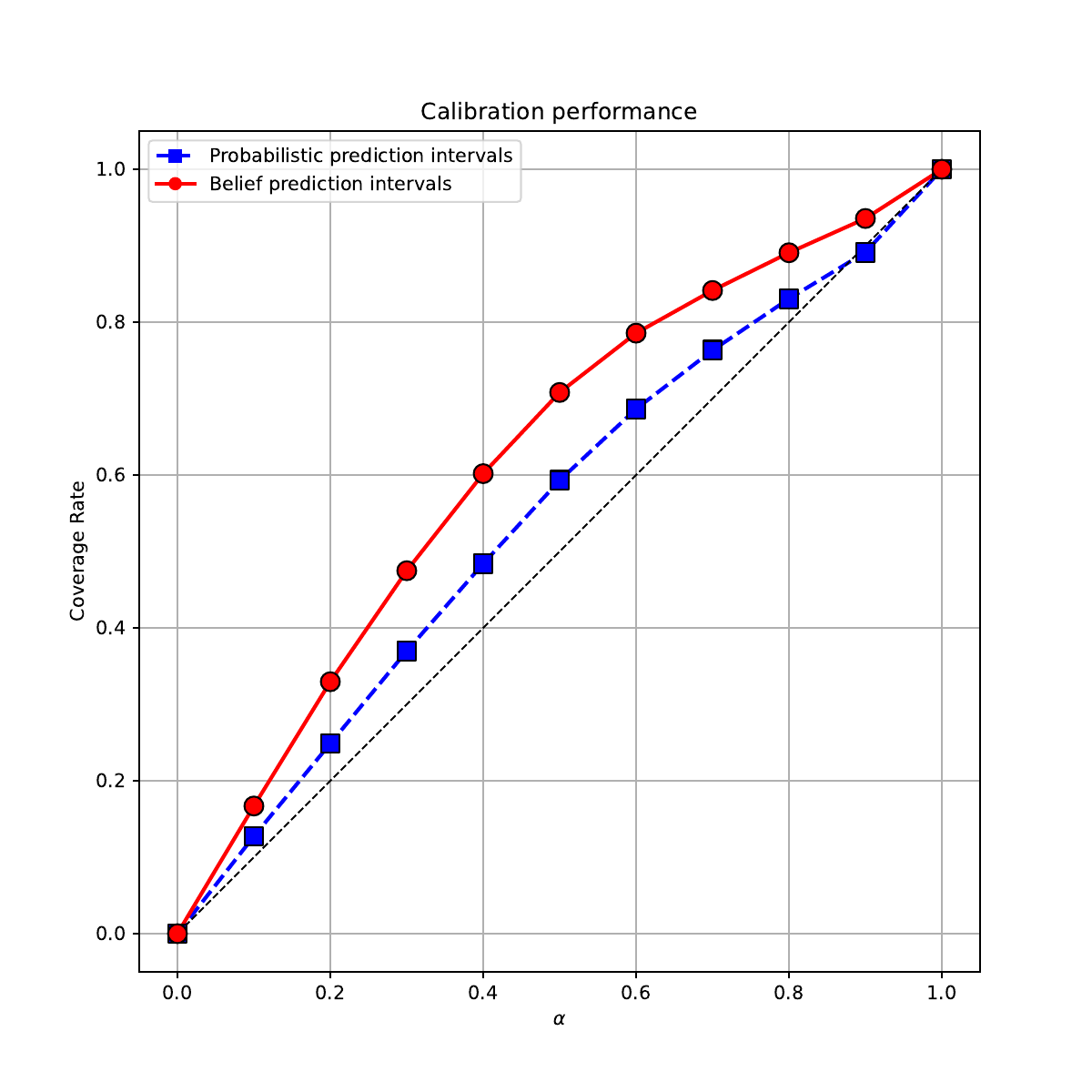}}
\caption{Calibration plots for the simulated data: belief prediction intervals (red solid lines) and probabilistic prediction intervals (blue broken lines).  \label{fig:cal_simul}}
\end{figure}

\paragraph{Quantitative results} Figure \ref{fig: risk_ass} presents the time-dependent concordance index ($C_{idx}$) \eqref{eq:Cidx} and integrated Brier score ($IBS$) \eqref{eq:IBS} for the three simulated datasets. We report results with our method as well as three existing approaches: Cox regression (referred to as Cox) \cite{cox1972regression}, DeepSurv \cite{katzman2018deepsurv}, an extension of the Cox model with an MLP module, and Cox-Time \cite{kvamme2019time}, which extends the Cox model with time-dependent hazard assumption and an MLP module. As expected, all methods perform well when based on true assumptions. For instance, the Cox model has the best performance according to the three criteria when the simulated data follows exactly the LPH condition (Figures \ref{fig: c_idx_LPH}, \ref{fig: IBS_LPH} and \ref{fig: IBLL_LPH}). In contrast, the DeepSurv and Cox-Time models outperform the Cox model on non linear data NLPH (Figures \ref{fig: c_idx_NLPH}, \ref{fig: IBS_NLPH} and \ref{fig: IBLL_NLPH}), and the performance gap between Cox-time and DeepSurv is even bigger on the NLNPH dataset (Figures \ref{fig: c_idx_NLNPH}, \ref{fig: IBS_NLNPH} and \ref{fig: IBLL_NLNPH}). Our model performs uniformly well, reaching the best or second best results according to both performance criteria. Calibration plots, shown in Figure \ref{fig:cal_simul}, show that the predictions are calibrated in the sense recalled in Section \ref{subsec: criteria}. These results demonstrate our model's ability to flexibly adapt to a wide range of underlying data distribution patterns.

\begin{figure}
\centering
\subfloat[\label{fig: LPH_base}True $S(t|x)$ of LPH]{\includegraphics[width=0.33\textwidth]{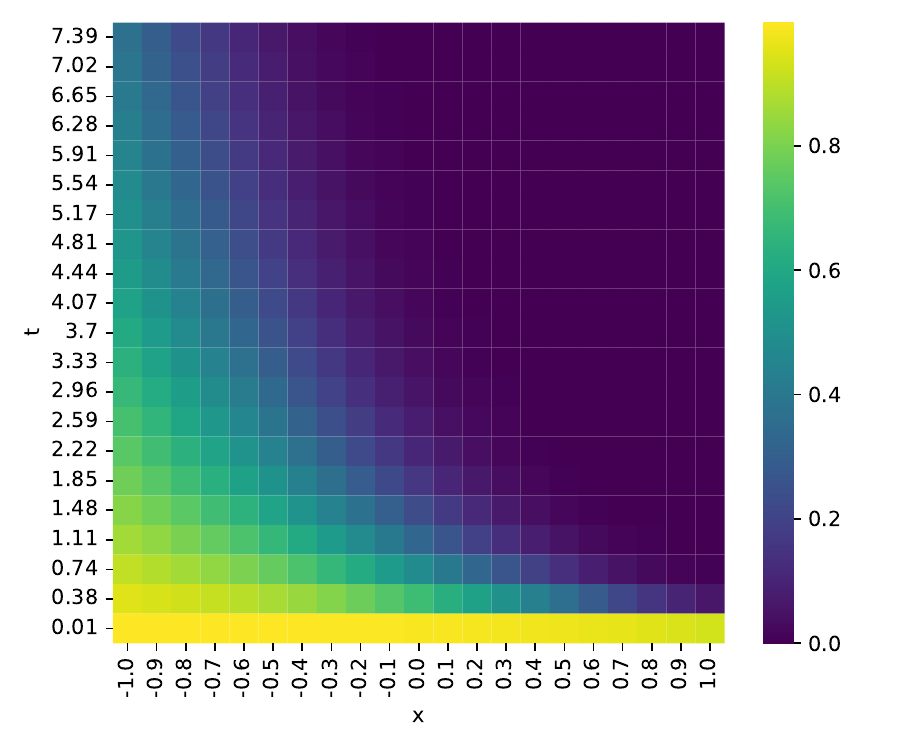}}
\subfloat[\label{fig: LPH_bel}Estimated $S_{*}(t|x)$ of LPH]{\includegraphics[width=0.33\textwidth]{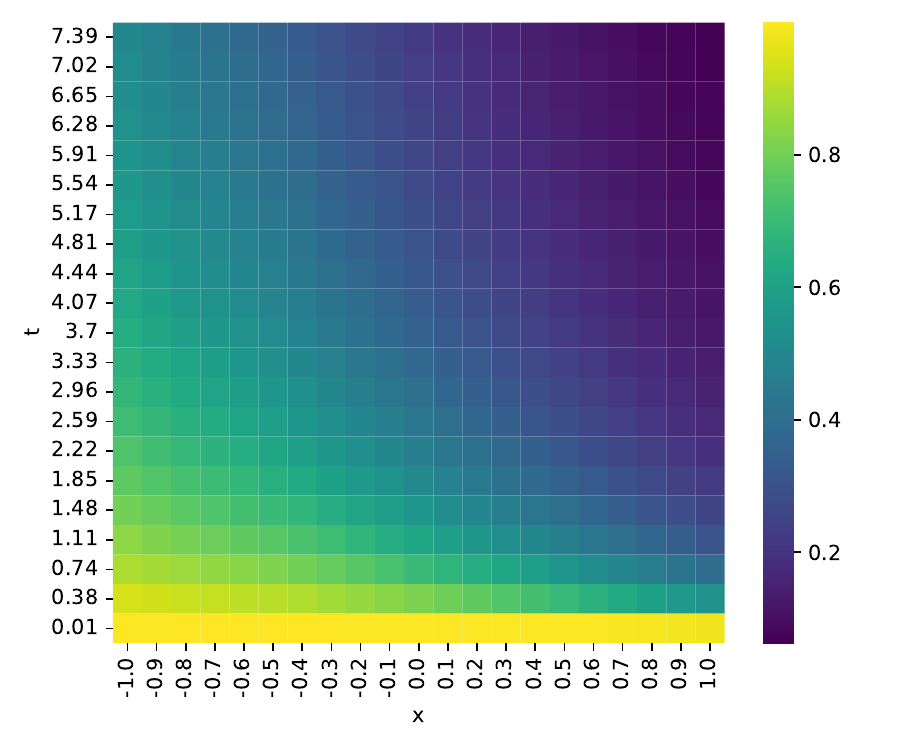}}
\subfloat[\label{fig: LPH_pl}Estimated $S^{*}(t|x)$ of LPH]{\includegraphics[width=0.33\textwidth]{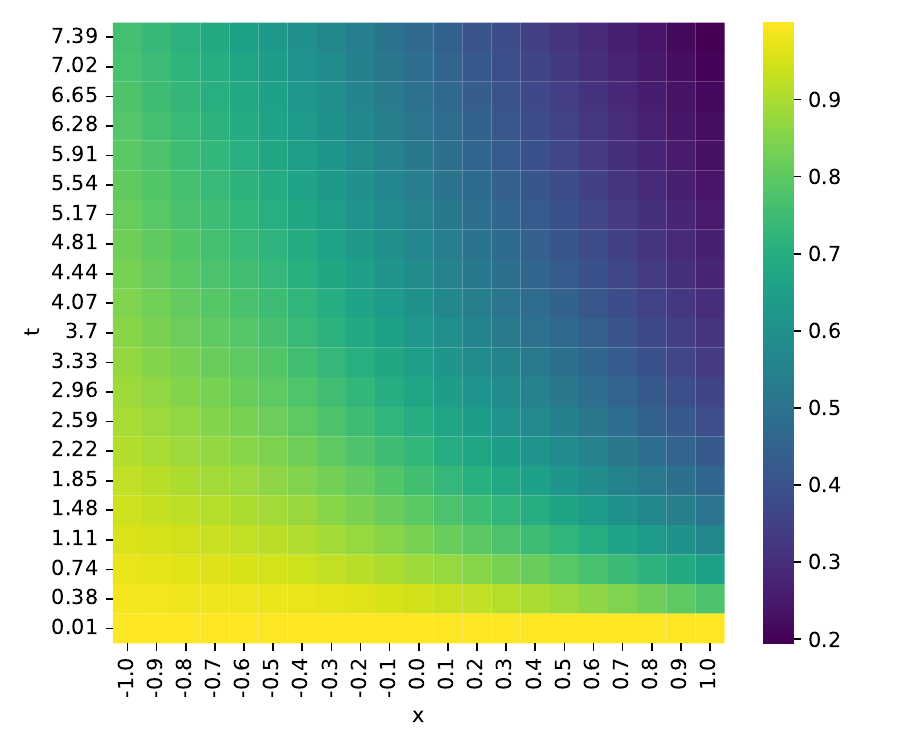}}
\\

\subfloat[\label{fig: NLPH_base}True $S(t|x)$ of NLPH]{\includegraphics[width=0.33\textwidth]{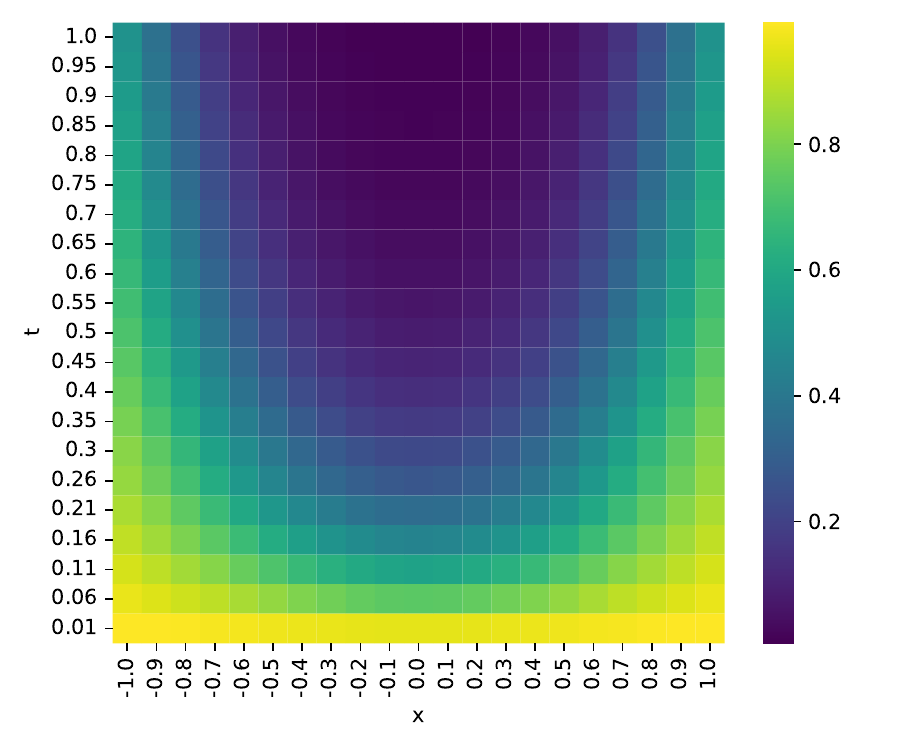}}
\subfloat[\label{fig: NLPH_bel}Estimated $S_{*}(t|x)$ of NLPH]{\includegraphics[width=0.33\textwidth]{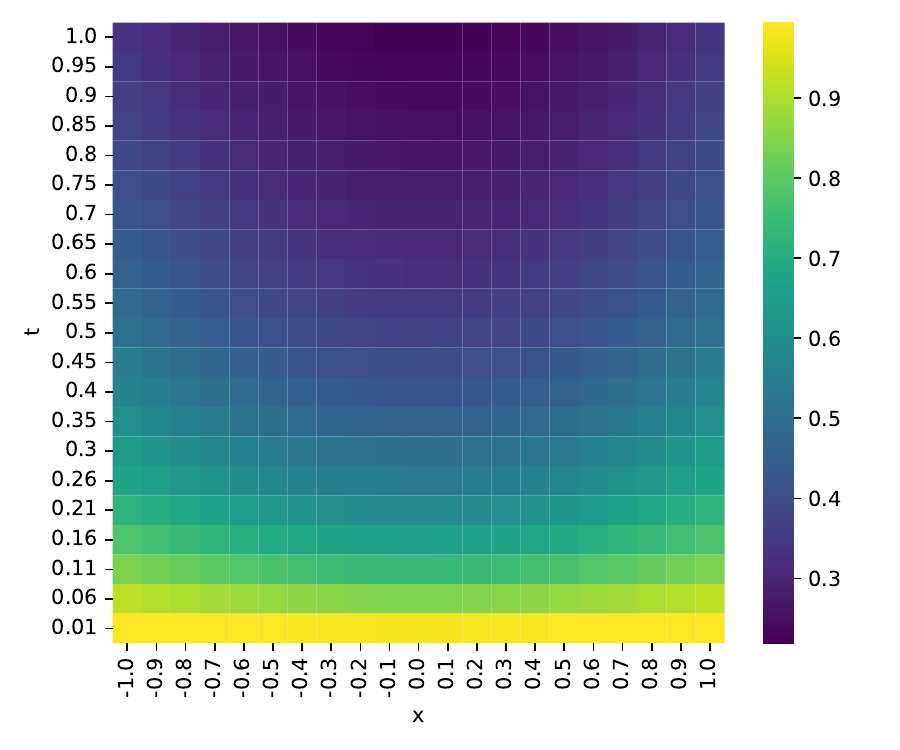}}
\subfloat[\label{fig: NLPH_pl}Estimated $S^{*}(t|x)$ of NLPH]{\includegraphics[width=0.33\textwidth]{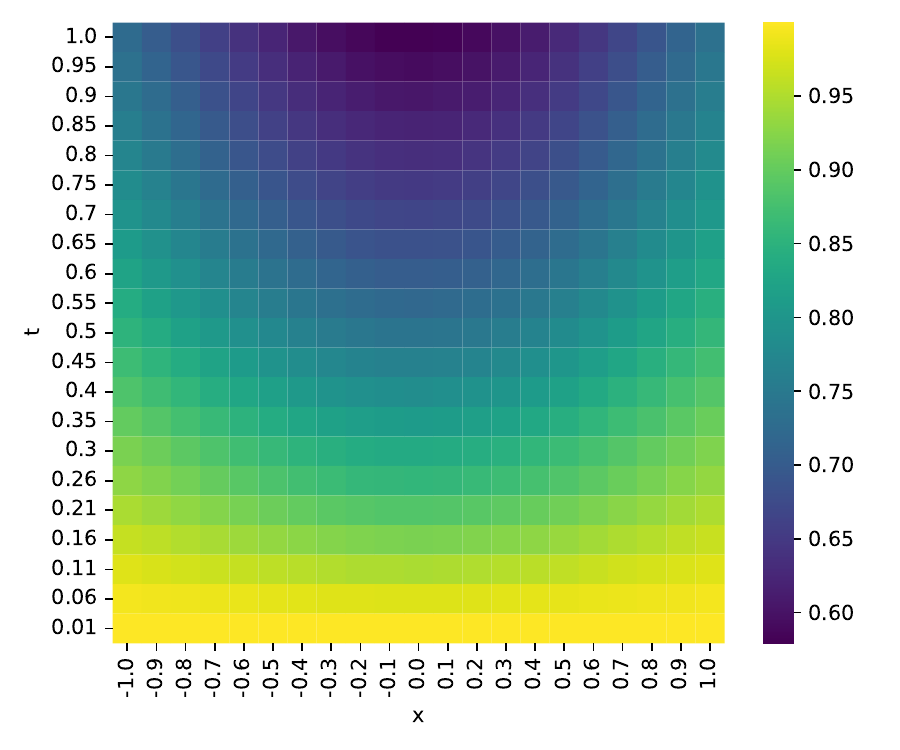}}\\

\subfloat[\label{fig: NLNPH_base}True $S(t|x)$ of NLNPH]{\includegraphics[width=0.33\textwidth]{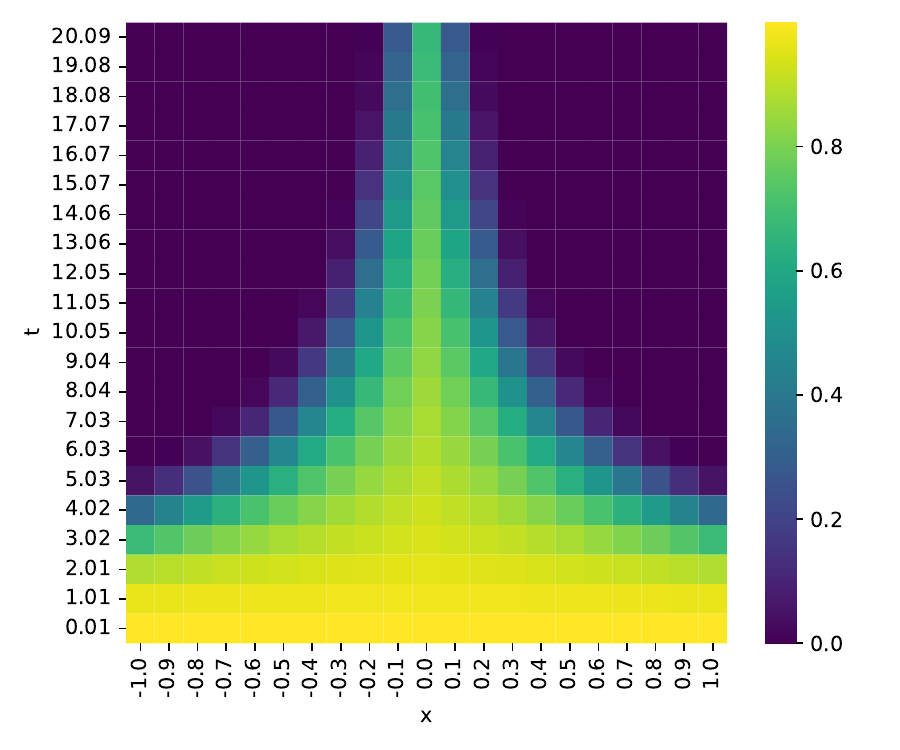}}
\subfloat[\label{fig: NLNPH_bel}Estimated $S_{*}(t|x)$ of NLNPH]{\includegraphics[width=0.33\textwidth]{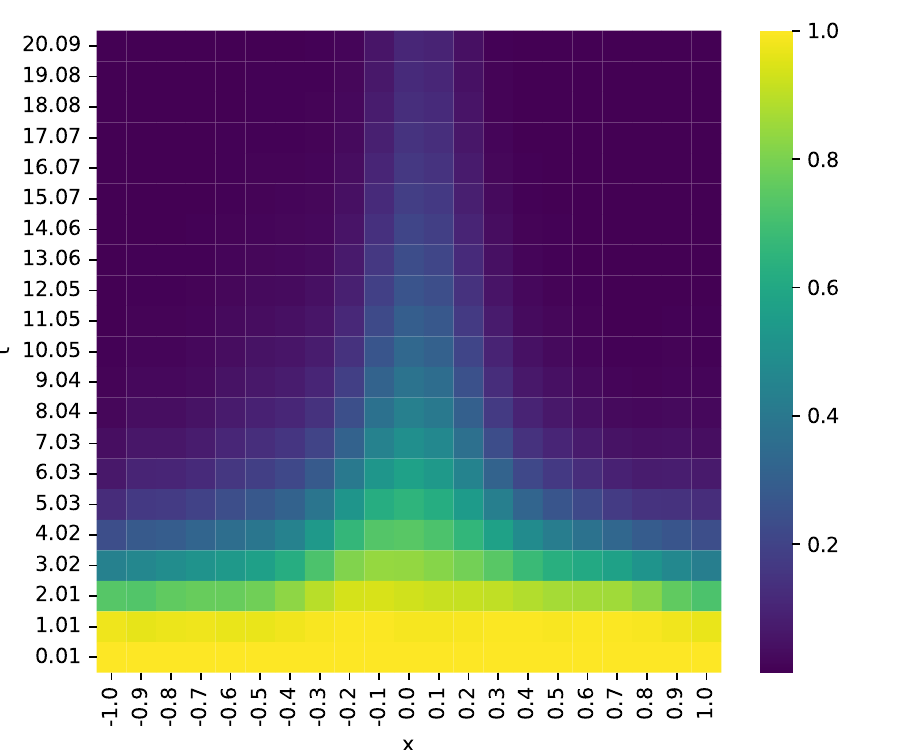}}
\subfloat[\label{fig: NLNPH_pl}Estimated $S^{*}(t|x)$ of NLNPH]{\includegraphics[width=0.33\textwidth]{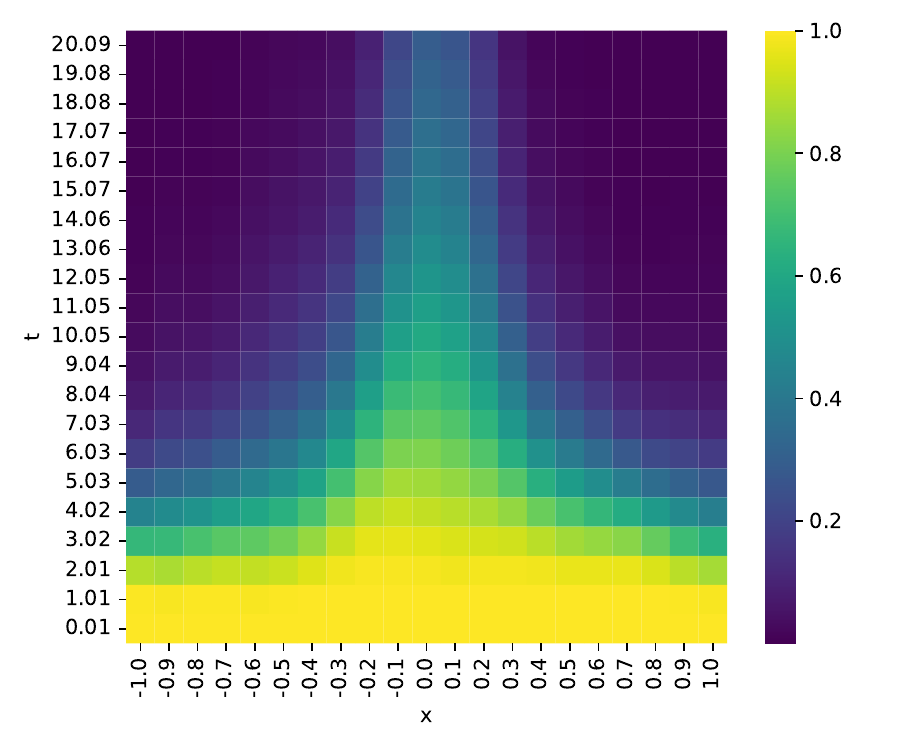}}
\\
\caption{Survival heat maps showing the conditional survival $S(t|x)$ for equally-spaced values of $x$ under the LPH, NLPH, and NLNPH assumptions (left columns). The middle and right columns display the estimated lower and upper survival function $S_{*}(t|x)$ and $S^{*}(t|x)$  obtained by ENNreg.}
\label{fig: heatmap}
\end{figure}

\paragraph{Prediction of survival functions} Figure \ref{fig: heatmap} shows true and survival heat maps generated under the LPH, NLPH, and NLPNL assumptions. Each map shows the true or estimated conditional survival function $S(t|x)$ for different values of one of the covariates $x$ (the other covariates being fixed), on a $21\times 21$ grid.  The first row (Figures \ref{fig: LPH_base}, \ref{fig: NLPH_base}, and \ref{fig: NLNPH_base}) displays the true survival functions under the LPH, NLPH, and NLNPH assumptions, respectively. The second and third rows show, respectively,  the predicted lower and upper survival functions \eqref{eq:lower_upper} computed by ENNreg.
We can see that ENNreg effectively captures both linear and nonlinear relationships between covariate $x$ and their survival function $y$. Notably, Figures \ref{fig: NLNPH_base}, \ref{fig: NLNPH_bel} and \ref{fig: NLNPH_pl} demonstrate the ability of ENNreg to model the nonlinear nonproportional conditions.

\begin{figure}
\centering
\subfloat[\label{fig: s(t)_LPH2_ennreg}]{\includegraphics[width=0.4\textwidth]{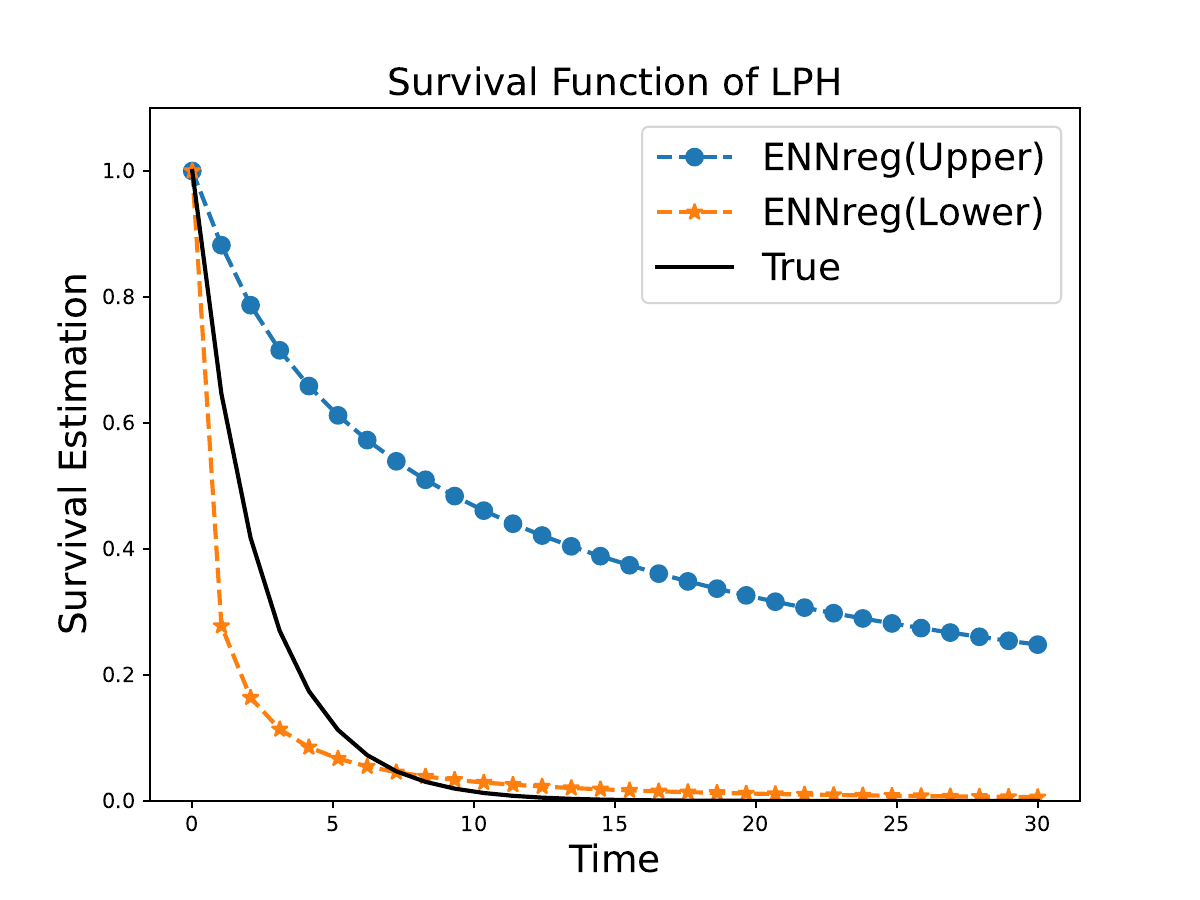}}
\subfloat[\label{fig: s(t)_LPH2_others}]{\includegraphics[width=0.4\textwidth]{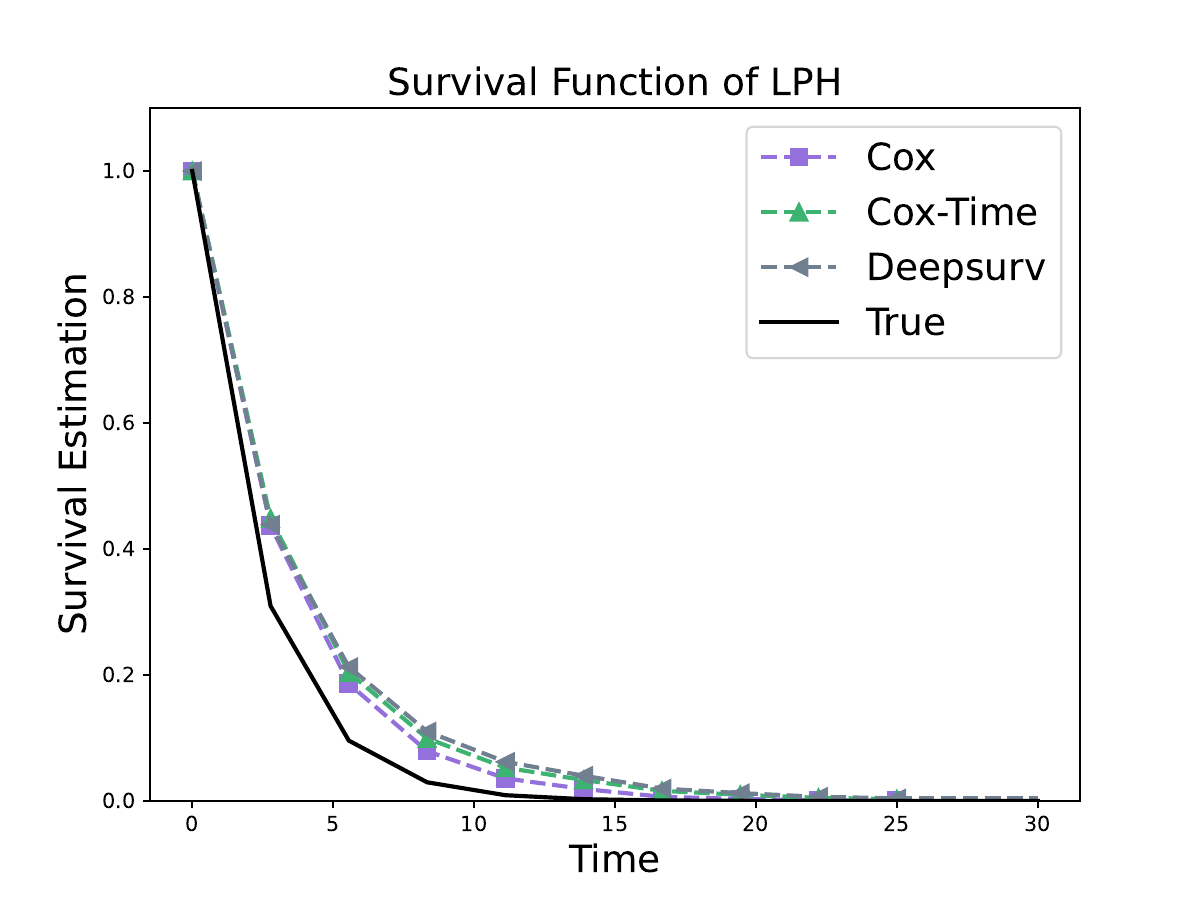}}\\
\subfloat[\label{fig: s(t)_LPH3_ennreg}]{\includegraphics[width=0.4\textwidth]{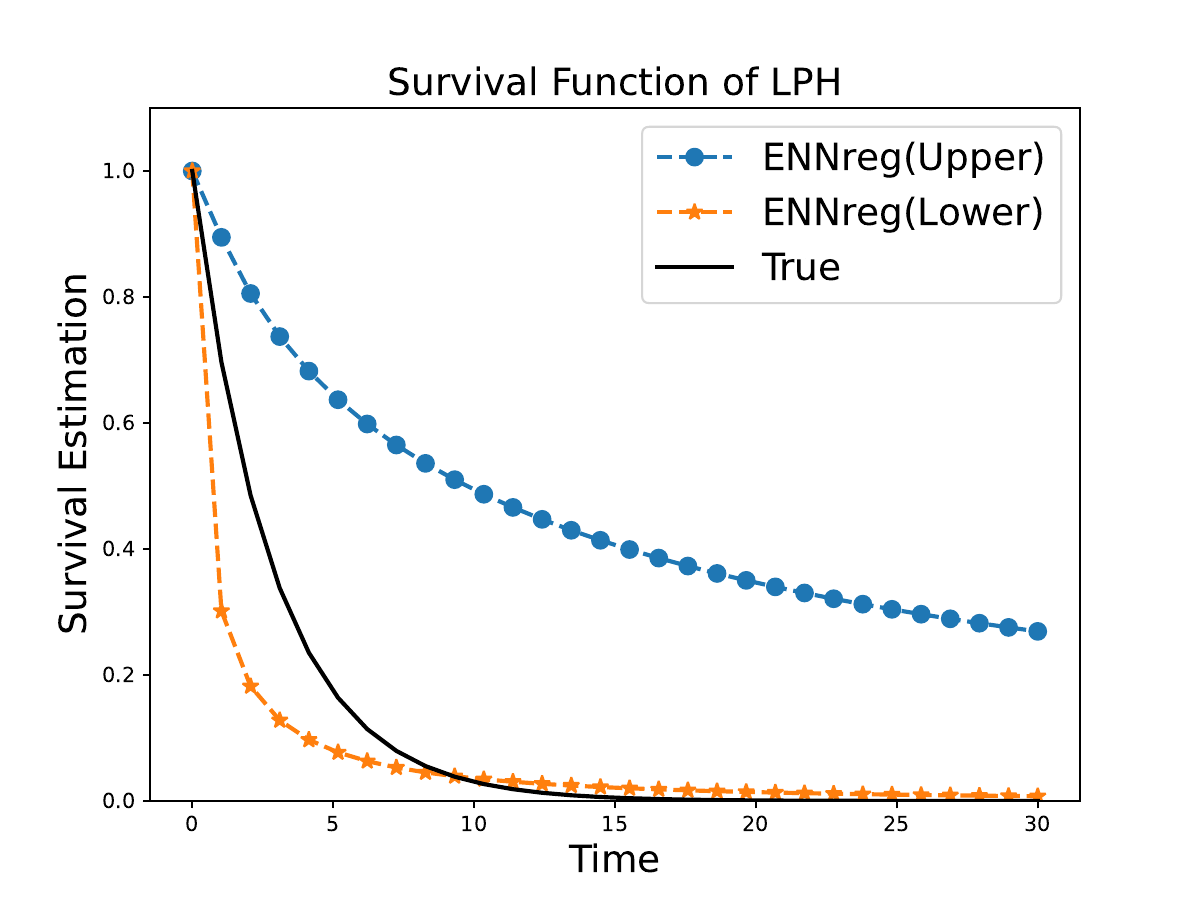}}
\subfloat[\label{fig: s(t)_LPH3_others}]{\includegraphics[width=0.4\textwidth]{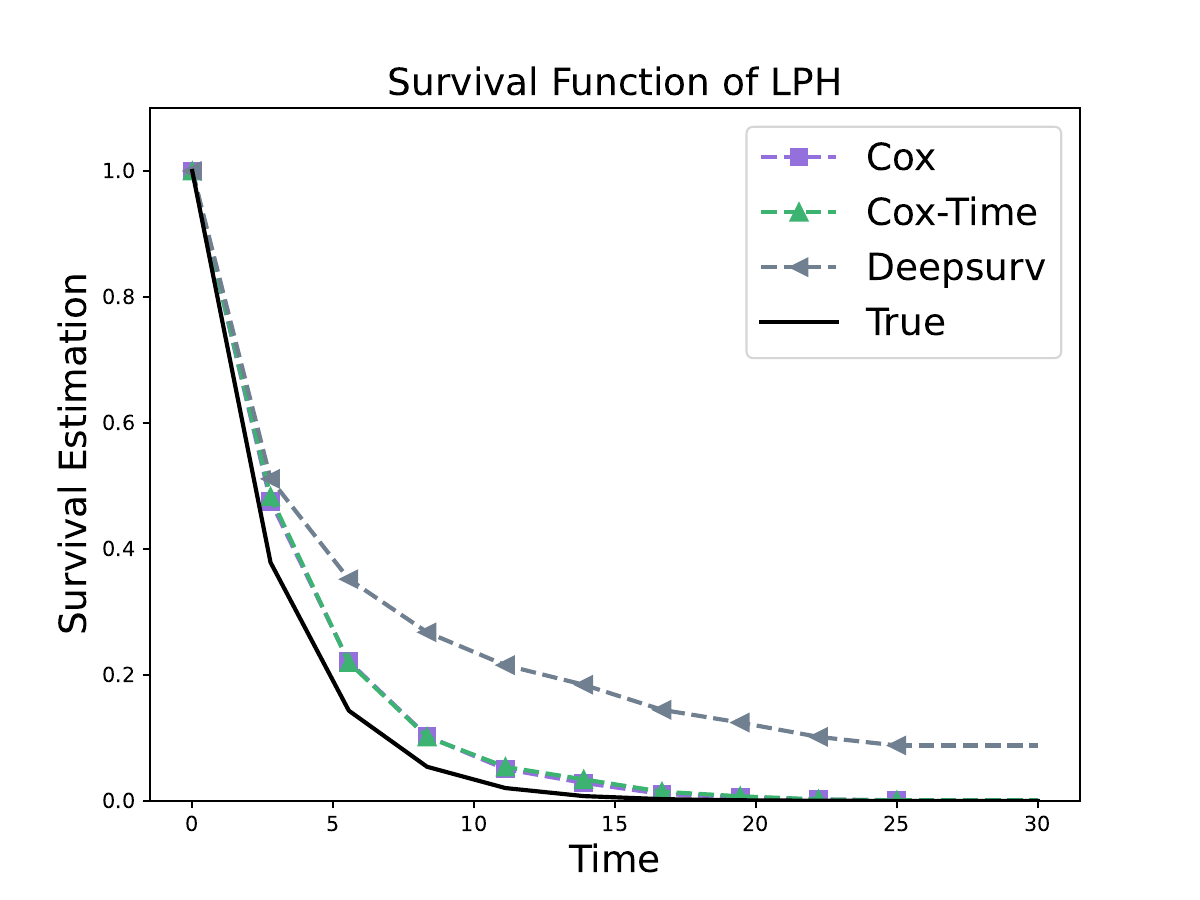}}
\caption{Survival curves for two values of $x$ predicted by ENNreg (a,c) and other methods (b,d) for the dataset generated under the LPH assumption. For ENNreg, the lower and upper  survival functions are plotted.}
\label{fig:st_LPH}
\end{figure}

\begin{figure}
\centering
\subfloat[\label{fig: s(t)_NLPH2_ennreg}]{\includegraphics[width=0.4\textwidth]{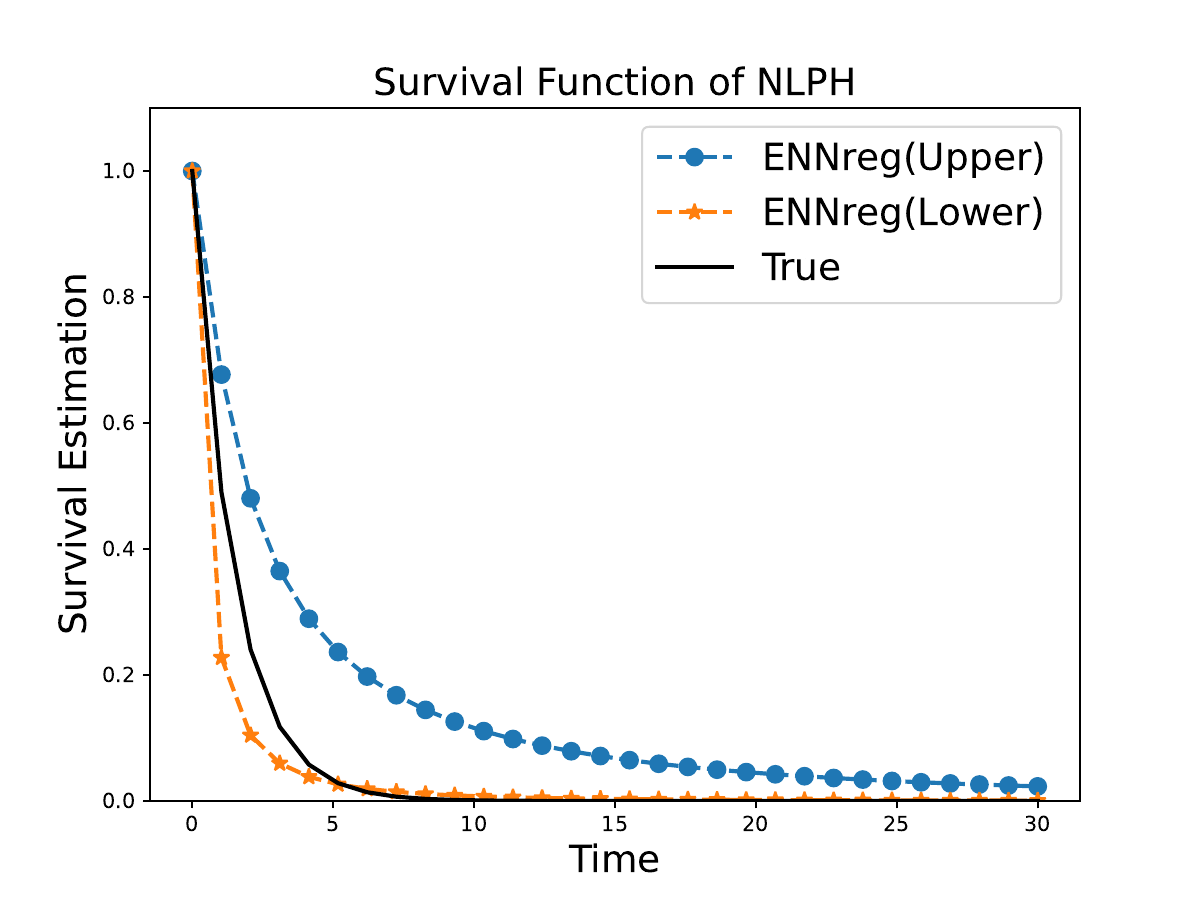}}
\subfloat[\label{fig: s(t)_NLPH2_others}]{\includegraphics[width=0.4\textwidth]{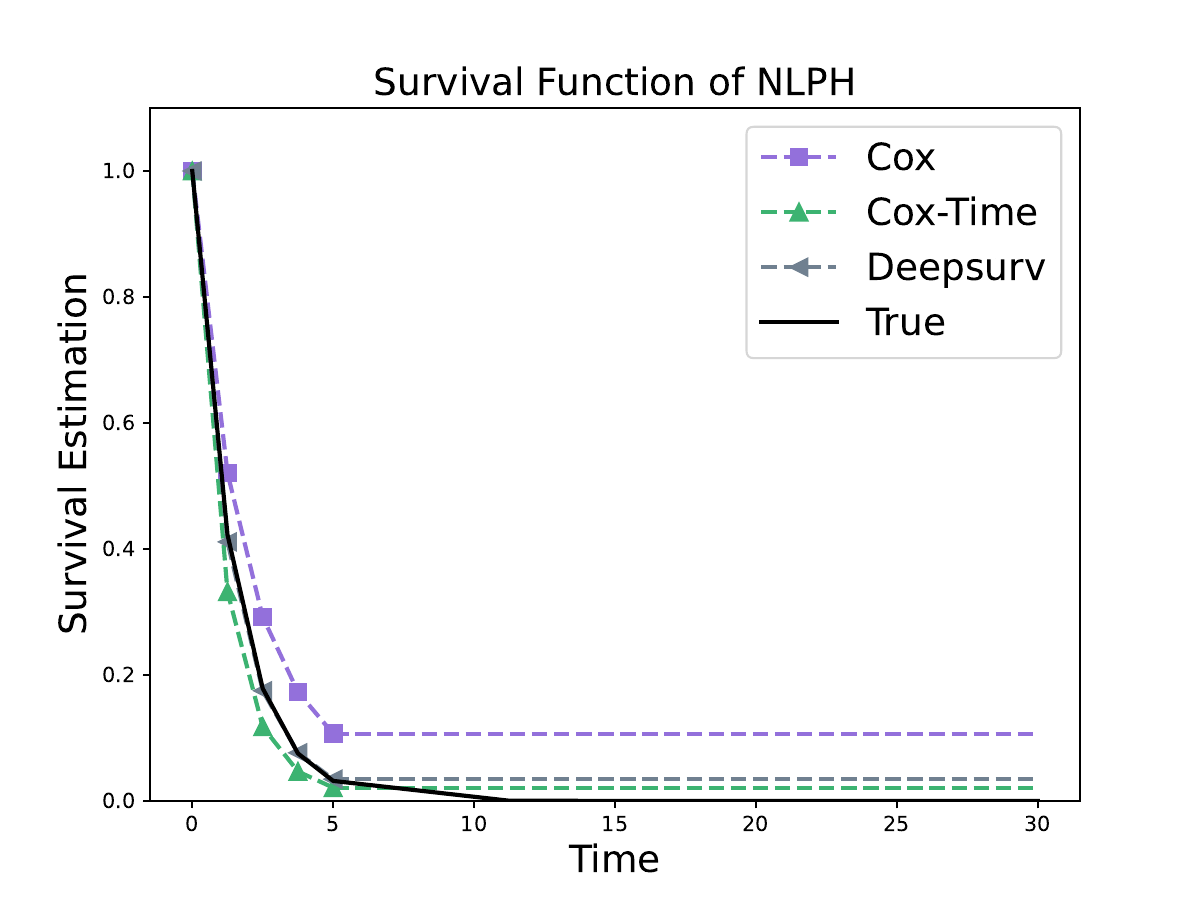}}\\
\subfloat[\label{fig: s(t)_NLPH3_ennreg}]{\includegraphics[width=0.4\textwidth]{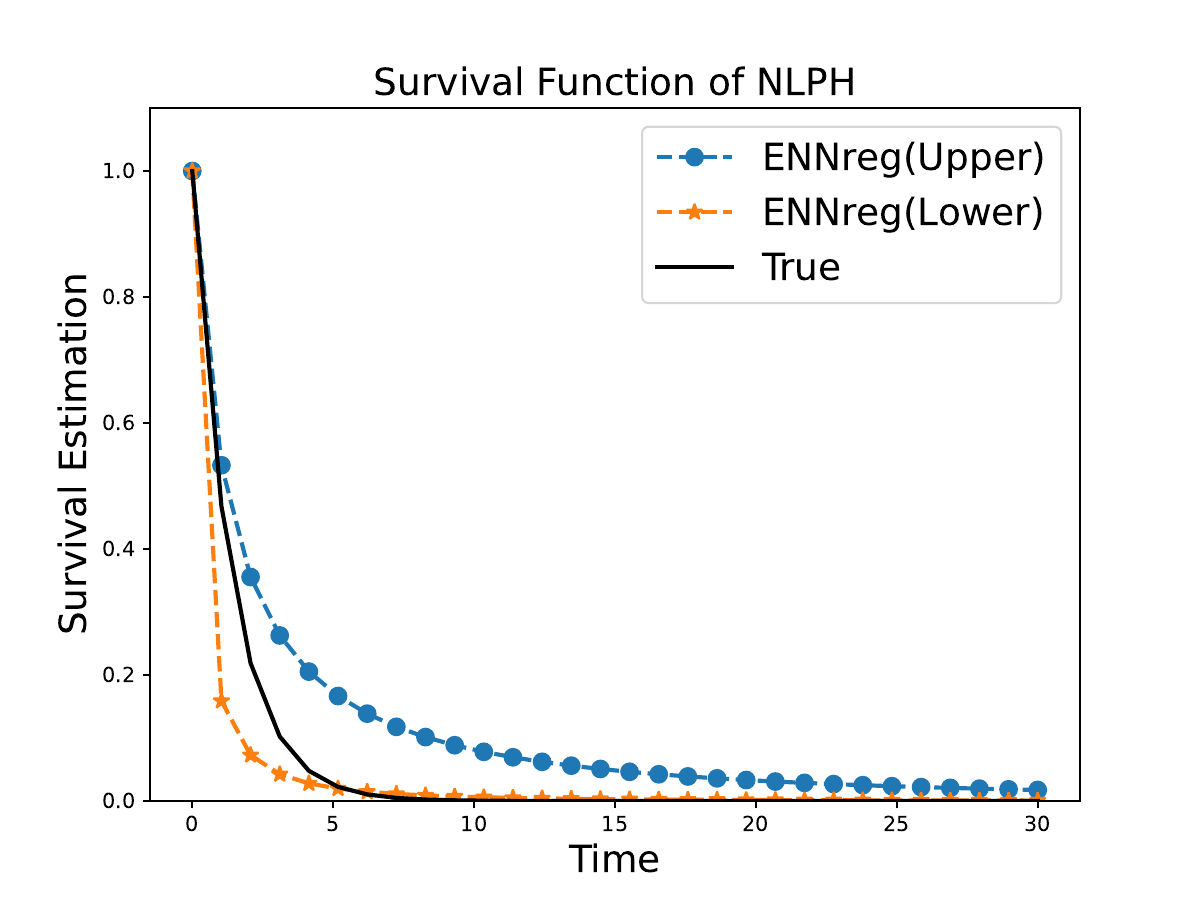}}
\subfloat[\label{fig: s(t)_NLPH3_others}]{\includegraphics[width=0.4\textwidth]{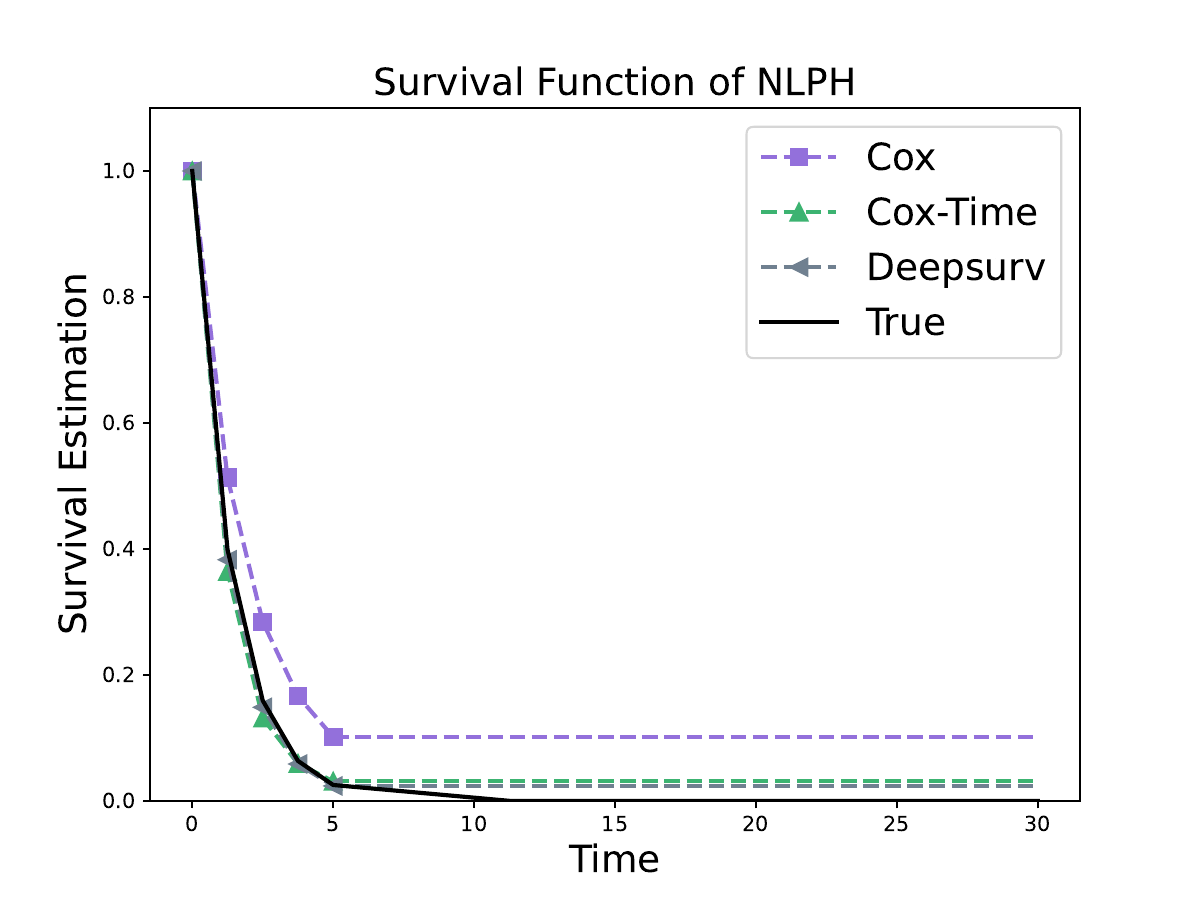}}
\caption{Survival curves for two values of $x$ predicted by ENNreg (a,c) and other methods (b,d) for the dataset generated under the NLPH assumption. For ENNreg, the lower and upper survival functions are plotted.}
\label{fig:st_NLPH}
\end{figure}

\begin{figure}
\centering
\subfloat[\label{fig: s(t)_NLNPH2_ennreg}]{\includegraphics[width=0.4\textwidth]{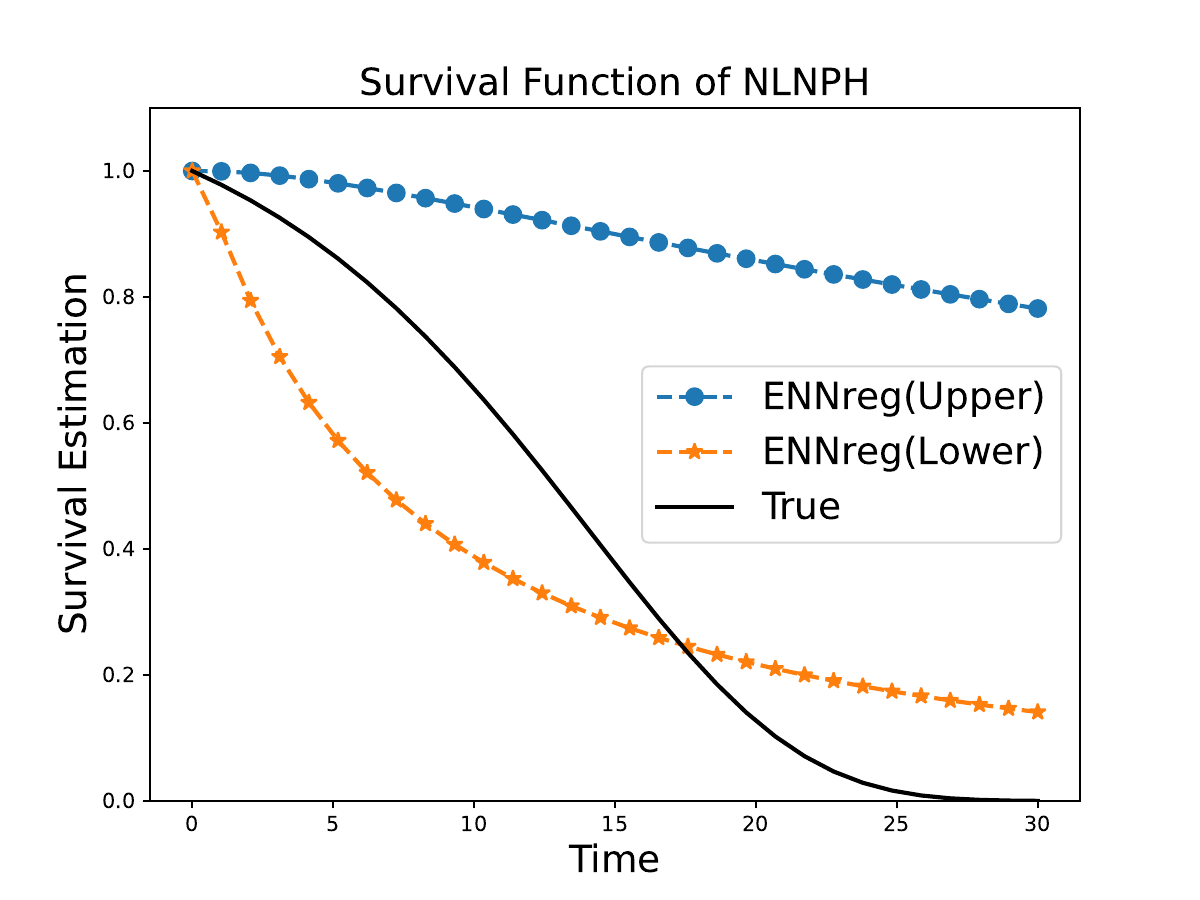}}
\subfloat[\label{fig: s(t)_NLNPH2_others}]{\includegraphics[width=0.4\textwidth]{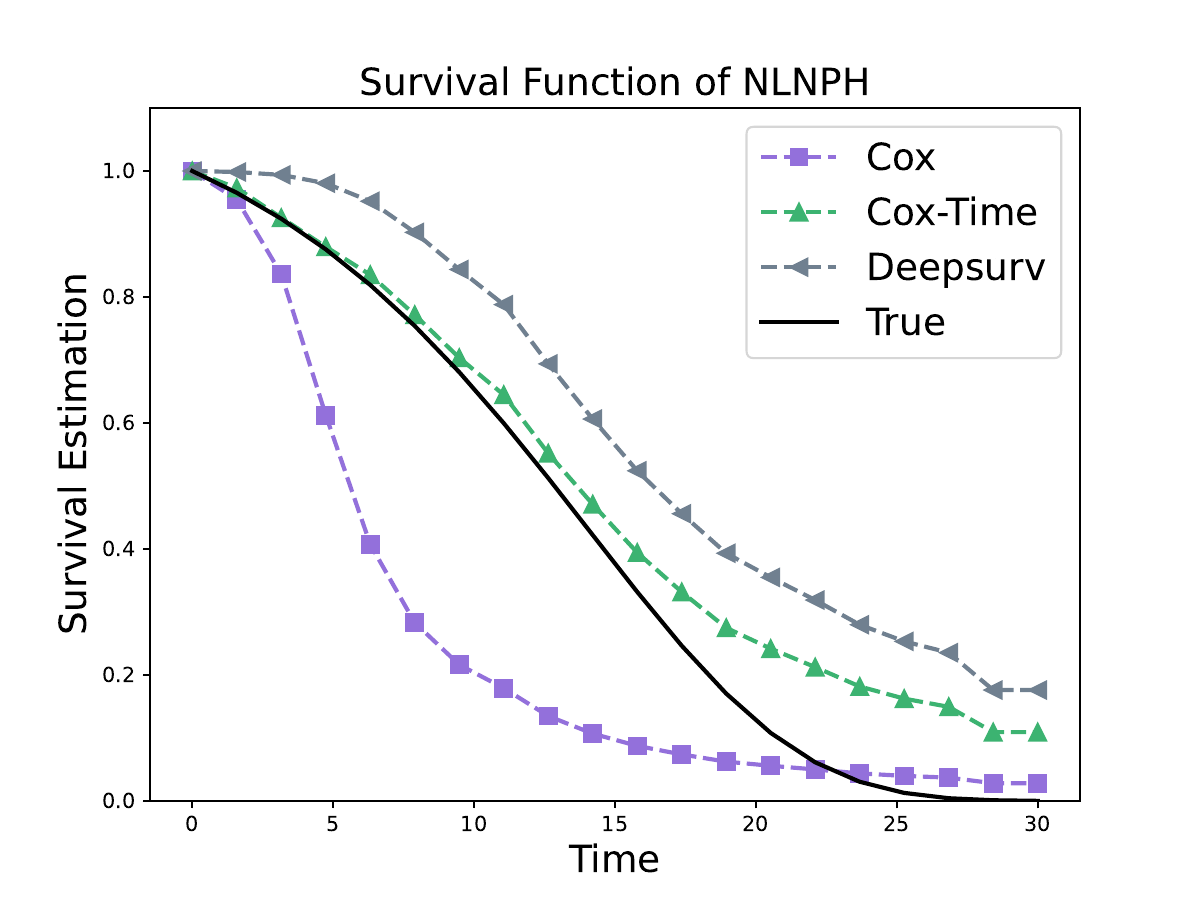}}\\
\subfloat[\label{fig: s(t)_NLNPH3_ennreg}]{\includegraphics[width=0.4\textwidth]{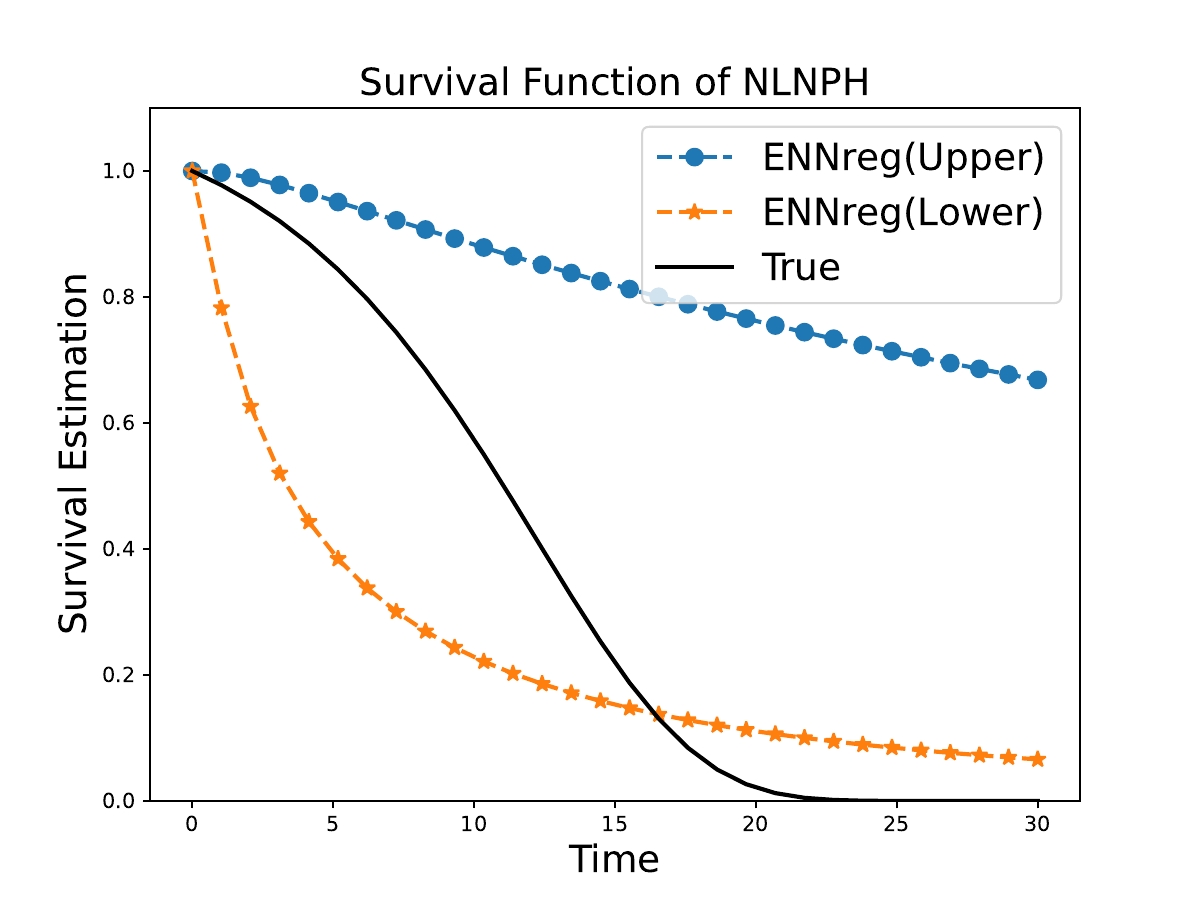}}
\subfloat[\label{fig: s(t)_NLNPH3_others}]{\includegraphics[width=0.4\textwidth]{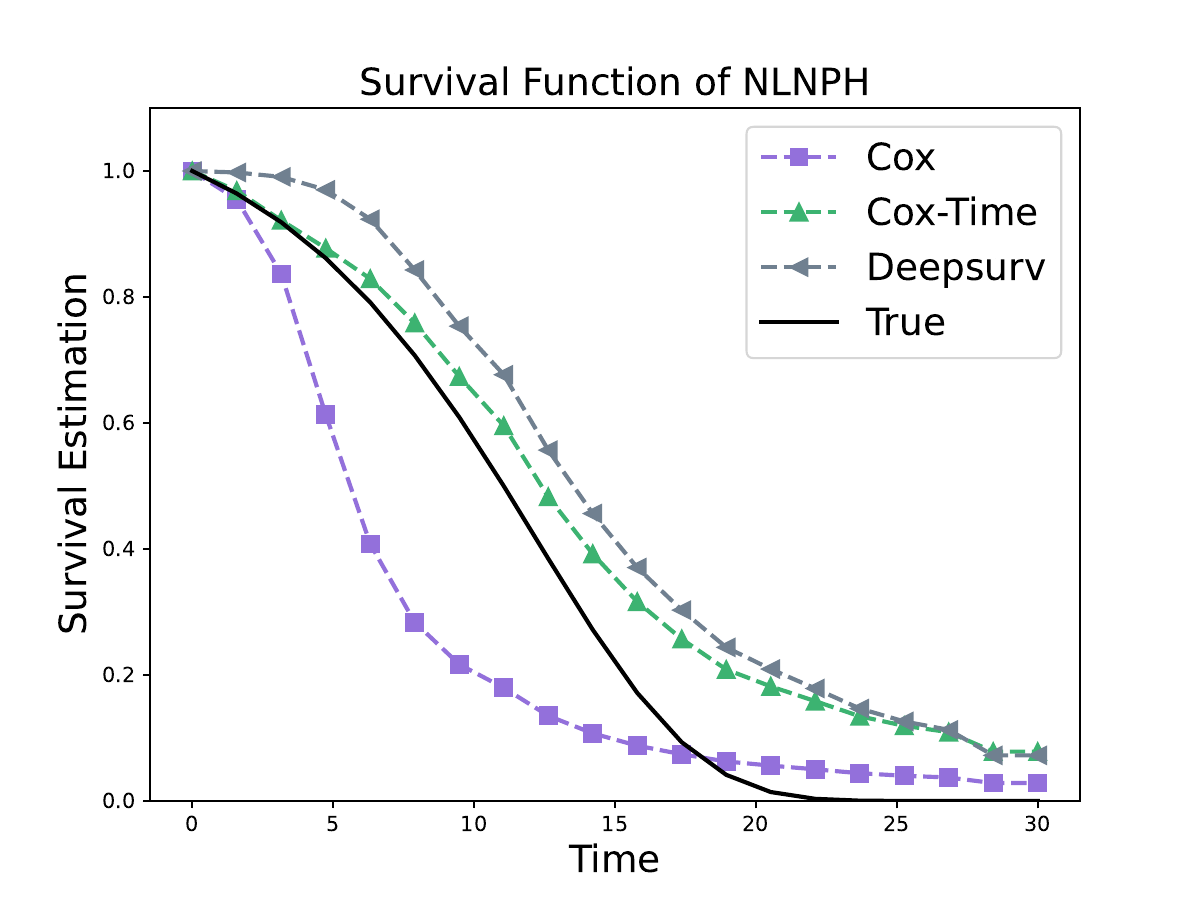}}
\caption{Survival curves for two values of $x$ estimated by ENNreg (a,c) and other methods (b,d) for the dataset generated under the NLNPH assumption. For ENNreg, the lower and upper survival functions are plotted.}
\label{fig:st_NLNPH}
\end{figure}

Figures \ref{fig:st_LPH}, \ref{fig:st_NLPH} and \ref{fig:st_NLNPH} display the survival curves predicted by ENNreg and the three reference methods for two different input values under, respectively, the LPH, NLPH, and NLNPH assumptions. We can see that Cox regression performs well under the LPH assumption (Figures \ref{fig: s(t)_LPH2_others} and  \ref{fig: s(t)_LPH3_others}), and Cox-Time as well as DeepSurv yield good results on the NLPH dataset (Figures \ref{fig: s(t)_NLPH2_others} and  \ref{fig: s(t)_NLPH3_others}). In contrast, none of the reference methods appear to do very well on the NLNPH dataset, Cox-Time doing slightly better than others (Figures \ref{fig: s(t)_NLNPH2_others} and  \ref{fig: s(t)_NLNPH3_others}). The lower and upper survivals functions computed by ENNreg appear to be quite conservative. They are reasonably good imprecise approximations of the true curves for the LPH and NLPH datasets (Figures \ref{fig: s(t)_LPH2_ennreg},  \ref{fig: s(t)_LPH3_ennreg}, \ref{fig: s(t)_NLPH2_ennreg} and \ref{fig: s(t)_NLPH3_ennreg}), less so on the NLNPH dataset (Figures \ref{fig: s(t)_NLNPH2_ennreg} and  \ref{fig: s(t)_NLNPH3_ennreg}). The very large intervals in this last case may reflect the greater difficulty of the learning task.

\subsection{Comparative results on real-world datasets}
\label{subsec: real_world}
The third experiment aims at comparing the performances of our method with those of state-of-the-art models using four real-world datsets. 

\paragraph{Datasets}

\begin{table}
\caption{Summary of the four real datasets used in the experiments.}
\centering
\begin{tabular}{cccccc}
\hline
Dataset &Size  & Covariates &   \multicolumn{2}{c}{Duration (days)} & Censoring rate\\
\cline{4-5}
               &  &  & Min &Max & \\
    \hline
    METABRIC & 1,904 & 9& 0  &355 &0.42 \\
    GBSG& 2,232 & 7 & 0.2  & 87& 0.43\\
    SUPPORT& 8,873 & 14 &2  &2029 & 0.32\\
    MIMIC-IV& 53,512 & 24 &0 & 12.3& 0.89\\
    \hline
    \end{tabular}
    
    \label{tab: dataset}
\end{table}

We used four real-world datasets whose main characteristics are summarized in Table \ref{tab: dataset}.  METABRIC, GBSG and SUPPORT are survival datasets publicly available through the Pycox python package\footnote{\url{https://github.com/havakv/pycox}} \cite{kvamme2019time} \cite{katzman2018deepsurv};  a detailed description of these three datasets, which do not require any preprocessing, can be found in \cite{kvamme2019time}. %In addition to survival datasets, we also considered an ICU dataset from the publicly available MIMIC-IV database \cite{johnson2020mimic}. 
 MIMIC-IV  \cite{johnson2020mimic} concerns stays in ICU; it contains detailed information such as vital signs, laboratory test results, medications, clinical notes, imaging reports, and demographic details. For this dataset, we extracted 24 features based on the first 24-hour clinical data following \cite{purushotham2018benchmarking}. The hospital length of stay is regarded as the event time. The event of interest is defined as the mortality after admission. The event time is observed if there is a record of death in the database; otherwise, the censored time is defined as the last time of being discharged from the hospital. Samples with more than 50\% missing variables were discarded directly, and multiple imputation was used to handle missing data. 

\paragraph{Methodology} As reference methods, we considered the classical Cox model \cite{cox1972regression}, two traditional ML-based survival models: RSF \cite{ishwaran2008random} and SurvivalSVM \cite{polsterl2015fast}; and four state-of-the-art deep learning-based survival models: DeepSurv \cite{katzman2018deepsurv}, Cox-CC \cite{kvamme2019time}, Cox-Time \cite{kvamme2019time}, and Deephit \cite{lee2018deephit}. For Cox regression, we used the implementation in the Python package {\tt lifelines}\footnote{\url{https://lifelines.readthedocs.io/en/latest/}}. For RSF and SurvivalSVM, we used the Python package {\tt scikit-survival}\footnote{\url{https://scikit-survival.readthedocs.io/en/stable/}}. 
For the deep neural network-based methods, we used the implementation in the Python package {\tt Pycox}. 
Hyperparameters for all methods included in the comparison were set according to the recommendations in their original papers or software packages when applicable. However, for the larger-scale MIMIC-IV dataset, testing Cox and RSF posed challenges due to the high correlation among covariates or computational costs. To address these issues, we applied necessary parameter adjustments for Cox and  RSF to ensure compatibility with MIMIC-IV. Specifically, for Cox, we set the penalizer to 0.1 to address high covariate correlation and improve estimation stability. For RSF, the number of trees was reduced to 50 to reduce computational costs. For ENNreg, the settings were the same as described in Section \ref{subsec: simulated}. Each dataset was also randomly split into training, validation, and test sets containing, respectively, 60\%, 20\%, and 20\% of the observations, and these random splits were repeated five times. For all compared methods, we report the $C_{idx}$, $IBS$, and $IBLL$ metrics, except for SurvivalSVM: for this method, we only calculated the $C_{idx}$ as it is designed to maximize the concordance index for pairs of observations, and its outputs do not make it possible to compute the $IBS$ and $IBLL$ measures.

\paragraph{Results on the survival datasets} As shown in Tables \ref{tab:result_metabric}-\ref{tab:result_support}, ENNreg achieved the best performance according to the $C_{idx}$ metric on all three datasets, while  DeepHit, Cox-Time and RSF yielded the second best performances on the METABRIC, GBSG and SUPPORT datasets, respectively. Cox, SurvivalSVM and DeepSurv performed rather poorly in terms of prediction accuracy on the three datasets. This could be expected in the case of Cox and DeepSurv, as they rely on strong distributional assumptions. In terms of prediction calibration, as measured by the $IBS$ and $IBLL$ metrics, our method yields the best results on the GBSG dataset (Table \ref{tab:result_GBSG}) and the second best on the METABRIC dataset (Table \ref{tab:result_metabric}). Overall, ENNreg is quite competitive on these three survival datasets in terms of prediction accuracy and calibration, as compared to state-of-the-art methods.

\begin{table}
\centering
\caption{Comparison of prediction performance (mean and standard error) on the METABRIC dataset. The best and second best results are, resp., in bold and underlined.}
\begin{tabular}{|l|c|c|c|}
\hline
%\rowcolor[HTML]{E0E0E0} 
\textbf{Methods} & \multicolumn{1}{c|}{\textbf{Accuracy }} & \multicolumn{2}{c|}{\textbf{Uncertainty }}\\ \hline
%\rowcolor[HTML]{E0E0E0} 
 &  \multicolumn{1}{c|}{\textbf{$C_{idx}\uparrow$}} & \multicolumn{1}{c|}{\textbf{IBS $\downarrow $}} & 
\multicolumn{1}{c|}{ \textbf{IBLL$\downarrow $} } \\ \hline
%\rowcolor[HTML]{E0E0E2} 
%\multicolumn{4}{|c|}{\textbf{METABRIC}}  \\\hline
Cox  &  0.643$\pm0.010$ &0.172$\pm0.002$ &0.515$\pm0.005$	\\
RSF  &0.626$\pm0.008$  & 0.176$\pm0.002$ &0.523$\pm0.006$  \\
SurvivalSVM &0.649$\pm0.011$  & $-$ &$-$   \\
DeepSurv &0.642$\pm0.037$	 &\underline{0.167}$\pm0.012$ &\textbf{0.502}$\pm0.031$	 \\
Cox-CC   &0.643$\pm0.006$& 0.171$\pm0.001$ &0.514$\pm0.004$  \\
Cox-Time &0.661$\pm0.005$	 & \textbf{0.165}$\pm0.002$	& \textbf{0.502}$\pm0.010$	\\
DeepHit  &\underline{0.669}$\pm0.010$ & 0.180$\pm0.002$ & 0.534$\pm0.004$\\
ENNreg & \textbf{0.672}$\pm0.007$&\underline{0.167}$\pm0.001$&\underline{0.508}$\pm0.002$ \\
 \hline
 \end{tabular}
\label{tab:result_metabric}
\end{table}
 
 \begin{table}
\centering
 \caption{Comparison of prediction performance (mean and standard error) on the GBSG dataset. The best and second best results are, resp., in bold and underlined.
}
\begin{tabular}{|l|c|c|c|}
\hline
%\rowcolor[HTML]{E0E0E0} 
\textbf{Methods} & \multicolumn{1}{c|}{\textbf{Accuracy }} & \multicolumn{2}{c|}{\textbf{Uncertainty }}\\ \hline
%\rowcolor[HTML]{E0E0E0} 
 &  \multicolumn{1}{c|}{\textbf{$C_{idx}\uparrow$}} & \multicolumn{1}{c|}{\textbf{IBS $\downarrow $}} & 
\multicolumn{1}{c|}{ \textbf{IBLL$\downarrow $} } \\ \hline
%\rowcolor[HTML]{E0E0E2} 
%\multicolumn{4}{|c|}{ \textbf{GBSG}}\\\hline
Cox   & 0.662$\pm0.006$ &0.183$\pm0.0009$& 0.539$\pm0.002$\\
RSF &0.663$\pm0.006$  & 0.180$\pm0.001$ &0.531$\pm0.003$  \\
SurvivalSVM &0.666$\pm0.006$  & $-$ & $-$ \\
DeepSurv  & 0.669$\pm0.043$& 0.180$\pm0.007$& 0.531$\pm0.021$\\
Cox-CC    &  {0.673}$\pm0.003$ & \underline{0.179}$\pm0.001$ & 0.536$\pm0.004$\\
Cox-Time & \underline{0.672}$\pm0.005$ &\underline{0.179}$\pm0.002$ &\underline{0.530}$\pm0.006$\\
DeepHit & 0.667$\pm0.005$& 0.200$\pm0.004$& 0.582$\pm0.014$\\
ENNreg & \textbf{0.679}$\pm0.005$&\textbf{0.178}$\pm0.001$&\textbf{0.525}$\pm0.003$\\
 \hline
  \end{tabular}
\label{tab:result_GBSG}
\end{table}

 \begin{table}
\centering
\caption{Comparison of prediction performance (mean and standard error) on the SUPPORT dataset. The best and second best results are, resp., in bold and underlined.
}
\begin{tabular}{|l|c|c|c|}
\hline
%\rowcolor[HTML]{E0E0E0} 
\textbf{Methods} & \multicolumn{1}{c|}{\textbf{Accuracy }} & \multicolumn{2}{c|}{\textbf{Uncertainty }}\\ \hline
%\rowcolor[HTML]{E0E0E0} 
 &  \multicolumn{1}{c|}{\textbf{$C_{idx}\uparrow$}} & \multicolumn{1}{c|}{\textbf{IBS $\downarrow $}} & 
\multicolumn{1}{c|}{ \textbf{IBLL$\downarrow $} } \\ \hline

%\rowcolor[HTML]{E0E0E2} 
%\multicolumn{4}{|c|}{\textbf{SUPPORT}} \\\hline
Cox  &0.568$\pm0.002$&0.204$\pm0.0009$&0.593$\pm0.002$ \\
RSF  & \underline{0.624}$\pm0.002$  & \textbf{0.192}$\pm0.001$ & \textbf{0.564}$\pm0.003$  \\
SurvivalSVM  &0.572$\pm0.001$  &$-$  &$-$\\
DeepSurv& 0.602$\pm0.004$&0.194$\pm0.001$& 0.569$\pm0.004$ \\
Cox-CC   & 0.603$\pm0.002$ &\textbf{0.192}$\pm0.0008$ &\underline{0.567}$\pm0.003$ \\
Cox-Time& 0.610$\pm0.002$& \underline{0.193}$\pm0.001$&\underline{0.567}$\pm0.003$\\
DeepHit  &  \textbf{0.626}$\pm0.002$& 0.206$\pm0.0008$& 0.600$\pm0.001$ \\
ENNreg  &  \textbf{0.626}$\pm0.004$& 0.196$\pm0.001$& 0.574$\pm0.003$\\
\hline
\end{tabular}
\label{tab:result_support}
\end{table}

\paragraph{Results on the MIMIC-IV dataset}

Table \ref{tab: result_mimic} shows the results on the MIMIC-IV dataset. On this dataset, ENNreg has the highest accuracy as measured by the  $C_{idx}$ metric, and it yields the second-based results in terms of calibration, as measured by the $IBS$ and $IBLL$ metrics. We can remark that,  with such a large dataset,  DeepHit does not make as accurate predictions as those of continuous models such as DeepSurv and Cox-Time because of the loss of information resulting in event time discretization. However, it makes well-calibrated predictions. Overall, ENNreg is quite competitive on this dataset, which confirms the conclusions of the previous analysis on survival datasets. 

From the results reported in this section as well as those reported in Section \ref{subsec: exp_illustrative} and  \ref{subsec: simulated}, we can conclude that our evidence-based time-to-event prediction model, relying on minimal assumptions, exhibits greater flexibility and robustness than those of classical and deep-learning models. Interestingly, our model achieves very good performance even a censoring rate as high as 89\% observed with the  MIMIC-IV dataset, which confirms the qualitative analysis performed in Section \ref{subsec: exp_illustrative}.

\begin{table}
\centering
\caption{Comparison of prediction performance (mean and standard error) on the MIMIC-IV dataset. The best and second best results are, resp., in bold and underlined.
}
\begin{tabular}{|l|c|c|c|}
\hline
%\rowcolor[HTML]{E0E0E2} 
%\multicolumn{4}{|c|}{\textbf{MIMIC-IV}} \\\hline

\textbf{Methods} & \multicolumn{1}{c|}{\textbf{Accuracy}} & \multicolumn{2}{c|}{\textbf{Uncertainty}}\\ \hline
%\rowcolor[HTML]{E0E0E0} 
%\rowcolor[HTML]{E0E0E0} 
 &  \multicolumn{1}{c|}{\textbf{$C_{idx}\uparrow$}} & \multicolumn{1}{c|}{\textbf{IBS $\downarrow $}} & 
\multicolumn{1}{c|}{ \textbf{IBLL$\downarrow $} } \\ \hline

Cox  &0.779$\pm0.005$& 0.193$\pm0.005$ &0.581$\pm0.018$\\
RFS & 0.814$\pm0.004$ &0.250$\pm0.005$ & 0.692$\pm0.013$\\
SurvivalSVM  & 0.791$\pm0.005$  &$-$  &$-$\\
DeepSurv  & 0.834$\pm0.004$ &0.220$\pm0.007$&0.698 $\pm0.033$ \\
Cox-CC    & 0.837$\pm0.003$ &{0.209}$\pm0.010$ &{0.663}$\pm0.050$ \\
Cox-Time & \underline{0.841}$\pm0.003$&0.193$\pm0.009$&0.575$\pm0.025$\\
DeepHit & 0.823$\pm0.004$&\textbf{0.184}$\pm0.008$&\textbf{0.548}$\pm0.022$  \\
ENNreg &  \textbf{0.845}$\pm0.003$&\underline{0.190}$\pm0.008$&\underline{0.572}$\pm0.026$ \\
\hline
\end{tabular}
\label{tab: result_mimic}
\end{table}

\section{Conclusion}
\label{sec:conclu} 

In this paper, we extended the evidential regression neural network introduced in \cite{denoeux2023quantifying} to account for data censoring and applied it to time-to-event prediction tasks. In contrast with most survival prediction algorithms, our method is not based on a model relating the hazard function to attributes. Instead, it directly predicts the time to event using a lognormal random fuzzy number.  The model is fit using a generalized negative log-likelihood function that accounts for data censoring. The output random fuzzy set induces a predictive belief function that can be used to compute, e.g., lower and upper conditional survival functions. It is also possible to compute belief intervals on the time to event, which were shown experimentally to be well calibrated.   
Experimental results with simulated datasets under various data distribution and censoring scenarios, as well as with real-world datasets across different clinical settings and tasks, show that our model consistently achieves both accurate and reliable performance compared to state-of-the-art methods.

This study can be extended in several directions. For complex and large-scale data, a deep feature-extraction architecture could be combined with ENNreg, resulting in a deep evidential neural network as introduced in \cite{tong2021evidential} for classification tasks. For multi-source or multimodal data, the fusion of source-dependent predictions using discounting and combination operations of ERFS theory could be considered,  using a  approach  similar to that explored in \cite{huang2025deep} for image segmentation. The dependence between sources could also be taken into account in the fusion process, as proposed  in \cite{denoeux2024combination}. Finally, the ENNreg model could be modified to output a mixture of GFRNs, as introduced in \cite{denoeux2023parametric}, for more complex prediction tasks arising, e.g., in financial or environmental applications.

\section*{Acknowledgements} 
This research was funded by A*STAR, CISCO Systems (USA) Pte. Ltd, and the National University of Singapore under its Cisco-NUS Accelerated Digital Economy Corporate Laboratory (Award I21001E0002) and the National Research Foundation Singapore under the AI Singapore Programme (Award AISG-GC-2019-001-2B).

%%%%%%%%%%%%%%%%%%

%%%%%%%%%%%%%%%%%%

%\appendix

%\section{Sample Appendix Section}
%Lorem ipsum dolor sit amet, consectetur adipiscing elit, sed do eiusmod tempor section \ref{sec:sample1} incididunt ut labore et dolore magna aliqua. Ut enim ad minim veniam, quis nostrud exercitation ullamco laboris nisi ut aliquip ex ea commodo consequat. Duis aute irure dolor in reprehenderit in voluptate velit esse cillum dolore eu fugiat nulla pariatur. Excepteur sint occaecat cupidatat non proident, sunt in culpa qui officia deserunt mollit anim id est laborum.

% \bibliographystyle{elsarticle-num} 
 \bibliographystyle{abbrv}
 \bibliography{cas-refs}





% \end{thebibliography}
\end{document}

%% file: preambule.tex
\usepackage{amsfonts,amstext,amsmath,amssymb,alltt}
\usepackage{algorithmic,algorithm}
\usepackage{url}

\newcommand{\bi}{\begin{itemize}}
\newcommand{\ei}{\end{itemize}}
\newcommand{\be}{\begin{enumerate}}
\newcommand{\ee}{\end{enumerate}}
\newcommand{\bd}{\begin{description}}
\newcommand{\ed}{\end{description}}

\newcommand{\calB}{\mathcal{B}}

\newcommand{\calF}{\mathcal{F}}
\newcommand{\calL}{\mathcal{L}}

\newcommand{\tF}{{\widetilde{F}}}

\newcommand{\tP}{{\widetilde{P}}}
\newcommand{\tN}{{\widetilde{N}}}

\newcommand{\tX}{{\widetilde{X}}}

\newcommand{\tY}{{\widetilde{Y}}}
\newcommand{\tT}{{\widetilde{T}}}

\newcommand{\tTheta}{{\widetilde{\Theta}}}

\newcommand{\tpsi}{{\widetilde{\psi}}}

\def\hbar{\overline{h}}

\newcommand{\reels}{\mathbb{R}}                                       % Entiers

\newcommand{\GFN}{\mathsf{GFN}}

\newcommand{\bmat}{\begin{pmatrix}}
\newcommand{\emat}{\end{pmatrix}}

\newcommand{\fracpar}[2]{\left(\frac{#1}{#2}\right)}
\def\simH0{\stackrel{H_0}{\sim}}

%% file: IJAR2024.bbl
\begin{thebibliography}{10}

\bibitem{antolini2005time}
L.~Antolini, P.~Boracchi, and E.~Biganzoli.
\newblock A time-dependent discrimination index for survival data.
\newblock {\em Statistics in medicine}, 24(24):3927--3944, 2005.

\bibitem{breiman2001random}
L.~Breiman.
\newblock Random forests.
\newblock {\em Machine learning}, 45:5--32, 2001.

\bibitem{chai2024uncertainty}
H.~Chai, S.~Lin, J.~Lin, M.~He, Y.~Yang, Y.~OuYang, and H.~Zhao.
\newblock An uncertainty-based interpretable deep learning framework for predicting breast cancer outcome.
\newblock {\em BMC bioinformatics}, 25(1):88, 2024.

\bibitem{chapfuwa2020calibration}
P.~Chapfuwa, C.~Tao, C.~Li, I.~Khan, K.~J. Chandross, M.~J. Pencina, L.~Carin, and R.~Henao.
\newblock Calibration and uncertainty in neural time-to-event modeling.
\newblock {\em IEEE transactions on neural networks and learning systems}, 34(4):1666--1680, 2020.

\bibitem{christ2017survivalnet}
P.~F. Christ, F.~Ettlinger, G.~Kaissis, S.~Schlecht, F.~Ahmaddy, F.~Gr{\"u}n, A.~Valentinitsch, S.-A. Ahmadi, R.~Braren, and B.~Menze.
\newblock Survivalnet: Predicting patient survival from diffusion weighted magnetic resonance images using cascaded fully convolutional and 3d convolutional neural networks.
\newblock In {\em 2017 IEEE 14th International Symposium on Biomedical Imaging (ISBI 2017)}, pages 839--843. IEEE, 2017.

\bibitem{ciampi1981approach}
A.~Ciampi, R.~Bush, M.~Gospodarowicz, and J.~Till.
\newblock An approach to classifying prognostic factors related to survival experience for non-hodgkin's lymphoma patients: Based on a series of 982 patients: 1967--1975.
\newblock {\em Cancer}, 47(3):621--627, 1981.

\bibitem{couso11}
I.~Couso and L.~S\'anchez.
\newblock Upper and lower probabilities induced by a fuzzy random variable.
\newblock {\em Fuzzy Sets and Systems}, 165(1):1--23, 2011.

\bibitem{cox1972regression}
D.~R. Cox.
\newblock Regression models and life-tables.
\newblock {\em Journal of the Royal Statistical Society: Series B (Methodological)}, 34(2):187--202, 1972.

\bibitem{dawid1982well}
A.~P. Dawid.
\newblock The well-calibrated bayesian.
\newblock {\em Journal of the American Statistical Association}, 77(379):605--610, 1982.

\bibitem{dempster67}
A.~P. Dempster.
\newblock Upper and lower probabilities induced by a multivalued mapping.
\newblock {\em Annals of Mathematical Statistics}, 38:325--339, 1967.

\bibitem{denoeux2021belief}
T.~Den{\oe}ux.
\newblock Belief functions induced by random fuzzy sets: A general framework for representing uncertain and fuzzy evidence.
\newblock {\em Fuzzy Sets and Systems}, 424:63--91, 2021.

\bibitem{denoeux22}
T.~Den{\oe}ux.
\newblock An evidential neural network model for regression based on random fuzzy numbers.
\newblock In S.~Le~H{\'e}garat-Mascle, I.~Bloch, and E.~Aldea, editors, {\em Belief Functions: Theory and Applications}, pages 57--66, Cham, 2022. Springer International Publishing.

\bibitem{denoeux2023parametric}
T.~Den{\oe}ux.
\newblock Parametric families of continuous belief functions based on generalized {Gaussian} random fuzzy numbers.
\newblock {\em Fuzzy Sets and Systems}, 471:108679, 2023.

\bibitem{denoeux2023quantifying}
T.~Den{\oe}ux.
\newblock Quantifying prediction uncertainty in regression using random fuzzy sets: the {ENNreg} model.
\newblock {\em IEEE Transactions on Fuzzy Systems}, 31:3690--3699, 2023.

\bibitem{denoeux2023reasoning}
T.~Den{\oe}ux.
\newblock Reasoning with fuzzy and uncertain evidence using epistemic random fuzzy sets: General framework and practical models.
\newblock {\em Fuzzy Sets and Systems}, 453:1--36, 2023.

\bibitem{denoeux2024combination}
T.~Denoeux.
\newblock Combination of dependent and partially reliable {Gaussian} random fuzzy numbers.
\newblock {\em Information Sciences}, 681:121208, 2024.

\bibitem{denoeux2024uncertainty}
T.~Den{\oe}ux.
\newblock Uncertainty quantification in logistic regression using random fuzzy sets and belief functions.
\newblock {\em International Journal of Approximate Reasoning}, 168:109159, 2024.

\bibitem{denoeux20b}
T.~Den{\oe}ux, D.~Dubois, and H.~Prade.
\newblock Representations of uncertainty in artificial intelligence: Beyond probability and possibility.
\newblock In P.~Marquis, O.~Papini, and H.~Prade, editors, {\em A Guided Tour of Artificial Intelligence Research}, volume~1, chapter~4, pages 119--150. Springer Verlag, 2020.

\bibitem{dietterich2002ensemble}
T.~G. Dietterich et~al.
\newblock Ensemble learning.
\newblock {\em The handbook of brain theory and neural networks}, 2(1):110--125, 2002.

\bibitem{evers2008sparse}
L.~Evers and C.-M. Messow.
\newblock Sparse kernel methods for high-dimensional survival data.
\newblock {\em Bioinformatics}, 24(14):1632--1638, 2008.

\bibitem{faraggi1995neural}
D.~Faraggi and R.~Simon.
\newblock A neural network model for survival data.
\newblock {\em Statistics in medicine}, 14(1):73--82, 1995.

\bibitem{fard2016bayesian}
M.~J. Fard, P.~Wang, S.~Chawla, and C.~K. Reddy.
\newblock A {Bayesian} perspective on early stage event prediction in longitudinal data.
\newblock {\em IEEE Transactions on Knowledge and Data Engineering}, 28(12):3126--3139, 2016.

\bibitem{fotso2018deep}
S.~Fotso.
\newblock Deep neural networks for survival analysis based on a multi-task framework.
\newblock {\em arXiv preprint arXiv:1801.05512}, 2018.

\bibitem{genders2011clinical}
T.~S. Genders, E.~W. Steyerberg, H.~Alkadhi, S.~Leschka, L.~Desbiolles, K.~Nieman, T.~W. Galema, W.~B. Meijboom, N.~R. Mollet, P.~J. de~Feyter, et~al.
\newblock A clinical prediction rule for the diagnosis of coronary artery disease: validation, updating, and extension.
\newblock {\em European heart journal}, 32(11):1316--1330, 2011.

\bibitem{graf1999assessment}
E.~Graf, C.~Schmoor, W.~Sauerbrei, and M.~Schumacher.
\newblock Assessment and comparison of prognostic classification schemes for survival data.
\newblock {\em Statistics in medicine}, 18(17-18):2529--2545, 1999.

\bibitem{gyorffy2010online}
B.~Gy{\"o}rffy, A.~Lanczky, A.~C. Eklund, C.~Denkert, J.~Budczies, Q.~Li, and Z.~Szallasi.
\newblock An online survival analysis tool to rapidly assess the effect of 22,277 genes on breast cancer prognosis using microarray data of 1,809 patients.
\newblock {\em Breast cancer research and treatment}, 123:725--731, 2010.

\bibitem{hageman2024estimating}
S.~H. Hageman, R.~A. Post, F.~L. Visseren, J.~W. McEvoy, J.~W. Jukema, Y.~Smulders, M.~van Smeden, J.~A. Dorresteijn, U.-S. study group, et~al.
\newblock Estimating uncertainty when providing individual cardiovascular risk predictions: a {Bayesian} survival analysis.
\newblock {\em Journal of Clinical Epidemiology}, 173:111464, 2024.

\bibitem{harrell1982evaluating}
F.~E. Harrell, R.~M. Califf, D.~B. Pryor, K.~L. Lee, and R.~A. Rosati.
\newblock Evaluating the yield of medical tests.
\newblock {\em JAMA}, 247(18):2543--2546, 1982.

\bibitem{herold2020validation}
T.~Herold, M.~Rothenberg-Thurley, V.~V. Grunwald, H.~Janke, D.~Goerlich, M.~C. Sauerland, N.~P. Konstandin, A.~Dufour, S.~Schneider, M.~Neusser, et~al.
\newblock Validation and refinement of the revised 2017 european leukemianet genetic risk stratification of acute myeloid leukemia.
\newblock {\em Leukemia}, 34(12):3161--3172, 2020.

\bibitem{hothorn2006survival}
T.~Hothorn, P.~B{\"u}hlmann, S.~Dudoit, A.~Molinaro, and M.~J. Van Der~Laan.
\newblock Survival ensembles.
\newblock {\em Biostatistics}, 7(3):355--373, 2006.

\bibitem{hothorn2004bagging}
T.~Hothorn, B.~Lausen, A.~Benner, and M.~Radespiel-Tr{\"o}ger.
\newblock Bagging survival trees.
\newblock {\em Statistics in medicine}, 23(1):77--91, 2004.

\bibitem{huang2025deep}
L.~Huang, S.~Ruan, P.~Decazes, and T.~Den{\oe}ux.
\newblock Deep evidential fusion with uncertainty quantification and reliability learning for multimodal medical image segmentation.
\newblock {\em Information Fusion}, 113:102648, 2025.

\bibitem{huang2023application}
L.~Huang, S.~Ruan, and T.~Den{\oe}ux.
\newblock Application of belief functions to medical image segmentation: A review.
\newblock {\em Information fusion}, 91:737--756, 2023.

\bibitem{huang2024review}
L.~Huang, S.~Ruan, Y.~Xing, and M.~Feng.
\newblock A review of uncertainty quantification in medical image analysis: probabilistic and non-probabilistic methods.
\newblock {\em Medical Image Analysis}, page 103223, 2024.

\bibitem{huang2024evidential}
L.~Huang, Y.~Xing, T.~Denoeux, and M.~Feng.
\newblock An evidential time-to-event prediction model based on {Gaussian} random fuzzy numbers.
\newblock In {\em International Conference on Belief Functions}, pages 49--57. Springer, 2024.

\bibitem{ishwaran2008random}
H.~Ishwaran, U.~B. Kogalur, E.~H. Blackstone, and M.~S. Lauer.
\newblock Random survival forests.
\newblock {\em Annals of Applied Statistics}, 2(2):841--860, 2008.

\bibitem{ishwaran2011random}
H.~Ishwaran, U.~B. Kogalur, X.~Chen, and A.~J. Minn.
\newblock Random survival forests for high-dimensional data.
\newblock {\em Statistical Analysis and Data Mining: The ASA Data Science Journal}, 4(1):115--132, 2011.

\bibitem{johnson2020mimic}
A.~Johnson, L.~Bulgarelli, T.~Pollard, S.~Horng, L.~A. Celi, and R.~Mark.
\newblock {MIMIC-iv}.
\newblock \url{https://physionet. org/content/mimiciv/1.0/} (accessed August 23, 2021), 2021.

\bibitem{katzman2018deepsurv}
J.~L. Katzman, U.~Shaham, A.~Cloninger, J.~Bates, T.~Jiang, and Y.~Kluger.
\newblock Deepsurv: personalized treatment recommender system using a {Cox} proportional hazards deep neural network.
\newblock {\em BMC medical research methodology}, 18(1):1--12, 2018.

\bibitem{kvamme2019time}
H.~Kvamme, {\O}.~Borgan, and I.~Scheel.
\newblock Time-to-event prediction with neural networks and {Cox} regression.
\newblock {\em Journal of machine learning research}, 20(129):1--30, 2019.

\bibitem{lee2018deephit}
C.~Lee, W.~Zame, J.~Yoon, and M.~Van Der~Schaar.
\newblock Deephit: A deep learning approach to survival analysis with competing risks.
\newblock In {\em Proceedings of the AAAI conference on artificial intelligence}, volume~32, 2018.

\bibitem{li2020bayesian}
G.~Li, L.~Yang, C.-G. Lee, X.~Wang, and M.~Rong.
\newblock A {Bayesian} deep learning {RUL} framework integrating epistemic and aleatoric uncertainties.
\newblock {\em IEEE Transactions on Industrial Electronics}, 68(9):8829--8841, 2020.

\bibitem{lian2018joint}
C.~Lian, S.~Ruan, T.~Den{\oe}ux, H.~Li, and P.~Vera.
\newblock Joint tumor segmentation in {PET-CT} images using co-clustering and fusion based on belief functions.
\newblock {\em IEEE Transactions on Image Processing}, 28(2):755--766, 2018.

\bibitem{luck2017deep}
M.~Luck, T.~Sylvain, H.~Cardinal, A.~Lodi, and Y.~Bengio.
\newblock Deep learning for patient-specific kidney graft survival analysis.
\newblock {\em arXiv preprint arXiv:1705.10245}, 2017.

\bibitem{mehrtash2020confidence}
A.~Mehrtash, W.~M. Wells, C.~M. Tempany, P.~Abolmaesumi, and T.~Kapur.
\newblock Confidence calibration and predictive uncertainty estimation for deep medical image segmentation.
\newblock {\em IEEE transactions on medical imaging}, 39(12):3868--3878, 2020.

\bibitem{nguyen78}
H.~T. Nguyen.
\newblock On random sets and belief functions.
\newblock {\em Journal of Mathematical Analysis and Applications}, 65:531--542, 1978.

\bibitem{polsterl2015fast}
S.~P{\"o}lsterl, N.~Navab, and A.~Katouzian.
\newblock Fast training of support vector machines for survival analysis.
\newblock In {\em Machine Learning and Knowledge Discovery in Databases: European Conference, ECML PKDD 2015, Porto, Portugal, September 7-11, 2015, Proceedings, Part II 15}, pages 243--259. Springer, 2015.

\bibitem{purushotham2018benchmarking}
S.~Purushotham, C.~Meng, Z.~Che, and Y.~Liu.
\newblock Benchmarking deep learning models on large healthcare datasets.
\newblock {\em Journal of biomedical informatics}, 83:112--134, 2018.

\bibitem{raftery1996accounting}
A.~E. Raftery, D.~Madigan, and C.~T. Volinsky.
\newblock Accounting for model uncertainty in survival analysis improves predictive performance.
\newblock {\em Bayesian statistics}, 5:323--349, 1996.

\bibitem{shafer1976mathematical}
G.~Shafer.
\newblock {\em A mathematical theory of evidence}, volume~42.
\newblock Princeton University Press, 1976.

\bibitem{shivaswamy2007support}
P.~K. Shivaswamy, W.~Chu, and M.~Jansche.
\newblock A support vector approach to censored targets.
\newblock In {\em Seventh IEEE international conference on data mining (ICDM 2007)}, pages 655--660. IEEE, 2007.

\bibitem{tong2021evidential}
Z.~Tong, P.~Xu, and T.~Denoeux.
\newblock An evidential classifier based on {Dempster-Shafer} theory and deep learning.
\newblock {\em Neurocomputing}, 450:275--293, 2021.

\bibitem{van2007support}
V.~Van~Belle, K.~Pelckmans, J.~A. Suykens, and S.~Van~Huffel.
\newblock Support vector machines for survival analysis.
\newblock In {\em Proceedings of the third international conference on computational intelligence in medicine and healthcare (CIMED 2007)}, pages 1--8, 2007.

\bibitem{van2008survival}
V.~Van~Belle, K.~Pelckmans, J.~A. Suykens, and S.~Van~Huffel.
\newblock Survival {SVM}: a practical scalable algorithm.
\newblock In {\em ESANN}, pages 89--94, 2008.

\bibitem{van2011support}
V.~Van~Belle, K.~Pelckmans, S.~Van~Huffel, and J.~A. Suykens.
\newblock Support vector methods for survival analysis: a comparison between ranking and regression approaches.
\newblock {\em Artificial intelligence in medicine}, 53(2):107--118, 2011.

\bibitem{VANBELLE2011107}
V.~{Van Belle}, K.~Pelckmans, S.~{Van Huffel}, and J.~A. Suykens.
\newblock Support vector methods for survival analysis: a comparison between ranking and regression approaches.
\newblock {\em Artificial Intelligence in Medicine}, 53(2):107--118, 2011.

\bibitem{vinzamuri2017pre}
B.~Vinzamuri, Y.~Li, and C.~K. Reddy.
\newblock Pre-processing censored survival data using inverse covariance matrix based calibration.
\newblock {\em IEEE Transactions on Knowledge and Data Engineering}, 29(10):2111--2124, 2017.

\bibitem{wei1992accelerated}
L.-J. Wei.
\newblock The accelerated failure time model: a useful alternative to the {Cox} regression model in survival analysis.
\newblock {\em Statistics in medicine}, 11(14-15):1871--1879, 1992.

\bibitem{xu2016evidential}
P.~Xu, F.~Davoine, H.~Zha, and T.~Denoeux.
\newblock Evidential calibration of binary {SVM} classifiers.
\newblock {\em International Journal of Approximate Reasoning}, 72:55--70, 2016.

\bibitem{zadeh75}
L.~A. Zadeh.
\newblock The concept of a linguistic variable and its application to approximate reasoning --{I}.
\newblock {\em Information Sciences}, 8:199--249, 1975.

\bibitem{zadeh1978fuzzy}
L.~A. Zadeh.
\newblock Fuzzy sets as a basis for a theory of possibility.
\newblock {\em Fuzzy sets and systems}, 1(1):3--28, 1978.

\bibitem{zhang2018nonparametric}
Q.~Zhang and M.~Zhou.
\newblock Nonparametric {Bayesian Lomax} delegate racing for survival analysis with competing risks.
\newblock {\em Advances in Neural Information Processing Systems}, 31, 2018.

\bibitem{zhong2021gefitinib}
W.-Z. Zhong, Q.~Wang, W.-M. Mao, S.-T. Xu, L.~Wu, Y.-C. Wei, Y.-Y. Liu, C.~Chen, Y.~Cheng, R.~Yin, et~al.
\newblock Gefitinib versus vinorelbine plus cisplatin as adjuvant treatment for stage {II-IIIA (N1-N2) EGFR-mutant NSCLC}: final overall survival analysis of {CTONG1104 phase III trial}.
\newblock {\em Journal of clinical oncology}, 39(7):713, 2021.

\bibitem{zhu2017wsisa}
X.~Zhu, J.~Yao, F.~Zhu, and J.~Huang.
\newblock Wsisa: Making survival prediction from whole slide histopathological images.
\newblock In {\em Proceedings of the IEEE conference on computer vision and pattern recognition}, pages 7234--7242, 2017.

\end{thebibliography}
